\pdfoutput=1

\documentclass[sigconf]{aamas} 

\usepackage{balance} 
\usepackage{graphicx} 
\usepackage{subfig}
\usepackage{amsfonts}
\usepackage{amsmath}
\usepackage{placeins}
\usepackage{tabulary}
\usepackage{algorithm}
\usepackage{algorithmic}

\setcopyright{ifaamas}
\acmConference[AAMAS '24]{}{}{}{}
\copyrightyear{2024}
\acmYear{2024}
\acmDOI{}
\acmPrice{}
\acmISBN{}

\acmSubmissionID{XXX}

\title[Adaptive Discounting of Training Time Attacks]{Adaptive Discounting of Training Time Attacks}

\author{Ridhima Bector}
\affiliation{
  \institution{Nanyang Technological University}
  \country{Singapore}}
\email{ridhima001@e.ntu.edu.sg}

\author{Abhay Aradhya}
\affiliation{
  \institution{Nanyang Technological University}
  \country{Singapore}}
\email{abhayaradhya@ntu.edu.sg}

\author{Chai Quek}
\affiliation{
  \institution{Nanyang Technological University}
  \country{Singapore}}
\email{ashcquek@ntu.edu.sg}

\author{Zinovi Rabinovich}
\affiliation{
  \institution{Nanyang Technological University}
  \country{Singapore}}
\email{zinovi@ntu.edu.sg}

\begin{abstract}
Among the most insidious attacks on Reinforcement Learning (RL) solutions are 
training-time attacks (TTAs) that create loopholes and backdoors in the learned behaviour. Not limited to a simple disruption, {\em constructive} TTAs (C-TTAs) are now available, where the attacker forces a specific, target behaviour upon a training RL agent (victim). However, even state-of-the-art C-TTAs focus on target behaviours that could be naturally adopted by the victim if not for a particular feature of the environment dynamics, which C-TTAs exploit. In this work, we show that a C-TTA is possible even when the target behaviour is un-adoptable due to {\em both} environment dynamics {\em as well as} non-optimality with respect to the victim’s objective(s). To find efficient attacks in this context, we develop a specialised flavour of the DDPG algorithm, which we term $\gamma$DDPG, that learns this stronger version of C-TTA. $\gamma$DDPG dynamically alters the attack policy planning horizon based on the victim's current behaviour. This improves effort distribution throughout the attack timeline and reduces the effect of uncertainty the attacker has about the victim. To demonstrate the features of our method and better relate the results to prior research, we borrow a 3D grid domain from a state-of-the-art C-TTA for our experiments. Code is available at "bit.ly/github-rb-gDDPG".
\end{abstract}

\keywords{Dynamic Discount, Adaptive Discount, Constructive Training-Time Attacks, Environment Poisoning, Reinforcement Learning}

\newcommand{\BibTeX}{\rm B\kern-.05em{\sc i\kern-.025em b}\kern-.08em\TeX}

\begin{document}


\pagestyle{fancy}
\fancyhead{}


\maketitle 

\section{Introduction}
\label{sec:Introduction}


Success of RL stands threatened by a wide variety of attacks \cite{chen2019adversarial, ilahi2021challenges, demontis2022survey}, most insidious of which are training-time attacks (TTAs) that “pre-program” back-doors and behavioural triggers into an RL strategy \cite{sun2020vulnerability, rakhsha2021reward, banihashem2021defense, Hang2, Hang3, lu2022adversarial}. In TTAs, the attacker learns to optimally modify/poison a victim RL agent’s internal aspects (i.e., sensor(s), processor(s), memory) and/or external influences (i.e., environment) while the victim agent trains to learn its task. The level of information access assumed by the adversary categorises a TTA as white-box or black-box. White-box attacks \cite{rakhsha2020policy,zhang2020adaptive,sun2020vulnerability} assume access to one or more {\em internal} aspects of a victim, while black-box attacks \cite{Hang2,lu2022adversarial,Hang3} focus on external influences, addressing more realistic settings. This paper aims to develop and study a constructive environment-poisoning black-box TTA which modifies/poisons the dynamics of the victim agent’s environment without accessing any internal mechanism of the victim. Like prior works on constructive environment-poisoning black-box TTAs \cite{Hang2,Hang3}, the adversary in this research is an RL agent which learns the optimal TTA to be applied on the victim RL agent. However, unlike the prior works that enforce un-adoptable but optimal target behaviour on the victim agent and train the attack by infusing all attack objectives into the reward of the optimisation problem; our attack enforces a non-optimal target behaviour which is learned by distributing the attack objectives into the reward and the reward discounting factor of the attacker's optimisation problem.


In detail, we seek a constructive attack, i.e., the objective is to push the victim to acquire an attacker-desired target-behaviour. A target-behaviour can be any behaviour that the victim agent will not learn by itself in the original/default environment. The un-adoptability of this target behaviour in the original environment can be due to environment dynamics, non-optimality with respect to victim’s objectives, or both. Prior works focus on feasibility and hence experimented with target behaviours that are optimal (discrete environments) \citep{Hang2,Hang3} or nearly optimal (continuous environments) \citep{Hang3} with respect to the victim’s objective(s), but un-adoptable due to environment dynamics. This work develops and studies attacks wherein the target behaviour is un-adoptable in the original environment due to both, environment dynamics as well as non-optimality with respect to the victim’s objective. In addition to pushing the victim agent towards this non-optimal target behaviour, the attack must also preserve the environment as much as possible or, equivalently, reduce the effort expended to modify it. Attack actions are thus constrained by the magnitude of change a single attack action is permitted to make, as well as by treating environment modification effort as a second objective in the attacker's optimisation problem. The attacker, therefore, faces a multi-objective problem of finding an attack strategy that: a) generates the target behaviour in the victim with high accuracy, and b) has low-effort environment modifications.

Now, commonly, an RL agent's objectives are represented by a reward signal and the agent strives to find a behaviour/policy which, when executed in the given environment, maximises the produced cumulative reward. Likewise, in the attack domain, the attacker's reward typically inculcates both attack objectives: the accuracy with which the victim adopts the target behaviour, and the effort applied by the attacker, in terms of environment modifications, to achieve this accuracy. This can be done either by having several reward terms, allowing for prioritisation (through weights) of the attacker's objectives; or, by measuring the discrepancy between combined behaviour-environment pairs, as done in ~\cite{Hang2,Hang3}. More specifically, these works utilise the Kullback Leibler Divergence Rate (KLR) to provide a unified estimate of effort and effectiveness of an attack by measuring the discrepancy between the combination of the victim's current behaviour with the poisoned environment {\em and} the combination of the target behaviour with the default environment. However, both the aforementioned approaches have their shortcomings. Due to high symmetry, the KLR-based approach cannot properly distinguish between a high-accuracy, medium-effort behaviour-environment pair and a medium-accuracy, low-effort pair; while, weighted multiple terms of reward cannot address the fact that some behaviour-environment discrepancies cancel each other and, are thus, irrelevant.

In this paper, we propose an alternative route. We avoid packing both attack effort and effectiveness into a single element of the attacker's problem. Rather, we use both the reward and the reward discounting factor to encode {\em and} prioritise these objectives. We propose a modification of DDPG \cite{DBLP:journals/corr/LillicrapHPHETS15} called $\gamma$DDPG that supports dual-priority dual-objective optimisation with the aid of a dynamic discount function. Herein, the discount function, $\gamma$ adapts in response to the current level of effort exerted by the attacker (and the current level of attack accuracy achieved with that effort) to create a bounded search space that bounds the lower priority objective (attacker effort), and enables the attacker to optimise the higher-priority objective (attack accuracy) within this bounded space. Furthermore, given that large discount factors lead to unreliable optimisation in uncertain environments \citep{kim2022adaptive}, the bounded search space (created by the bounded discount function) in this work improves the optimisation capability of $\gamma$DDPG by reducing the effect of uncertainty in the given black-box environment.


\section{Related Work}
\label{sec:RelatedWork}

This section positions this paper in terms of the proposed methodology and the overall framework through comparison with literature on non-constant discounts and adversarial RL respectively.

    \subsection{Non-Constant Discounts}
    \label{subsec:RelatedWork/Discounts}


    Humans, as well as animals, tend to subjugate impulsive behaviours when the behaviours' absence increases the probability of a larger reward in the future \cite{miyazaki2012role,mischel1970attention}. In an attempt to explore reward optimisation with respect to flexible time horizons several works adapt the discount during training. \cite{franccois2015discount,hou2021improvement} gradually increase the discount factor over time and experimentally demonstrate the higher efficiency and better performance of this approach. \cite{zinzuvadiya2021state} increases the discount of states with high return estimates based on the intuition that the expected return of states on or near the optimal trajectory, increases during training. \cite{kim2022adaptive} strives to prevent reward overestimation in uncertain environments via advantage-based discount adaptation in policy gradient algorithms. This work computes two value functions using a small and a large discount respectively, and iteratively increases the small discount or decreases the large discount if it results in a lower advantage value than the other.
    
    \cite{wei2011markov,yoshida2013reinforcement} introduce optimal value functions using state-dependent discounts in model-based and model-free settings. The latter proves convergence to this optimal function and experimentally demonstrates the better performance of state-dependent discounts optimised via an evolutionary algorithm, compared to constant discounts. \cite{strand2022radar} first trains a set of low-level agents that use different constant discounts and then trains a high-level meta agent that learns to choose between the different low-level agent policies depending on the current state. \cite{white2017unifying} proposes a unified specification of episodic and non-episodic tasks using transition-based discounts and \cite{sharma2021transition} uses transition-based discounts in model-free algorithms to achieve faster convergence and learn risk-averse policies. The former is achieved via the inculcation of pessimism into the learning algorithm by increasing the discount of transitions that result in negative reward; while the latter is achieved by decreasing the discount for the preferred risk-averse trajectory. \cite{gu2021proximal} improves the accuracy of value estimation by utilising the probabilities of the current policy as the discount factor.
        
    In Multi-Objective MDPs (MO-MDPs) either the weight of each objective/reward is known or found via hyperparameter tuning and the MO-MDP is solved via transformation into a single-objective MDP or these weights are unknown and each objective is optimised separately to be either combined at the decision-making stage or be chosen by the user \cite{roijers2013survey}. \cite{gunarathna2022intelligent} falls under the former scenario and extends state-dependent discounts to multi-objective MDPs such that different objectives have different time scales. This work computes the expected return using a separate but constant discount for each objective in order to preserve the temporal scale of each objective. The $n^{th}$ step target Q-value inside the TD($\lambda$) update is discounted using a single variable discount which takes the value of the higher priority objective's discount as long as the reward corresponding to that objective is non-zero. This work, on one hand, requires specification of several hyperparameters in terms of weights as well as discounts of the different objectives, and on the other hand, works with a set of constant discounts. 

    This research introduces a novel dual-priority dual-objective MDP framework that neither needs weights and discounts to be known/estimated nor requires separate optimisation of each objective. In this framework, the higher priority (primary) objective is taken as the RL agent's reward while the lower priority (secondary) objective is used to condition the discount function. The discount adapts to the current state and modifies the algorithm's search space so as to optimise the primary objective while keeping the secondary objective bounded.

    \subsection{Adversarial Reinforcement Learning}
    \label{subsec:RelatedWork/constructiveTTA}

    In a typical adversarial RL study, an adversarial system is constructed that encompasses the RL agent, its environment, and its task. In these systems, the RL agent is regarded as the victim, its RL environment, the victim environment, and the task, the victim task. In addition to the victim, the system includes an adversary, tasked with attacking the victim agent. Since the attacker’s task is no easier than the victim’s, machine learning solutions (and RL, in particular) are generally deployed on the attacker’s side as well. All attack solutions are commonly classified by 4 features: the intent of attack (Destructive vs. constructive), the mode of attack (Reward vs. Observation vs. Environment), the level of access (White-box vs. Black-box) to the victim’s inner workings granted to the attacker, and the stage of attack (Training vs. Testing).

    \textbf{Attack Intent (Destructive vs. Constructive):} Destructive attacks \cite{huang2017adversarial,kos2017delving,zhang2021robust} degrade the victim RL agent's policy such that the policy performs poorly with respect to the victim's task. The victim, therefore, becomes incapable of solving its task. On the other hand, constructive attacks \cite{Hang2,Hang3,rakhsha2020policy} force the victim RL agent to learn a target policy that the victim will not learn by itself in the absence of the attack. The victim, therefore, becomes capable of carrying out the attacker desired task (target task). Given the higher level of complexity associated with constructive RL attacks, they form the subject of the current research investigation.

    \textbf{Attack Mode (Reward vs. Observation vs. Environment):} In reward \cite{zhang2020adaptive,rakhsha2021reward,banihashem2021defense} and observation \cite{zhang2021robust,huai2020malicious,sun2020vulnerability,lu2022adversarial} poisoning, the attacker generally possesses the ability to interfere with the victim's gratification and sensory circuitry and thereby modifies the victim's rewards and observations, respectively. In order to efficiently poison the rewards and/or observations of a victim, the attacker requires intrinsic information regarding the victim in the form of its learning algorithm, policy function, and/or preferences. However, such detailed victim-specific knowledge is impractical and often impossible to acquire in the real world. Environment poisoning \cite{rakhsha2020policy,Hang2,Hang3} mode of attack, on the other hand, does away with most of these requirements by directing the attack at the dynamics of the victim environment and is thus the chosen mode of attack in this research.

    \textbf{Attack Access (White- vs. Black-box):} The level of information access granted to the adversary categorises an adversary system as white-box or black-box. White-box systems \cite{rakhsha2020policy,sun2020stealthy,lutjens2020certified} allow the attacker access to one or more internal aspects of a victim and support all three modes of attack (reward, observation, and environment poisoning). In contrast, black-box systems \cite{lu2022adversarial,Hang2,Hang3} focus only on the more realistic, external influences (e.g., via environment poisoning). 
    We follow suit and adopt a black-box attack.

    \textbf{Attack Stage (Training vs. Testing):} An RL agent has two broad stages of operation corresponding to the training and testing/deployment phases of an RL algorithm. These two stages lead to two very different kinds of RL attacks, categorized as training-time (e.g.,~\cite{Hang2,Hang3,behzadan2017vulnerability,rakhsha2020policy}) and test-time (e.g.,~\cite{sun2020stealthy,lutjens2020certified,huang2017adversarial,lin2017tactics}) attacks respectively. Training-time attacks take place when the victim agent is in the process of learning the optimal policy ( w.r.t. the victim's objectives). Consequently, during a training-time attack, the victim's policy undergoes periodic updates. In contrast, test-time attacks take place after the victim RL agent has finished training. Therefore, during a test-time attack, the victim's policy remains fixed. This research studies training-time attacks as they are potentially more malicious, given their capability of permeating loopholes and behavioural triggers into the RL agent's "apparently" optimal strategy. These behavioural triggers can then be used during deployment to make the victim RL agent carry out the target behaviour.

\begin{figure}[ht]
    \centering
    \includegraphics[width=0.8\linewidth]{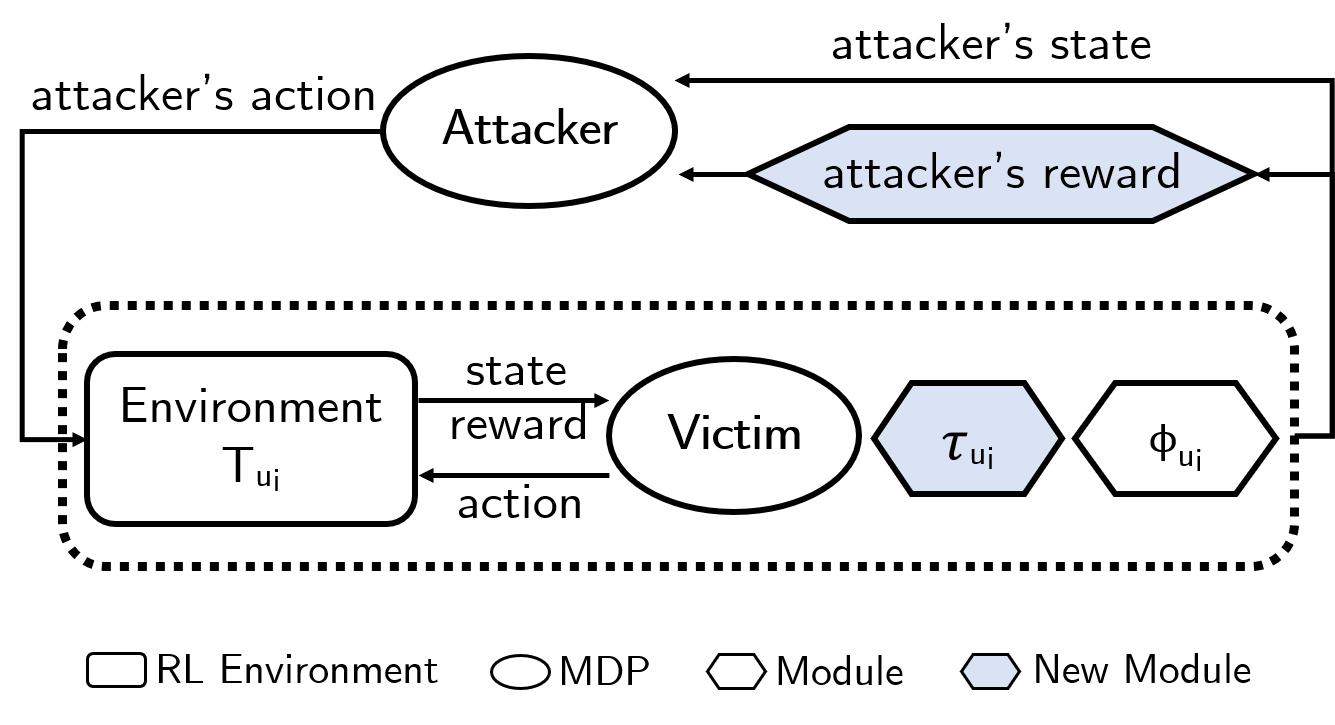}
    \caption{Bi-Level Attack Framework}
    \label{fig:Architecture}
\end{figure}

    The state-of-the-art constructive, environment-poisoning, black-box TTAs; TEPA \cite{Hang2} and DBB-EPA \cite{Hang3} follow a bi-level hierarchical framework (Figure~\ref{fig:Architecture}), wherein the attacker as well as the victim is an independent reinforcement-learning agent with its individual learning algorithm and policy. Within this framework, in order to learn a given task, the victim trains to maximise its cumulative discounted rewards by interacting with the victim environment, unaware of the attacker. The attacker, on the other hand, observes these interactions of the victim with its environment and takes an action to modify the victim environment. The goal of the attacker is to sequentially and minimally modify the victim's environment dynamics to drive the victim to adopt the attacker-desired target behaviour. Therefore, the overall system is formed by two nested closed-loop processes, wherein the attacker, as well as the victim, is modelled as a Markov Decision Process (MDP).
        
    \textbf{Victim MDP:} The victim's MDP can be denoted by the tuple $< S, A, T_{u_i}, R_v, q_0, \gamma_v >$ where $S = s_1, s_2, ...$, and $A = a_1,a_2, ...$ are the victim's states and actions respectively; $R_v: S \times A \times S \rightarrow \mathbb{R}$ is the reward function which encodes the victim's task; $\gamma_v \in (0,1)$ is the discount factor, $q_0(S)$ is the distribution over initial states; and, $T_{u_i}: S \times A \times S \rightarrow [0,1]$ is the probabilistic transition function, where $u_i$ denotes the environment parameterisation that has resulted from the first $i$ interventions on the environment, by the attacker. In particular, $T_{u_0}$ refers to the original, unaltered dynamics of the victim environment. The objective of the victim is to find an optimal policy within the experienced environment.

    \textbf{Attacker MDP:}
    The attacker's Markov process can be represented by the tuple $< X, U, F, R, \tau^*, \gamma >$,  where: $X$ is the attacker's state space;
    $U$ is the attacker's action space, i.e., the set of all permissible changes that can be applied to the victim environment dynamics.
    Aggregate attack notation is common, i.e., when action $u_i$ is applied on the environment with dynamics $T_{u_{i-1}}$, it results in an environment with dynamics $T_{u_i}$. Thus, environment changes by attack actions $u_0, u_1, ..., u_i$ {\em accumulate} to create $T_{u_i}$.
    %
%
    $F: X \times U \times X \rightarrow [0,1]$ is the stochastic transition function that describes the response of the victim to environmental experiences, i.e., how the victim's behaviour changes in response to changes in the environment dynamics; $R: X \times U \times X \rightarrow \mathbb{R}$ is the attacker's reward function that encapsulates the attacker's objectives and $\gamma: X \rightarrow [0.0, 1.0]$ is the discount function. 
    The attacker seeks the most {\em efficient} way to push the victim to converge to the target policy $\tau^*$.


\section{Methodology : $\mathbf{\gamma}$-variant DDPG}
\label{sec:Methodology}

The current research adopts the system architecture developed by the state-of-the-art constructive, environment-poisoning, black-box TTAs \cite{Hang2,Hang3} but modifies the attacker MDP to incorporate a novel state-space, a single-objective reward function and an adaptive/dynamic discount function.
Like prior works, the attacker's Markov process is represented by the tuple $< X, U, F, R, \tau^*, \gamma >$. However, unlike prior works: $X = [T_{u_{i-1}}, \phi_{u_{i-1}}]$ is the attacker's state space comprising the victim environment dynamics, $T_{u_{i-1}}$ and the victim's behaviour, $\phi_{u_{i-1}}$ that emerged in response to those dynamics; $R: X \times U \times X \rightarrow \mathbb{R}$ is the attacker's reward function that describes attack effectiveness, i.e., how close the victim's behaviour is to the target (attacker-desired ideal) behaviour $\tau^*$; and $\gamma: X \rightarrow [\gamma_{min},\gamma_{max}]$ is the adaptive/dynamic discount function that tunes the importance of long-term rewards based on the current attacker state $x_{u_{i-1}}$. The attacker seeks to optimise its expected total discounted reward, wherein the combination of $R$ and $\gamma$ simulate dual optimisation of (maximum) attack effectiveness with (minimum) effort, by a policy of the form $\sigma:X \rightarrow U$, $\sigma(u_i|x_{i-1})$.

\begin{algorithm}[t]
\caption{$\boldsymbol{\gamma}$DDPG Algorithm}
\begin{algorithmic}[1]

    \STATE Randomly initialise critic $Q(x,u|\theta^Q)$ and actor $\sigma(x|\theta^\sigma)$ networks with weights $\theta^Q$ and $\theta^\sigma$
    \STATE Initialise target networks $Q'$ and $\sigma'$ with weights $\theta^{Q'} \leftarrow \theta^Q, \theta^{\sigma'} \leftarrow \theta^\sigma$
    \STATE Initialise replay buffer $R$
    
    \FOR{episode = 1, $M_a$}
        \STATE Initialise a random process $\chi$ for action exploration
        \STATE Receive initial observation state $x_0$
        
        \FOR{t = 1, $T_a$}
            \STATE Select action $u_{t} = \sigma(x_{t-1}|\theta^\sigma) + \chi_{t}$ according to the current policy and exploration noise
            \STATE Execute action $u_t$ to poison environment $T_{u_{t-1}}$
            \STATE Observe reward $r_{t}$  
            \STATE $x_t \leftarrow [T_{u_t}$, Auto\_Encoder( Algorithm \ref{algo:tau} ($T_{u_t}$) )]
            \STATE \textbf{Compute $\boldsymbol{\gamma_t}$ using Equation \ref{eq:discountFunction}}
            \STATE \textbf{Store transition $\boldsymbol{(x_{t-1}, u_t, r_t, x_t, \gamma_t)}$ in $\boldsymbol{R}$}
            \STATE \textbf{Sample a random minibatch of $\boldsymbol{N}$ transitions $\boldsymbol{(x_{i-1}, u_i, r_i, x_i, \gamma_i)}$ from $\boldsymbol{R}$}
            \STATE \textbf{Set $\boldsymbol{y_i = r_i + \gamma_i Q'(x_i, \sigma'(x_i|\theta^{\sigma'})|\theta^{Q'})}$}
            \STATE Update critic $\sigma$ by minimising the loss: $L = \frac{1}{N} \sum_{i}(y_i - Q(x_{i-1}, u_i|\theta^Q))^2$
            \STATE Update the actor policy $Q$
            \STATE Update the target networks $Q'$ and $\sigma'$
            
        \ENDFOR

    \ENDFOR
\end{algorithmic}
\label{algo:DDPGgamma}
\end{algorithm}

\begin{algorithm}[t]
\caption{$\boldsymbol{\tau}$ Computation Algorithm}
\begin{algorithmic}[1]

    \STATE Receive victim environment with dynamics $T_{u_i}$
    \IF{i == 1}
        \STATE Initialise victim's Q table
    \ELSE
        \STATE Use victim's Q table from $T_{u_{i-1}}$
    \ENDIF
    \STATE Initialise $\tau_{u_i}$ with the no-action symbol for each state
    \STATE Initialise state

    \FOR{episode = 1, $M_v$}
        \WHILE{done != True}
            \STATE action = Softmax\_Action(Q)
            \STATE next\_state, reward, done = Env-$T_{u_i}$(action)
            \STATE $\tau_{u_i}$(state) $\leftarrow$ action
            \STATE Q $\leftarrow$ TD\_Update(Q)
            \STATE state $\leftarrow$ next\_state
        \ENDWHILE
    \ENDFOR
    
\end{algorithmic}
\label{algo:tau}
\end{algorithm}

    \subsection{State Space}
    \label{subsec:Methodology/StateSpace}
    
    $\gamma$DDPG (Algorithm \ref{algo:DDPGgamma}), a dual-priority dual-objective RL algorithm,  is developed to enable the attacker to learn attack strategies that can push the victim RL agent towards target behaviours that are non-optimal with respect to the victim's learning objectives.
    Every action $u_i$ of the attack is conditioned on the current victim behaviour $\phi_{u_{i-1}}$ and the current victim environment dynamics $T_{u_{i-1}}$. We assume that this attack conditioning occurs in a black-box setting, i.e., without any access to the victim's inner mechanisms or representations, during both, the attacker's training {\em and} testing. Thus, the victim's behaviour can only be approximated through across-policy behaviour traces that the attacker observes while the victim trains. In general, the victim would update its policy with a non-trivial frequency. This frequency can be so high that each state-action pair of its behaviour trace would originate from a slightly different policy. In our experiments, we assume as much. Now, to approximate the victim's policy we would need traces of multiple epochs of the victim's training process. But conditioning an attack on such a large volume of data is impractical. Instead, we preprocess these traces by storing the last observed victim action corresponding to each observed victim state and assign a "no-action" symbol to unvisited states \cite{ijcai2023p386}. This behaviour information will hereafter be denoted as $\tau_{u_{i-1}} = \{ s_1, a_1; s_2, a_2; ...; s_N, a_N\} \forall s_n \in S$, where $a_n$ is the latest action taken by the victim in state $n$ or the no-action symbol in case state $s_n$ was never visited by the victim, and $N$ is the total number of states in the victim environment. 
    
    As $\tau_{u_{i-1}}$ contains the latest action / no-action symbol corresponding to all states; $\tau_{u_{i-1}}$'s size can explode in high-dimensional environments. To combat this issue, this paper learns a low-dimensional latent space, $\Phi$ of victim behaviours using an auto-encoder model (Eq \ref{eq:ae}). The model consists of an encoder $q_{e}$ that takes the victim's $\tau_{u_{i-1}}$ as input and outputs the corresponding latent behaviour $\phi_{u_{i-1}}$; and a decoder $q_{d}$ that takes two inputs, the latent behaviour $\phi_{u_{i-1}}$ and a victim environment state $s_n$, and outputs the probability with which the victim will take each available action in the given state $s_n$. Finally, $\phi_{u_{i-1}}$ and the current environment dynamics $T_{u_{i-1}}$ together constitute the current state of the attacker (Eq \ref{eq:attackstate}).

    \begin{equation}
        \begin{aligned}
            \phi_{u_{i-1}} = q_{e}( &\tau_{u_{i-1}} )\\
            a =  argmax( q_{d}( &\phi_{u_{i-1}}, s_{n} ) )
        \end{aligned}
    \label{eq:ae}
    \end{equation}

    \begin{equation}
        \textrm{Attacker State} ~ = ~ x_{i-1} ~ = ~ [ T_{u_{i-1}}, \phi_{u_{i-1}} ]
    \label{eq:attackstate}
    \end{equation}

    \subsection{Adaptive Discount Function}
    \label{subsec:Methodology/Discount}

    As suggested in the attacker-MDP, this work seeks to balance attack effectiveness and effort. However, unlike prior works, where the balance was achieved through {\em reward} elements' merge, we distribute the responsibility between distinct MDP components. Herein, the reward function encapsulates the higher-priority objective (attack accuracy) while the dynamic discount manages both the higher and lower priority objectives (attack accuracy and effort). We implement this architecture in our $\gamma$DDPG algorithm (Algorithm \ref{algo:DDPGgamma}), appropriately subverting the original DDPG \citep{DBLP:journals/corr/LillicrapHPHETS15} in lines 12-15.
    
    By optimising attack accuracy within effort+accuracy-bounded search space, $\gamma$DDPG becomes capable of prioritising attack accuracy over attacker effort. At each attack timestep, $\gamma$DDPG's adaptive discount factor creates a bounded effort+accuracy search space similar to the trust regions created by adaptive step sizes \cite{schulman2015trust}; and the attacker looks for the highest accuracy state within this space. This bounded search space is created by adapting the attacker MDP's discount factor ($\gamma$) conditioned on the current attacker effort and attack accuracy. The adaptive discount factor modifies $\gamma$DDPG's Bellman update to alter the level of importance that the algorithm accords to long-term rewards. Given that smaller discounts bias an RL algorithm's precedence to short-term rewards, a decrease in discount results in tightening of the search space  of the attacker around the current state, while an increase in discount widens the search space of the attacker around the current state.

    The adaptive discount; which is a function of the effort executed by the attacker on the victim environment and the accuracy with which the victim adopts the target behaviour; can be modelled in different ways. As mentioned in Section \ref{sec:Introduction}, prior works \cite{rabinovich2010cultivating,Hang2} utilise negative KLR between the vanilla-current (current environment * current behaviour) and perfect (original environment * target behaviour) MDPs as the attacker reward because reducing this KLR pushes the current MDP towards the perfect MDP (high @Acc and low @Effort). This work adopts the same formulation (distance/divergence between vanilla-current and perfect MDPs) to construct dynamic discounts conditioned on both, attack accuracy and attacker effort and proposes a new formulation (distance/divergence between target-current and perfect MDPs) to construct dynamic discounts conditioned only on attacker effort. The target-current MDP models the current environment dynamics coupled with the victim's target behaviour as a stochastic Markov process, thereby computing target-behaviour conditioned effort associated with the current state. The vanilla-current $P^v_{u_i}$, target-current $P_{u_i}$ and perfect $P^*_{u_0}$ MDPs are defined in Equations \ref{eq:VCurrentProcess}, \ref{eq:TCurrentProcess}, and \ref{eq:PerfectProcess} where $j$ denotes victim-level time step, just as $i$ denotes attacker-level time step. This study employs the partial target behaviour design of \cite{Hang2} to construct the attacker-desired target policy $\tau^*$:

    \begin{equation}
        \tau^*_{u_i}(s) = 
        \begin{cases}
            a^*_n &  s_n \in S^* \\
            \tau_{u_i}(s_n) & s_n \not\in S^*
        \end{cases}
    \end{equation}

    Here $S^*$ is the target state set, $a^*_n$ is the target action for target state $s_n$, and $\tau_{u_i}(s)$ is the latest behaviour of the victim observed in the environment with transition dynamics $T_{u_i}$. As the attacker cannot access the victims' policy in the given black-box setting, it approximates the victim policy $\pi_{u_i}$, using the last $h$ actions taken by the victim in each state. The target policy distribution $\pi^*_{u_i}$ is then constructed by taking a copy of $\pi_{u_i}$ and modifying it by assigning probability 1.0 to all target actions (and 0.0 to non-target actions) w.r.t. each target state.

    \begin{equation}
        \begin{aligned}
            \textrm{Vanilla}&\textrm{-Current MDP} = P^v_{u_i}(s_{j+1},a_{j+1}|s_j,a_j) \\
            &= T_{u_i}(s_{j+1}|s_j,a_j) \ \pi_{u_i}(a_{j+1}|s_{j+1})
        \end{aligned}
    \label{eq:VCurrentProcess}
    \end{equation}
    
    \begin{equation}
        \begin{aligned}
            \textrm{Target}&\textrm{-Current MDP} = P_{u_i}(s_{j+1},a_{j+1}|s_j,a_j) \\
            &= T_{u_i}(s_{j+1}|s_j,a_j) \ \pi^*_{u_i}(a_{j+1}|s_{j+1})
        \end{aligned}
    \label{eq:TCurrentProcess}
    \end{equation}

    \begin{equation}
        \begin{aligned}
            \textrm{Pe}&\textrm{rfect MDP} = P^*_{u_0}(s_{j+1},a_{j+1}|s_j,a_j) \\
            &= T_{u_0}(s_{j+1}|s_j,a_j) \ \pi^*_{u_i}(a_{j+1}|s_{j+1})
        \end{aligned}
    \label{eq:PerfectProcess}
    \end{equation}

    Like prior works, we can use KLR to compute the divergence between these MDPs. KLR computes this divergence by calculating the divergence between probability distributions of different trajectories in the two given Markov processes. However, KLR does not take the underlying metric space into account. This work hypothesises that a measure that computes the distance between the $k^{th}$ step probability distributions of two Markov processes while respecting the underlying geometry of the metric space provides a better estimate of the difference between the two given Markov processes as compared to KLR. To test this hypothesis we use Wasserstein distance to compute the distance between the vanilla-current and perfect MDPs (termed WD) as well as target-current and perfect MDPs (termed TargetWD). Wasserstein distance possesses an additional property of being insensitive to small changes in the probability distributions. This property is advantageous in the current uncertain black-box setting where $\tau_{u_i}$ being an approximate representation of the actual policy of the victim can be noisy in nature.
    Let $p^{k^v}_{u_i}$ and $p^{k*}_{u_0}$ be the $k^{th}$ step probability distributions of the vanilla-current and perfect processes respectively. The Wasserstein 1-distance between these distributions termed WD is defined below. Here, $\beta$ is a transport plan, $d(x,y)$ is the distance between $x$ and $y$, and $p^{k^v}_{u_i}$ and $p^{k*}_{u_0}$ are written as $p^{k^v}$ and $p^{k*}$ respectively.

    \begin{equation}
        \begin{aligned}
            &\boldsymbol{\gamma} \propto \textrm{WD} (p^{k^v}, p^{k*} ):=\\
            &\left(\inf _{\beta \in \textrm{B} ( p^{k^v}, p^{k*} )} {\rm I\!E}_{(x,y)\sim\beta} d(x,y)\,\mathrm {d} \beta (x,y)\right)
        \end{aligned}
    \label{eq:discountFunction}
    \end{equation}

    In order to better understand the individual contributions of the Wasserstein metric (in comparison to KLR) and target-current MDP based distance in bounding the lower-priority objective (@Effort) and reducing uncertainty, this paper experiments with four adaptive/dynamic discount functions. These four adaptive discount functions must undergo normalisation as the range of KL divergence \citep{kullback1951information} and Wasserstein distance \citep{vaserstein1969markov} is $[0, \infty]$. Details regarding this normalisation are presented in Appendix 7.2.

    \begin{itemize}
        \item WD - WD(Vanilla-Current MDP, Perfect MDP)
        \item KLR - KLR(Vanilla-Current MDP, Perfect MDP)
        \item TargetWD - WD(Target-Current MDP, Perfect MDP)
        \item TargetKLR - KLR(Target-Current MDP, Perfect MDP)
    \end{itemize}

    \begin{figure*}[ht]%
        \begin{tabular}{cccc}
        \subfloat[@Acc KDE]{\includegraphics[width=0.2\linewidth,trim={0cm 0cm 7cm 0cm},clip]{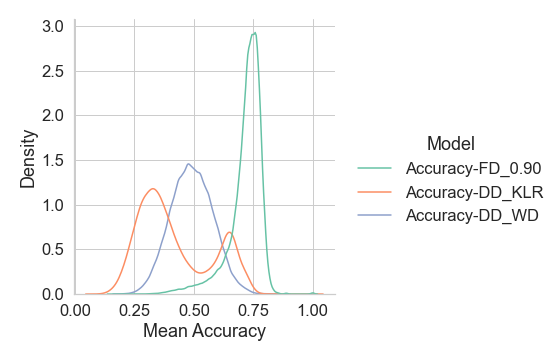}} &
        \subfloat[@SoftAcc KDE]{\includegraphics[width=0.2\linewidth,trim={0cm 0cm 7cm 0cm},clip]{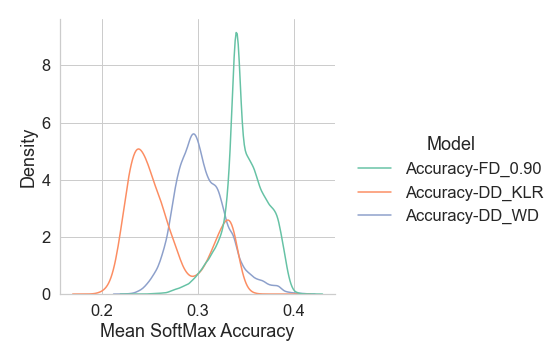}} &
        \subfloat[@Effort KDE]{\includegraphics[width=0.2\linewidth,trim={0cm 0cm 7cm 0cm},clip]{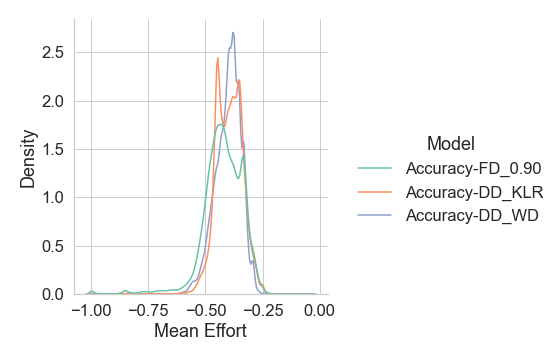}} &
        \subfloat[Legend]{\includegraphics[width=0.16\linewidth,trim={13cm 3cm 0cm 2cm},clip]{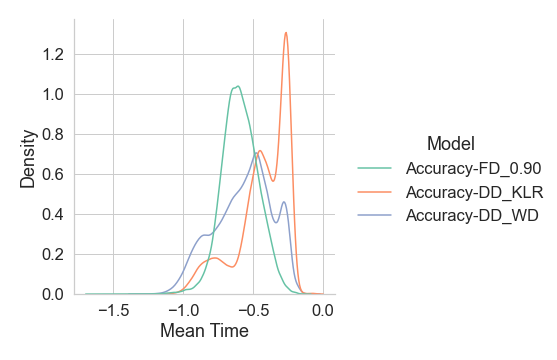}}\\
        
        \subfloat[@Acc Line Graph]{\includegraphics[width=0.24\linewidth,trim={2.5cm 0.0cm 2.5cm 2.5cm},clip]{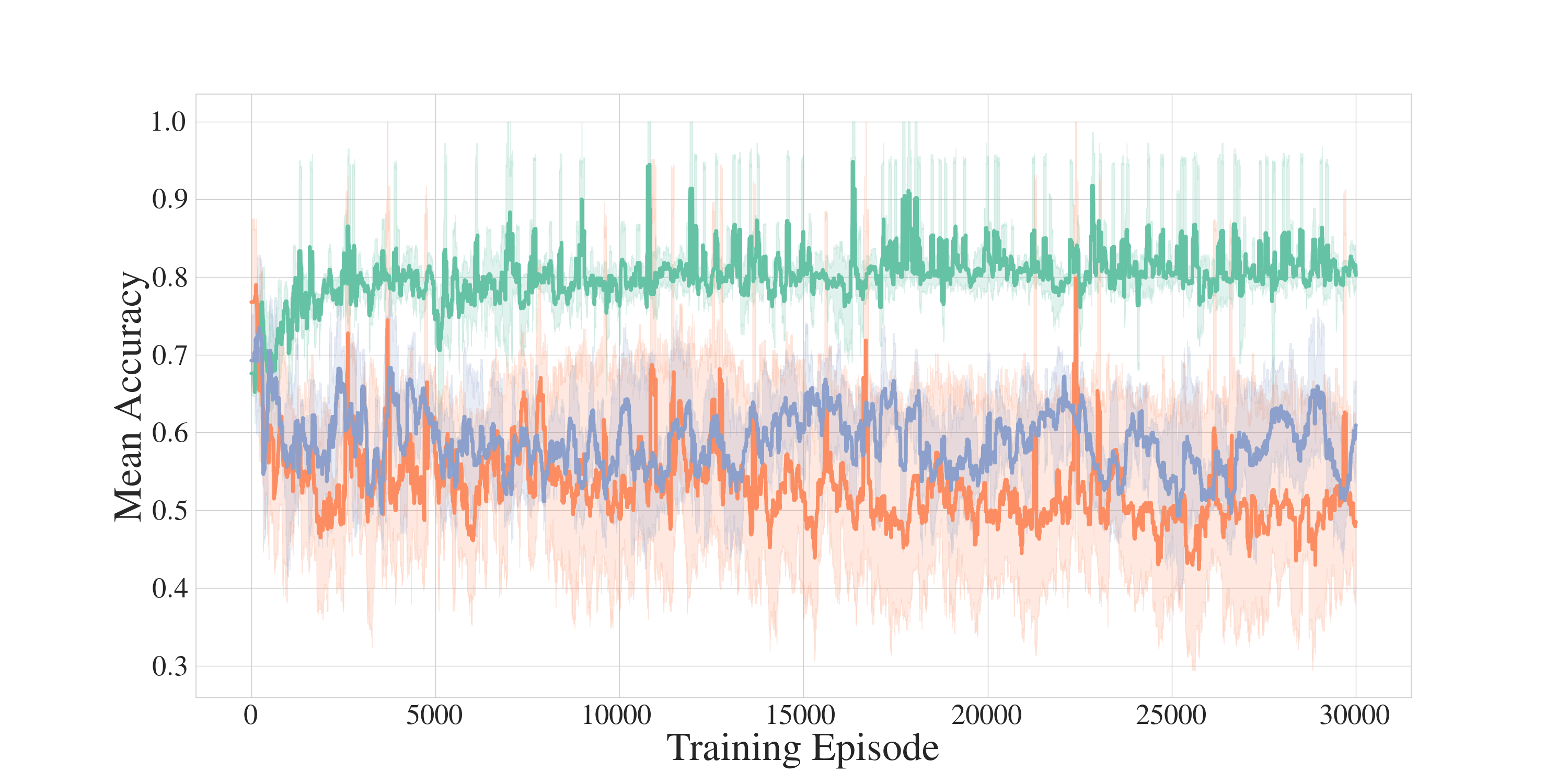}} &
        \subfloat[@SoftAcc Line Graph]{\includegraphics[width=0.24\linewidth,trim={2.5cm 0.0cm 2.5cm 2.5cm},clip]{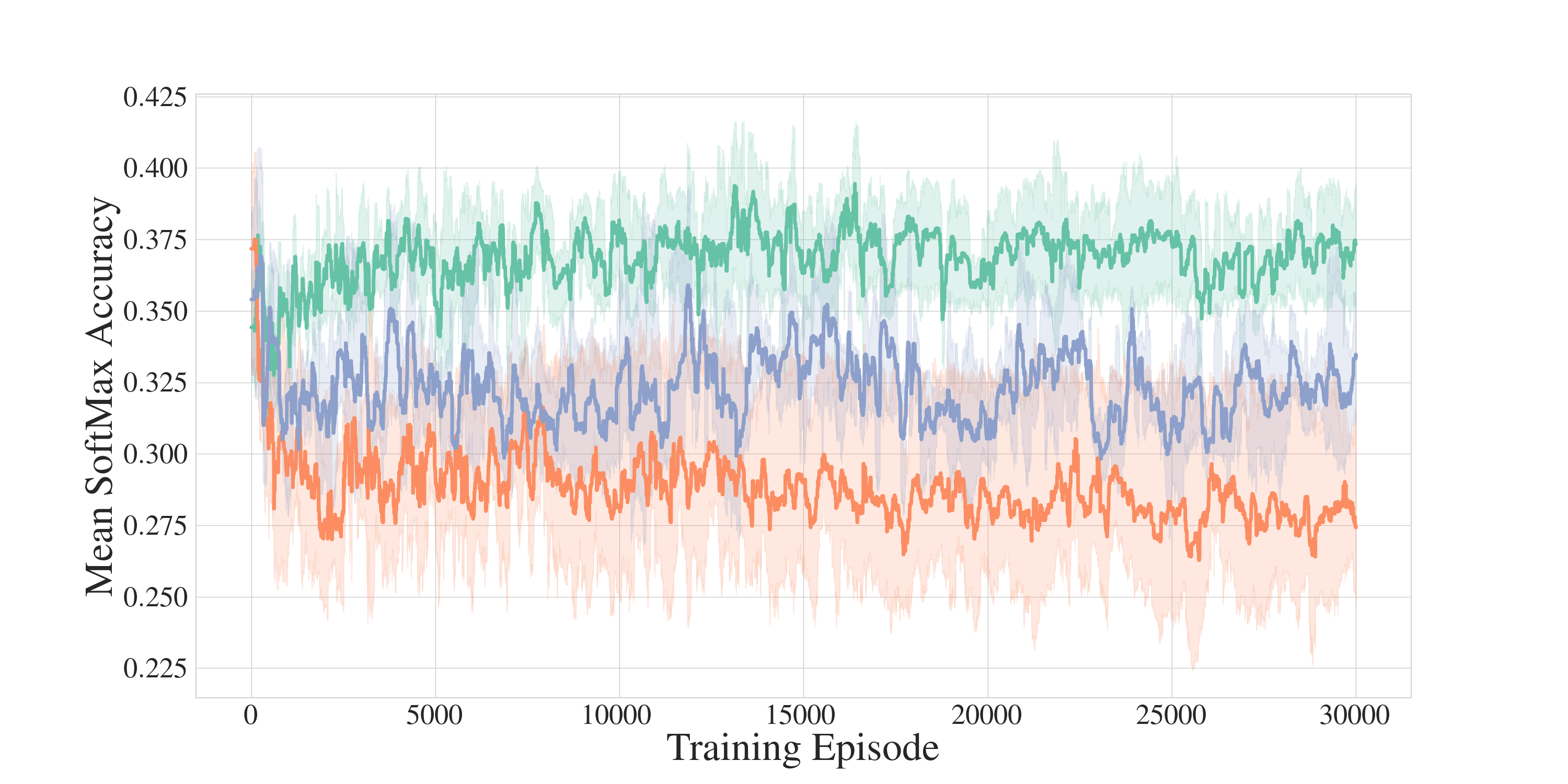}} &
        \subfloat[@Effort Line Graph]{\includegraphics[width=0.24\linewidth,trim={2.5cm 0.0cm 2.5cm 2.5cm},clip]{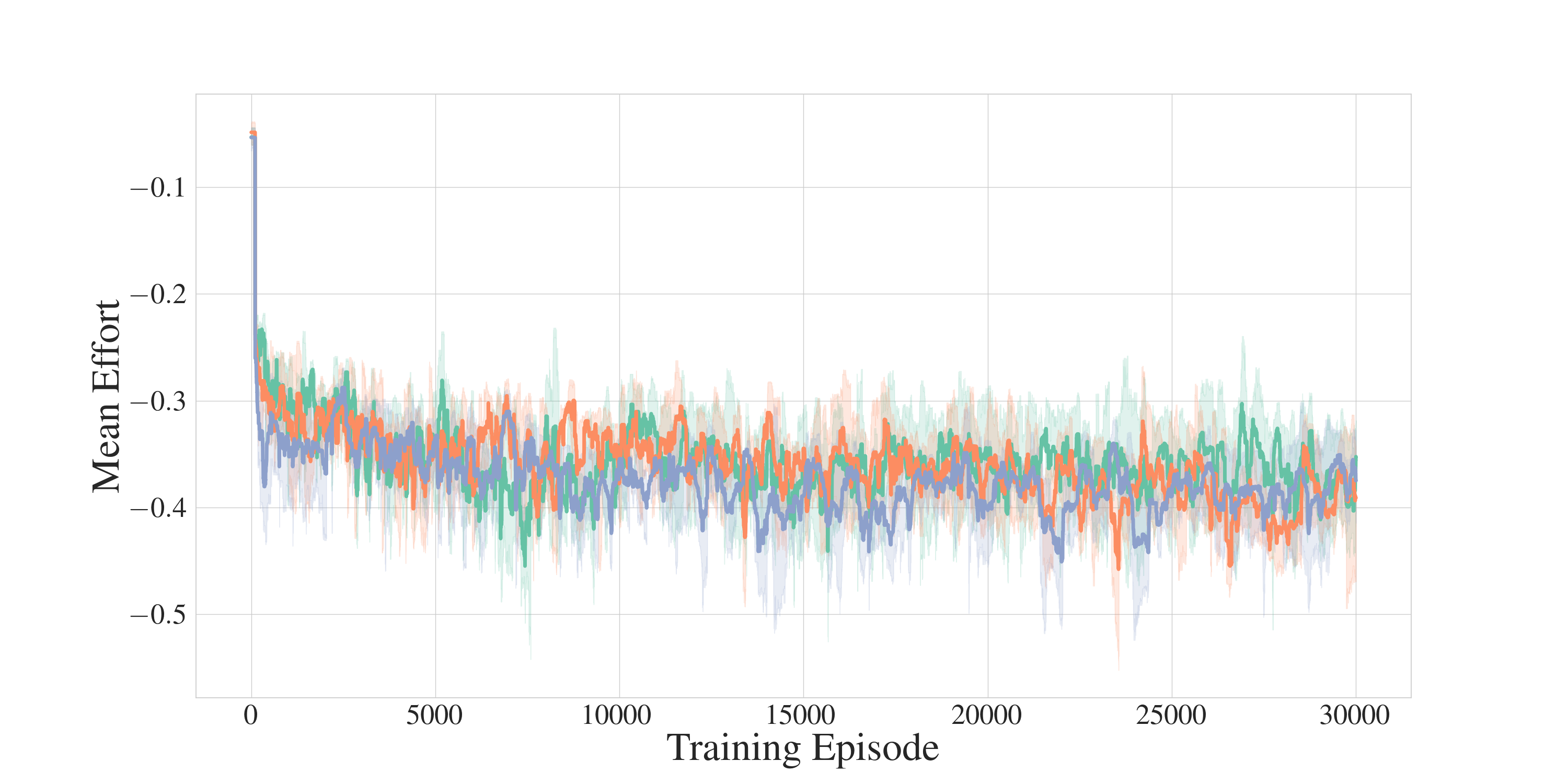}} &
        \subfloat[@Time Line Graph]{\includegraphics[width=0.24\linewidth,trim={2.5cm 0.0cm 2.5cm 2.5cm},clip]{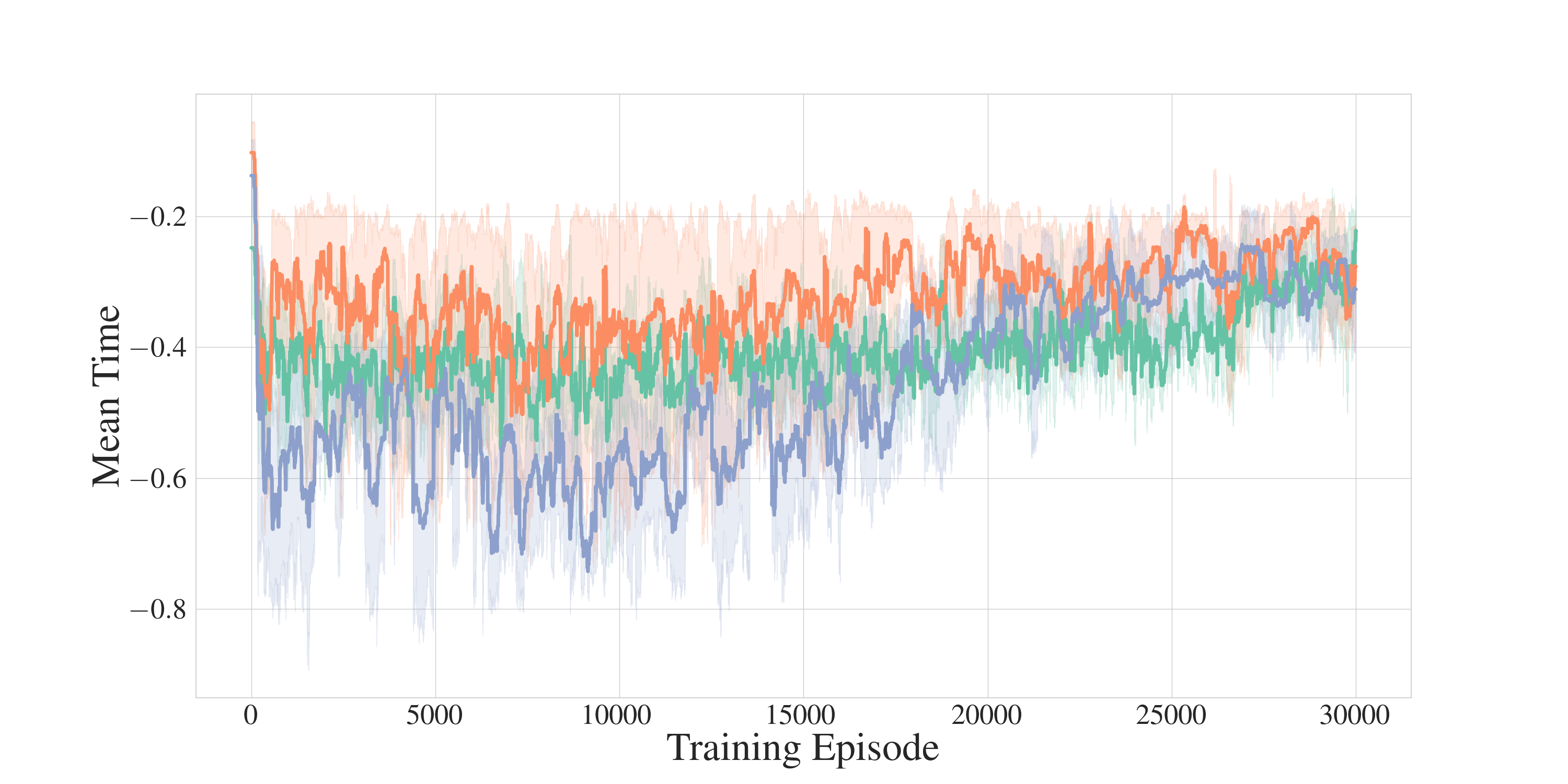}}\\

        \subfloat[Test-Time @Acc]{\includegraphics[width=0.2\linewidth,trim={0cm 0cm 5cm 0cm},clip]{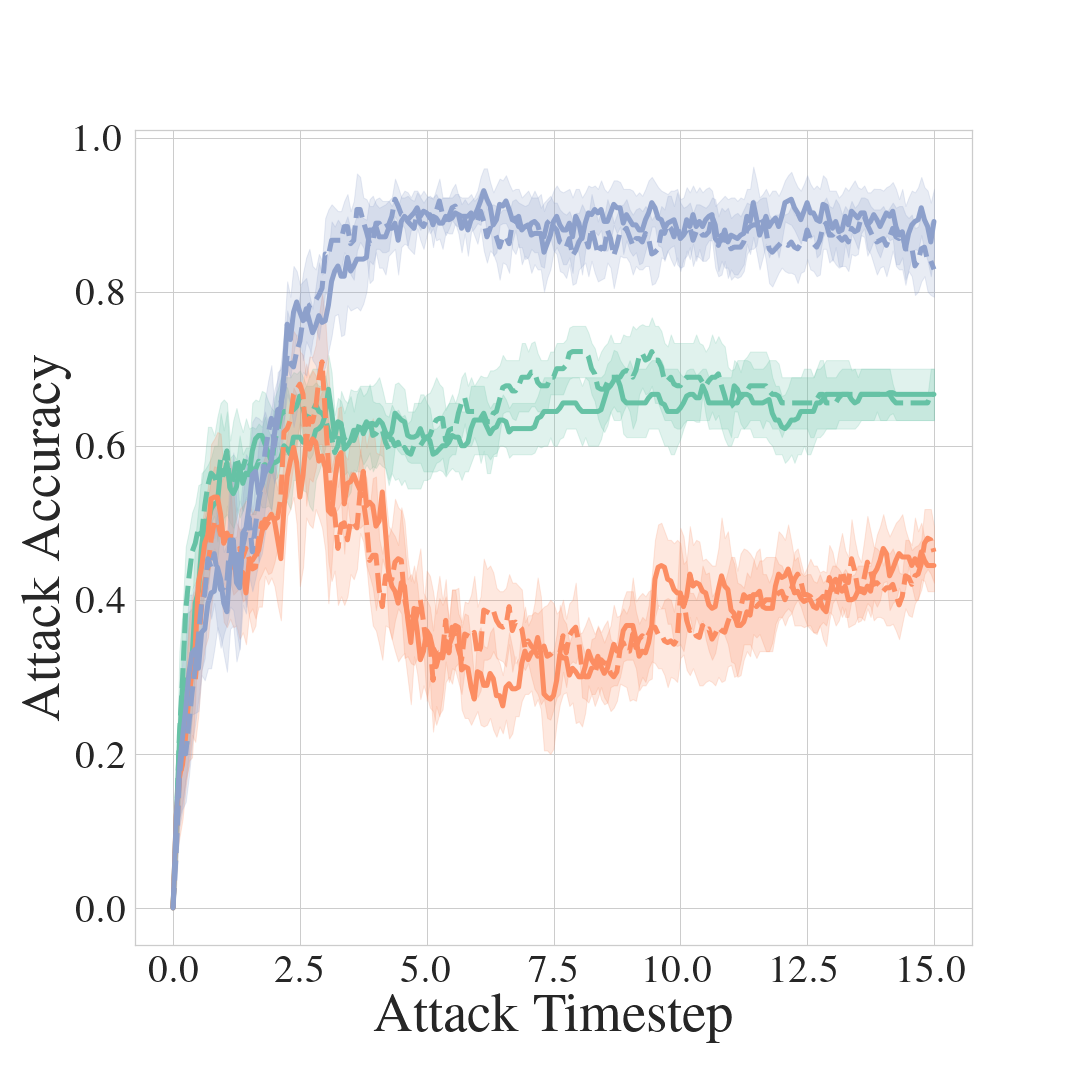}} &
        \subfloat[Test-Time @SoftAcc]{\includegraphics[width=0.2\linewidth,trim={0cm 0cm 5cm 0cm},clip]{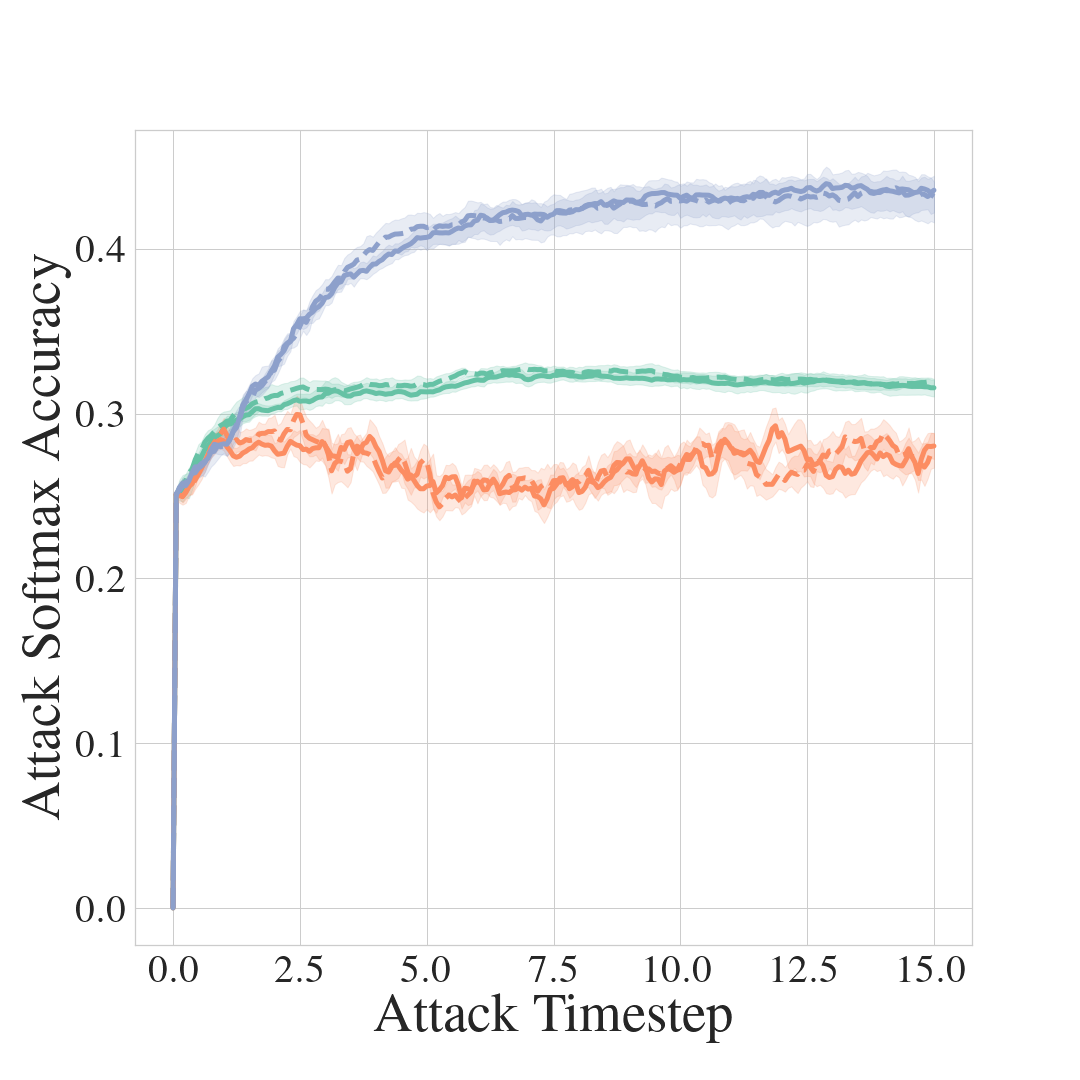}} &
        \subfloat[Test-Time @Effort]{\includegraphics[width=0.2\linewidth,trim={0cm 0cm 5cm 0cm},clip]{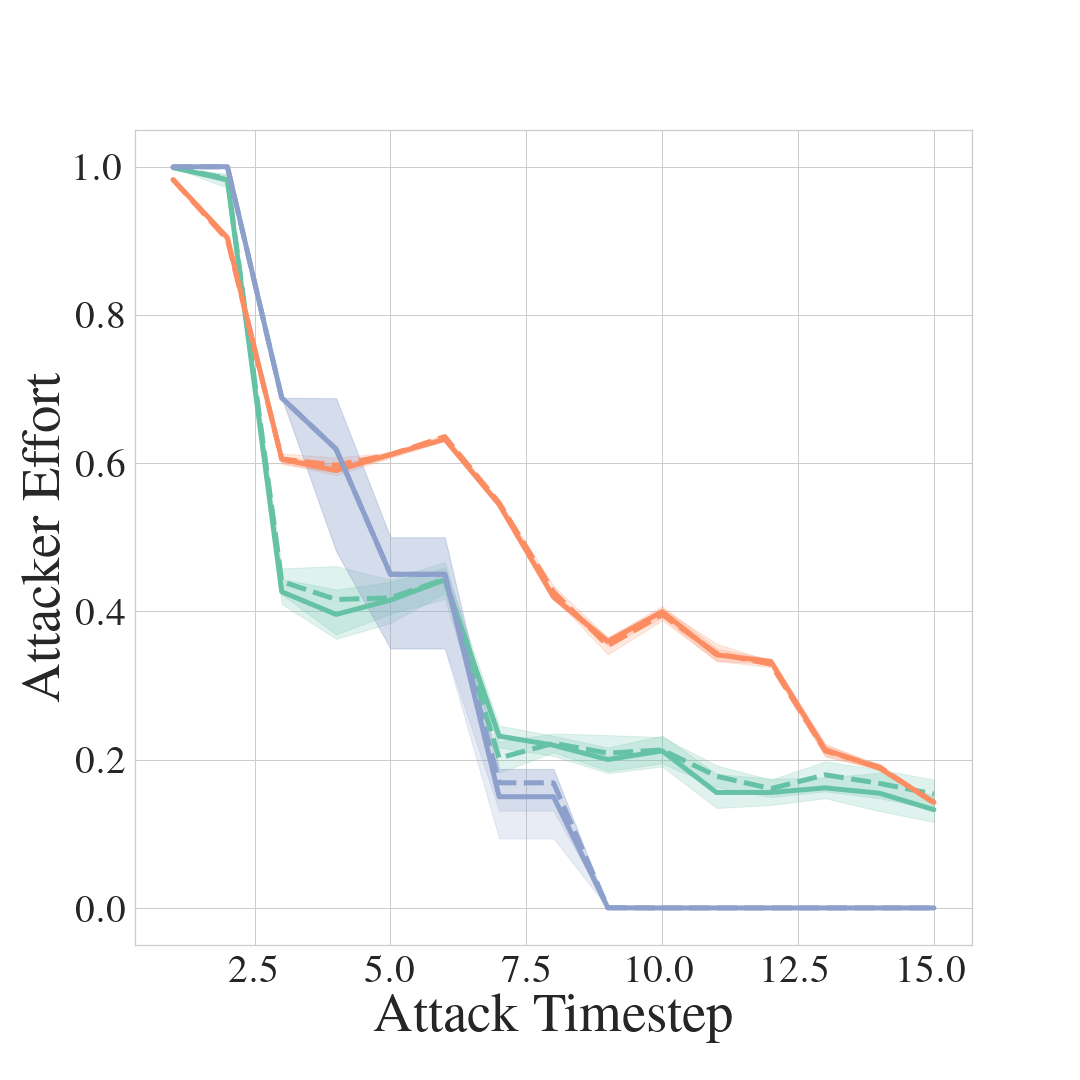}} &
        \subfloat[Test-Time @Time]
        {\includegraphics[width=0.2\linewidth,trim={0cm 0cm 5cm 0cm},clip]{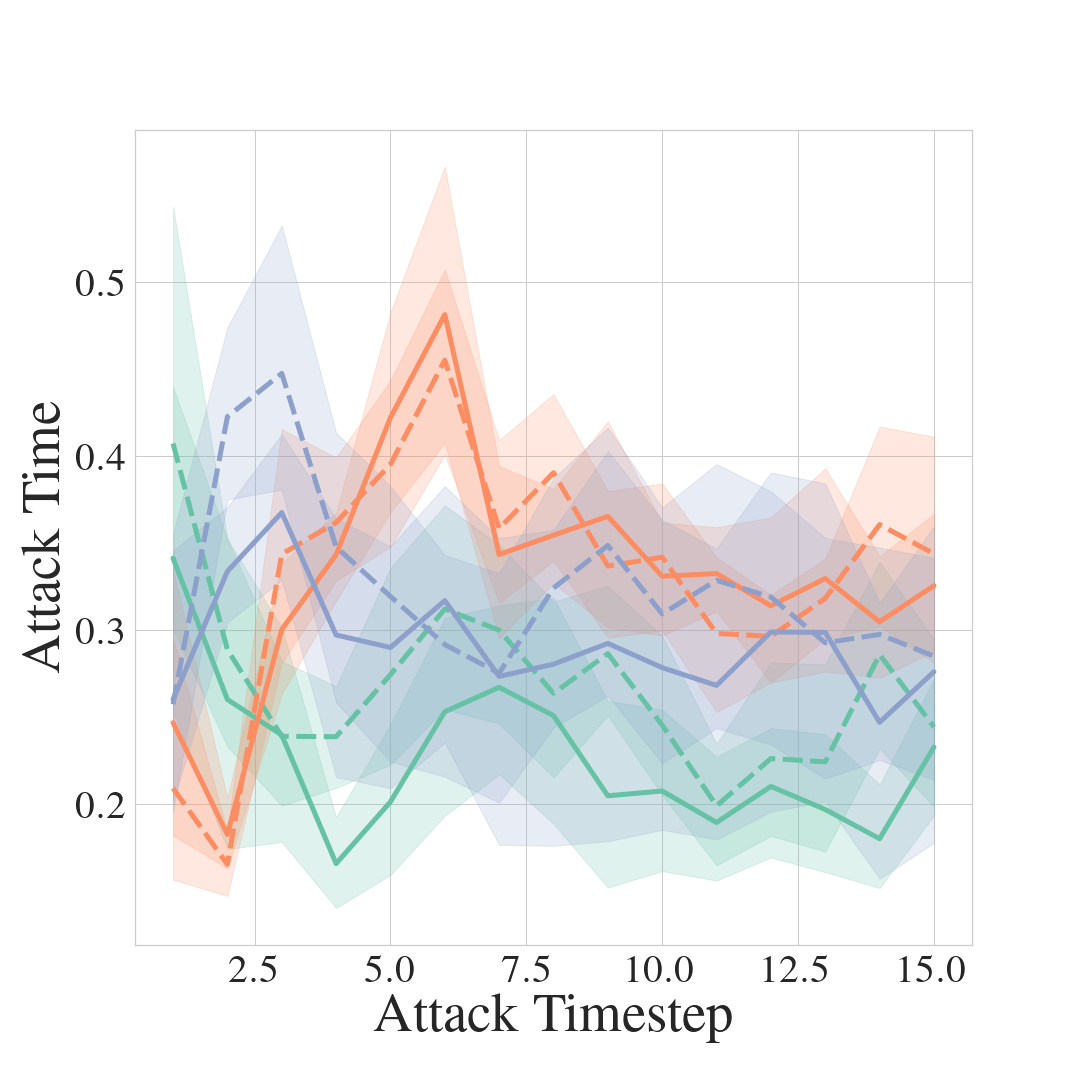}} \\
        \end{tabular}
        \caption{Training-Time statistics (a-c, e-h) and Test-Time performance (i-l) w.r.t. Accuracy (@Acc), Softmax Accuracy (@SoftAcc), Effort (@Effort), and Time (@Time) of $\gamma$DDPG with best fixed-discount (0.90) and dynamic discounts KLR and WD.  The dotted graphs in Test-Time plots (i-l) represent attacks on victims initialised with random numbers using different seeds.}
    \label{Fig:study2}%
    \end{figure*}

\section{Experiments}
\label{sec:Experiments}

Following a series of prior works\cite{rabinovich2010cultivating,Hang2,Hang3}, we experiment with attacks   
on a navigational agent (victim) whose objective is to find the shortest path to the goal state. The victim’s navigation environment can be discrete or continuous. 
The target behaviour is an alternate path to the goal state and is distinct from the path that the victim would learn in the absence of the attack. 
The current research aims to build an attacker that learns a high-accuracy, low-effort strategy to modify the stochastic dynamics of a discrete environment in order to push a training victim agent to learn a non-optimal target behaviour. This target behaviour must be un-adoptable in the default environment due to both, environment dynamics as well as non-optimality with respect to victim’s objective. In the following experiments, a path that is three times the length of the optimal path (shortest path under the original dynamics) is chosen as the non-optimal target behaviour.
This paper conducts 4 experimental studies. {\bf Study-1} is designed to demonstrate the capability of the discount factor to function as a means of bounding the lower-priority objective (minimise @Effort) while reducing the effect of uncertainty so as to aid in the optimisation of the primary objective (maximise @Acc) in a high-dimensional space and is carried out by training and testing $\gamma$DDPG with different fixed-discounts (Appendix 8). Herein, fixed-discount 0.90 is shown to achieve the best balance between @Acc and @Effort and is therefore chosen as the best fixed-discount w.r.t. the current setting. However, finding the best fixed-discount requires a grid-search and fixed-discounts cannot be used in settings where the victim task or victim environment changes with time, in a manner that the optimal discount factor of the attacker undergoes adaptation. These problems associated with fixed-discounts are solved with adaptive/dynamic discounts proposed in Section \ref{sec:Methodology}. {\bf Study-2} and {\bf Study-3} compare effort+accuracy-based and effort-based dynamic discounts with the best fixed-discount (0.90) while {\bf Study-4} compares the core contribution of this work i.e. $\gamma$DDPG with best dynamic discounts with the state-of-the-art baseline, TEPA \cite{Hang2}. Due to space constraints, studies 1 and 3 are moved to Appendix 8.

\textbf{Victim Environment \& Task:} In order to better align our contribution with prior works, we utilise the 3D Grid World \cite{rabinovich2010cultivating} to test and establish the quality of the proposed methodology. This environment simulates an uneven terrain on a $2D$ grid of cells. The unevenness corresponds to the $3^{rd}$ dimension of the grid and is due to the elevation/altitude associated with each grid cell. The relative elevation between two cells affects the transition probability between them. A change in this relative elevation thus changes how the environment responds to a navigating agent's actions. The navigating agent's task is to find the optimal path from the start cell to the goal cell and its state is its position inside the grid world. At each time step, the agent observes its state and takes one step in any of the four cardinal directions (N,S,E,W). The agent receives a reward of -1 for every action it takes until it reaches the goal state. A given agent training episode terminates once the agent reaches the goal state or maximum time has elapsed. This pushes the agent to find a shortest path to the goal cell. This environment also allows the presence of an additional agent, the elevation expert who can view the altitude of each grid cell and take a constrained action to modify it. The elevation expert's state space comprises of the grid cells' altitudes along with the navigational agent's behaviour, while its action space is a vector $[x^1,\dots, x^M]\in [-1.0,1.0]^M$, where $M$ is the total number of cells in the grid. In this work, the victim is a navigating agent while the attacker is the elevation expert.

\textbf{Performance Metrics:} The performance of the attacker is measured in terms of the accuracy (Attack Accuracy, abbrevaited @Acc) and strength (Attack SoftMax Accuracy, abbreviated @SoftAcc) with which the victim (unknowingly) adopts the target behaviour; the cumulative changes brought about in the victim environment by the attacker (Attacker Effort, or @Effort); and time taken to carry out the attack (Attack Time, or @Time). In more detail, @Acc measures the level of adoption of the target behaviour by the victim in the given environment. Given that $\pi_{u_i}$ is victim's probabilistic policy in the environment with dynamics $T_{u_i}$, $S^*$ is the set of target states, let $N^*$ be the number of target states, $a^*_n$ be the target action in target state $s_n$, and $f_a(s, \pi_{u_i})$ is an indicator function, triggered when the given policy assigns highest probability to the target action in a given state.

\begin{equation}
    f_a(s_n, \pi_{u_i}) = 
    \begin{cases}
        1 &  \pi_{u_i}(a^*_n|s_n) > \pi_{u_i}(a_n|s_n) ~ \forall ~ a_n \\
        0 & otherwise
    \end{cases}
\end{equation}

\begin{equation}
    \begin{aligned}
        \textrm{@Acc} = \frac{1}{N^*} \sum_{s_n \in S^*} f_a(s_n, \pi_{u_i})
    \end{aligned}
\end{equation}

@SoftAcc measures the strength with which the victim adopts the target behaviour and is computed as the probability assigned to the target path by the victim agent.

\begin{equation}
    \begin{aligned}
        \textrm{@SoftAcc}
        &= \frac{1}{N^*} \sum_{s_n \in S^*} \pi_{u_i}(a^*_n|s_n) ) 
    \end{aligned}
\end{equation} 

@Effort is the degree to which the attacker modifies the victim environment. Let $h^1_{u_i},\dots, h^M_{u_i}$ be the altitudes of the $M$ grid cells in victim environment with dynamics $T_{u_i}$.

\begin{equation}
    \begin{aligned}
        \textrm{@Effort} = \frac{1}{M} \sum_{m = 1}^{M}  \mid ~ h^m_{u_i} - h^m_{u_{i-1}} \mid
    \end{aligned}
\end{equation}

@Time is the computation time taken by the attacker to carry out an attack action. @Time includes an observation period wherein the attacker observes the victim as it trains in the attacked/poisoned/modified victim environment. 

\textbf{Graphs:} The $\gamma$-variant of DDPG, that we introduce in this work, allows sophisticated use of dynamic discounting. We contribute several dynamic discounting factors to boost an environment poisoning attack. To further underline their impact, we use the {\em tuned} performance of the SotA algorithm, TEPA. %
The training-time statistics of the various models are demonstrated via KDEs (histograms smoothed using kernel density estimation) as well as line graphs that depict these statistics across training time. KDEs represent the overall frequency of the mean value of each metric during the complete training period. For line graphs, we show a sliding-window maximum of metric means. Metric mean is computed for every attacker training episode, and the sliding window has a length of 75 such episodes. 
The attacker training episode is a 15-step sequential attack on a randomly initialised victim wherein attack step 0 corresponds to the original environment with default dynamics, and the episode ends when the victim has adopted the target behaviour with 1.0 @Acc or max attack steps (15) have elapsed. Each KDE and line graph represents four training runs with different seeds, except for Study-4, wherein a single training run is presented due to exorbitant runtime requirements of the baseline model. We plot the negative of mean @Effort and mean @Time values, so as to standardise all KDE graphs to "further right is better" and line graphs to "higher is better". 

The test-time performance graphs showcase the performance of the best attack strategy found by each attack model. The best attack strategy is the attack strategy that acquires the highest mean attack accuracy (@Acc) during the given model's training. The amount of changes made to the environment (@Effort), and the strength of target behaviour adoption (@SoftAcc) and time taken to execute the attack (@Time), are the secondary and the tertiary features of the attack. The use of dynamic discounting in $\gamma$DDPG allows us to tradeoff between them, and we plot them to show a complete performance picture.
The proficiency of each 
attack strategy is visualised by plotting the four 
performance metrics across each 15-step sequential attack carried out using the given strategy (plots i-l in each Figure). @Acc and @SoftAcc are measured along the victim timescale to observe how the accuracies change (and thereby understand how the victim behaves) in-between attack actions. @Effort and @Time on the other hand can only be measured corresponding to each attack action and hence are measured along the attacker timescale. Due to this difference, accuracy plots begin from attack step 0 while effort and time plots begin from attack step 1.

It is important that during attack training, the victim agent is (re)initialised using a single random pattern. 
On the other hand, during the test-time evaluation of the attack strategy, {\em multiple} victim initialisation patterns are used. Test-time evaluation, therefore, challenges the generalisability of the best training-stage attack strategies to differently initialised victim agents.
    

\textbf{Baseline:} Out of the two state-of-the-art constructive, environment-poisoning, black-box TTAs: TEPA \cite{Hang2} and DBB-EPA \cite{Hang3}; the former is chosen as the baseline for this research as like the current research, TEPA \cite{Hang2} develops attacks on known victim environments while DBB-EPA \cite{Hang3} develops attacks on unknown victim environments. TEPA is an auto-encoder-based model that is shown capable of pushing a victim agent in white-box and proxy-black-box adversarial settings to adopt a target behaviour that is optimal with respect to the victim's objective, but un-adoptable with respect to the original environment dynamics. 

\textbf{Results:} Our results show that effort+accuracy-based dynamic discount computed using Wasserstein distance finds strategies that maximise accuracy (higher-priority objective) and minimise effort (lower-priority objective) within reasonable @Time. This can be seen in Study 2 (Figure \ref{Fig:study2}) which compares KLR and WD adaptive discounts with the best fixed-discount (0.90) found via grid-search. Even though, the training-time statistics (plots a-c, e-h) of the 0.90 fixed-discount showcase better @Acc and @SoftAcc than both the dynamic discounts; during test-time (plots i-l), the generalisability of WD dynamic discount surpasses both KLR and best fixed-discount as it achieves the highest @Acc and @SoftAcc in second-best @Time with the least @Effort. 


\begin{figure*}[ht]%
    \begin{tabular}{cccc}
    \subfloat[@Acc KDE]{\includegraphics[width=0.2\linewidth,trim={0cm 0cm 11cm 0cm},clip]{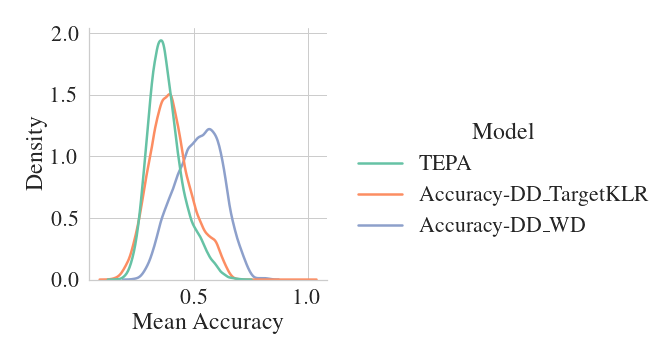}} &
    \subfloat[@SoftAcc KDE]{\includegraphics[width=0.2\linewidth,trim={0cm 0cm 11cm 0cm},clip]{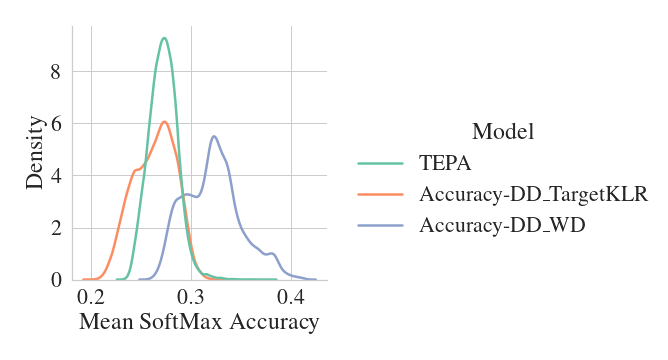}} &
    \subfloat[@Effort KDE]{\includegraphics[width=0.2\linewidth,trim={0cm 0cm 11cm 0cm},clip]{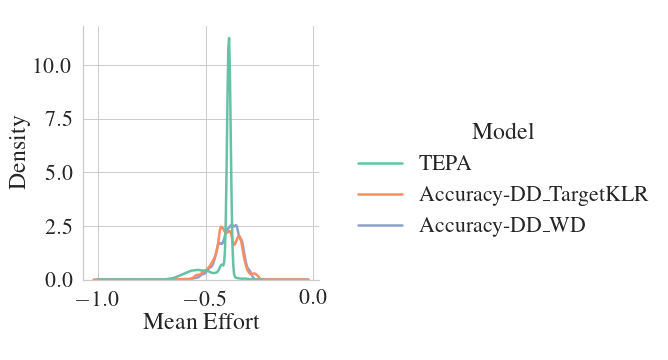}} &
    \subfloat[Legend]{\includegraphics[width=0.25\linewidth,trim={13cm 3cm 0cm 2cm},clip]{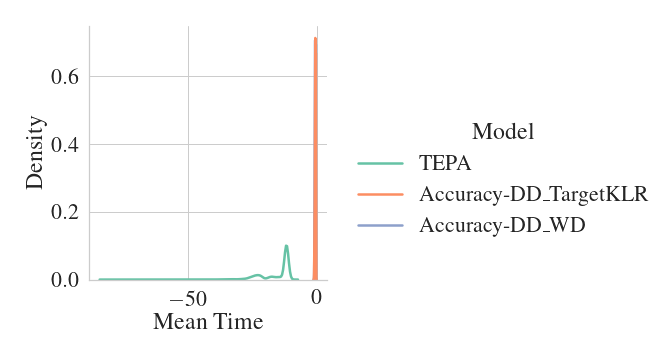}}\\
    
    \subfloat[@Acc Line Graph]{\includegraphics[width=0.24\linewidth,trim={2.5cm 0cm 2.5cm 2.5cm},clip]{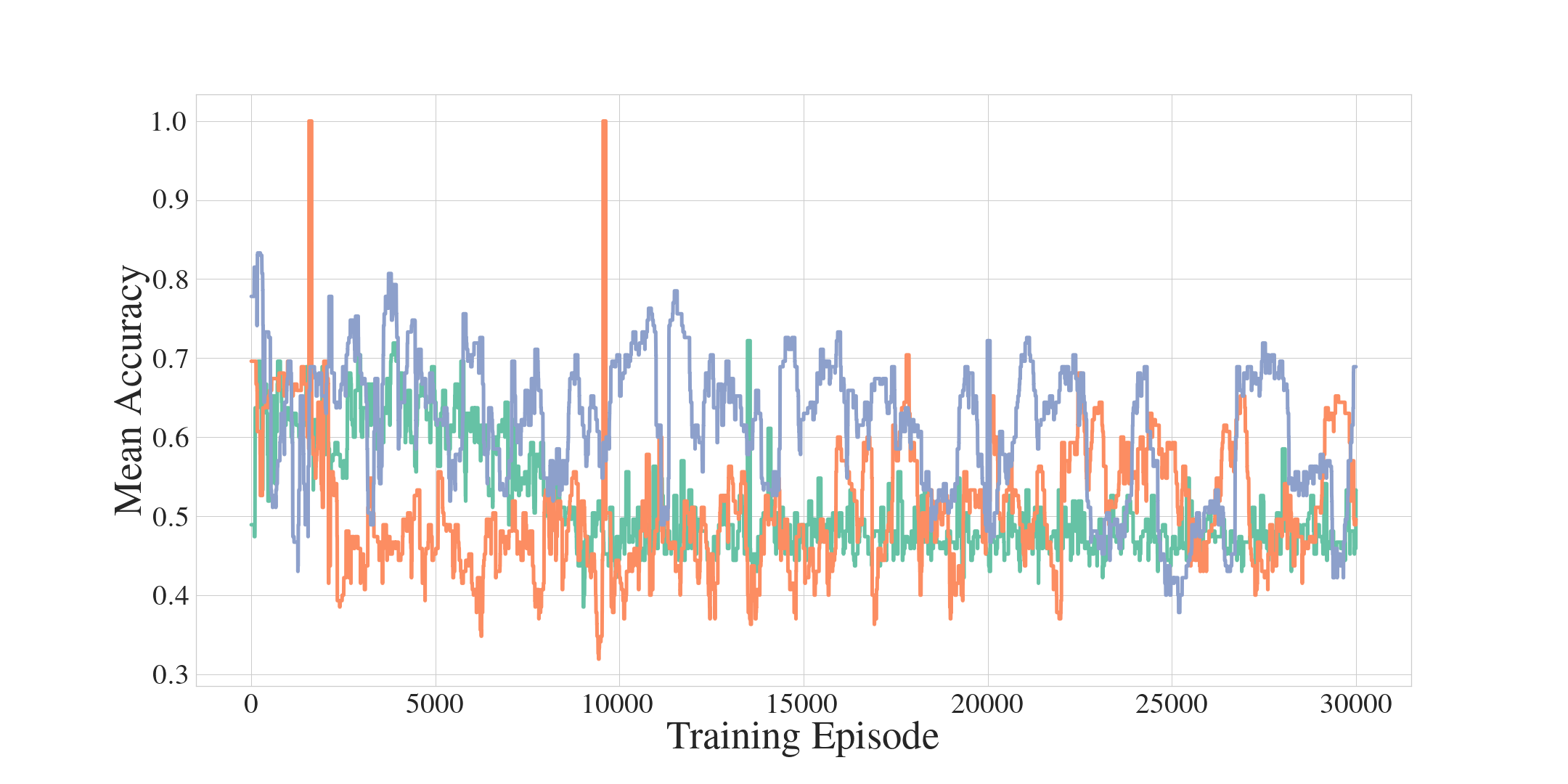}} &
    \subfloat[@SoftAcc Line Graph]{\includegraphics[width=0.24\linewidth,trim={2.5cm 0cm 2.5cm 2.5cm},clip]{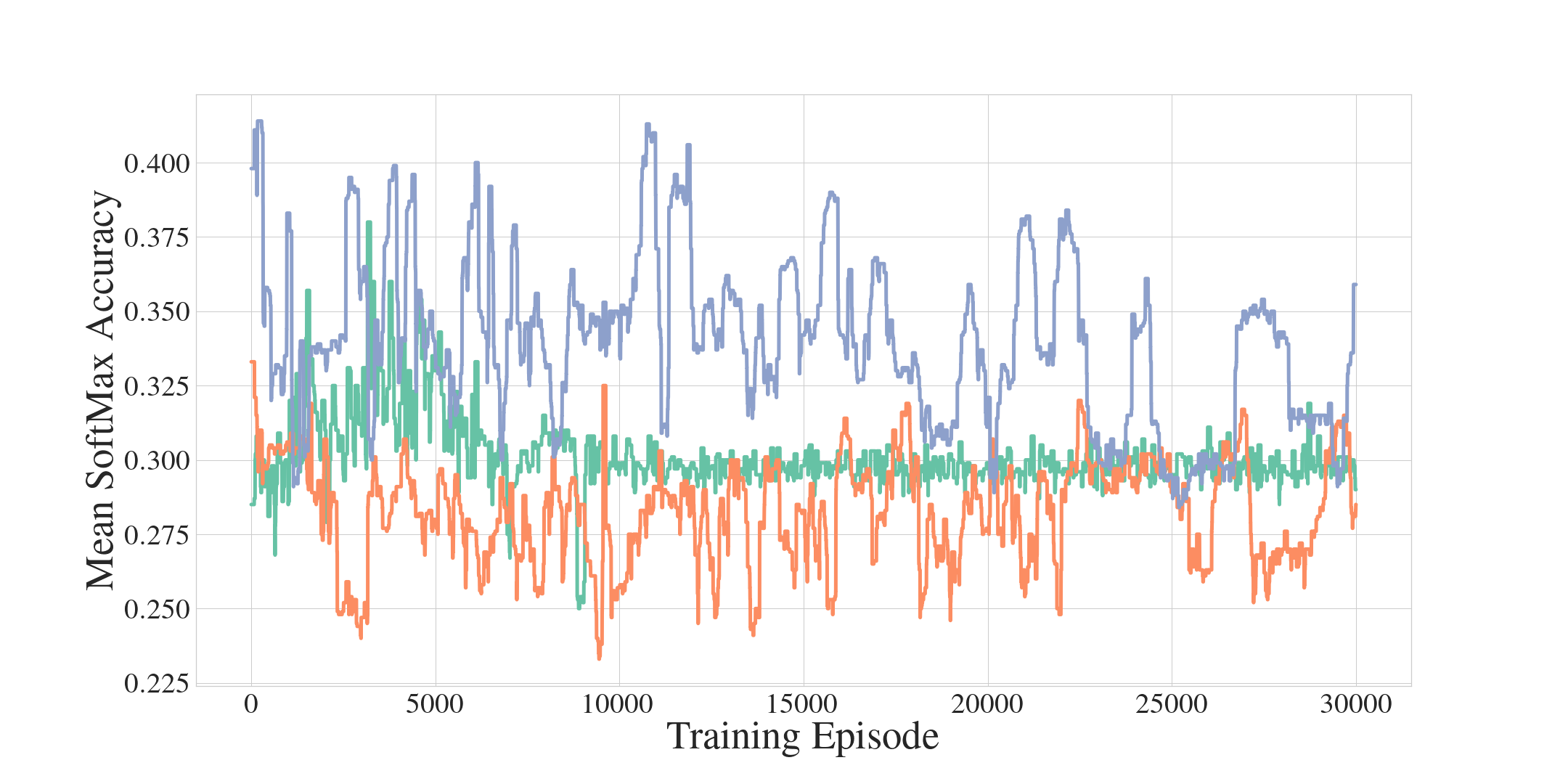}} &
    \subfloat[@Effort Line Graph]{\includegraphics[width=0.24\linewidth,trim={2.5cm 0cm 2.5cm 2.5cm},clip]{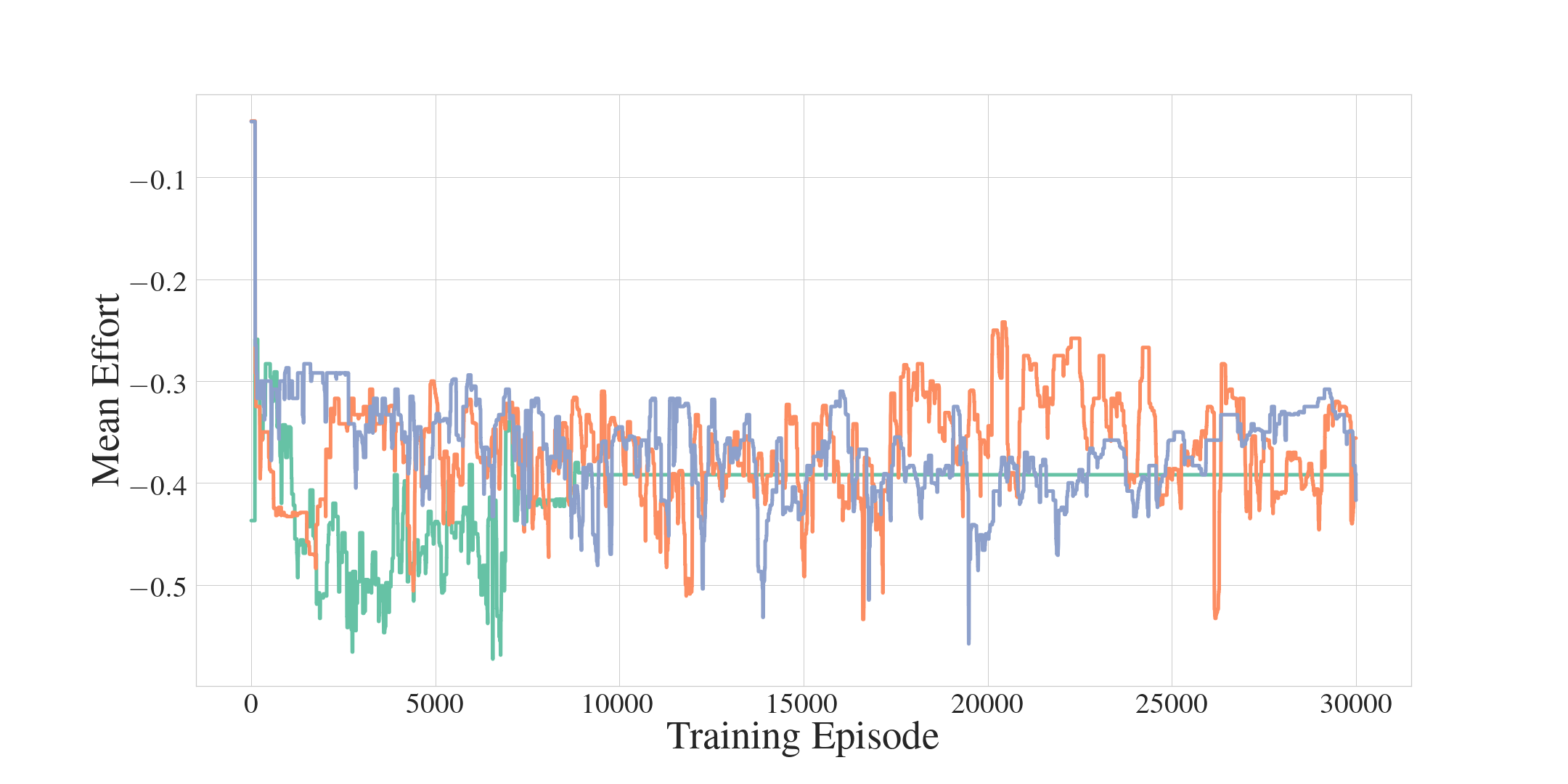}} &
    \subfloat[@Time Line Graph]{\includegraphics[width=0.24\linewidth,trim={2.5cm 0cm 2.5cm 2.5cm},clip]{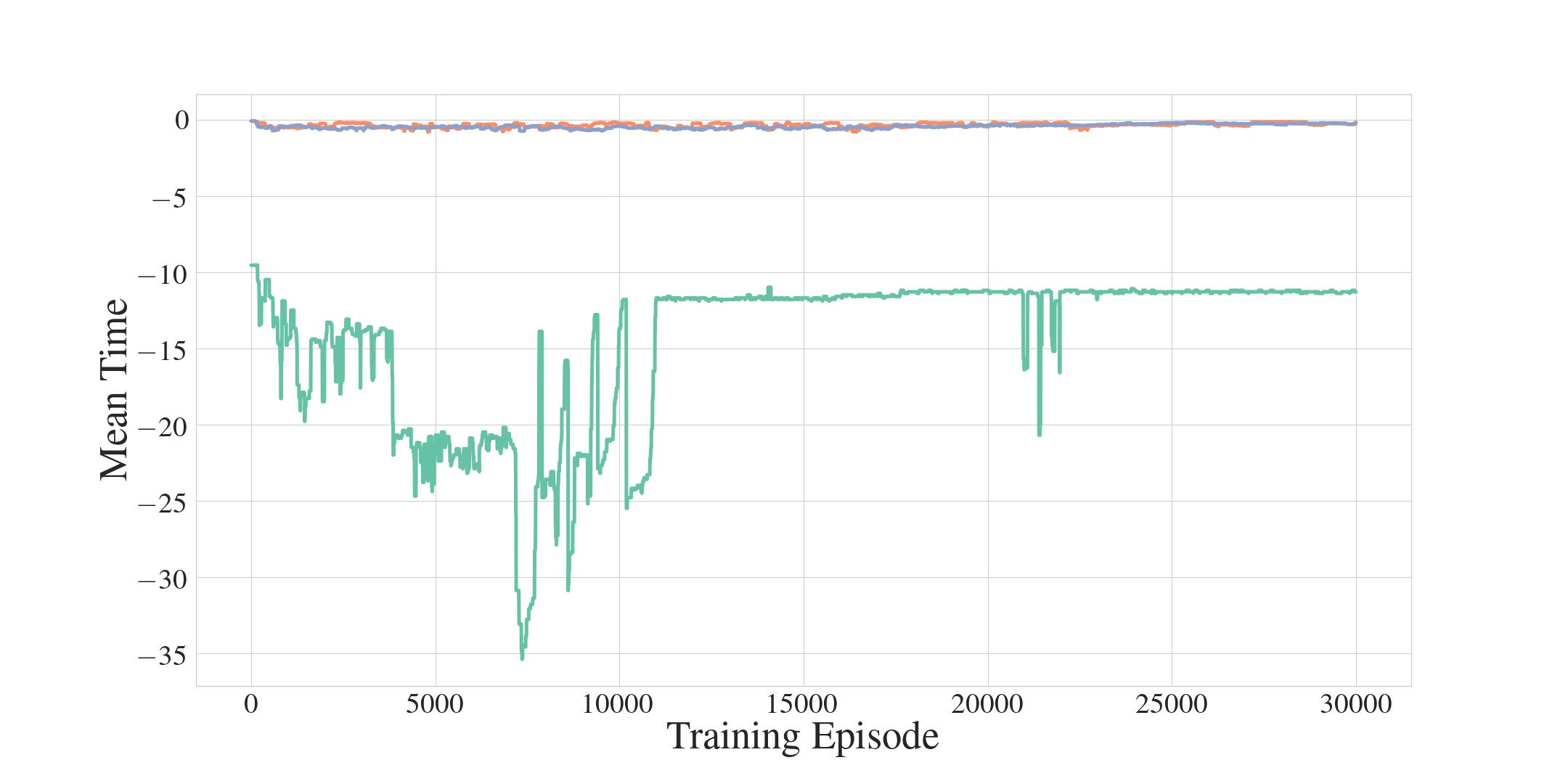}}\\

    \subfloat[Test-Time @Acc]{\includegraphics[width=0.2\linewidth,trim={0cm 0cm 5cm 0cm},clip]{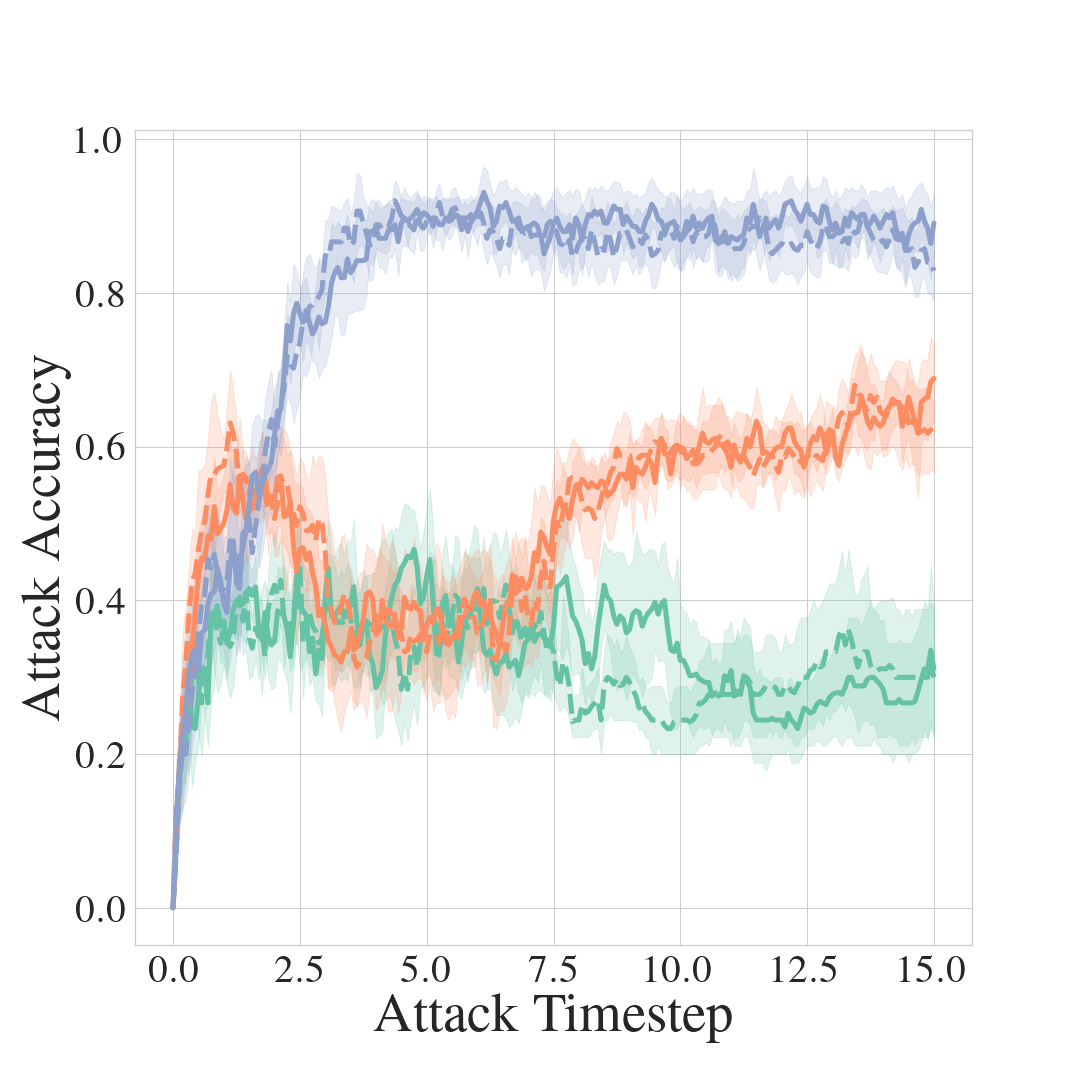}} &
    \subfloat[Test-Time @SoftAcc]{\includegraphics[width=0.2\linewidth,trim={0cm 0cm 5cm 0cm},clip]{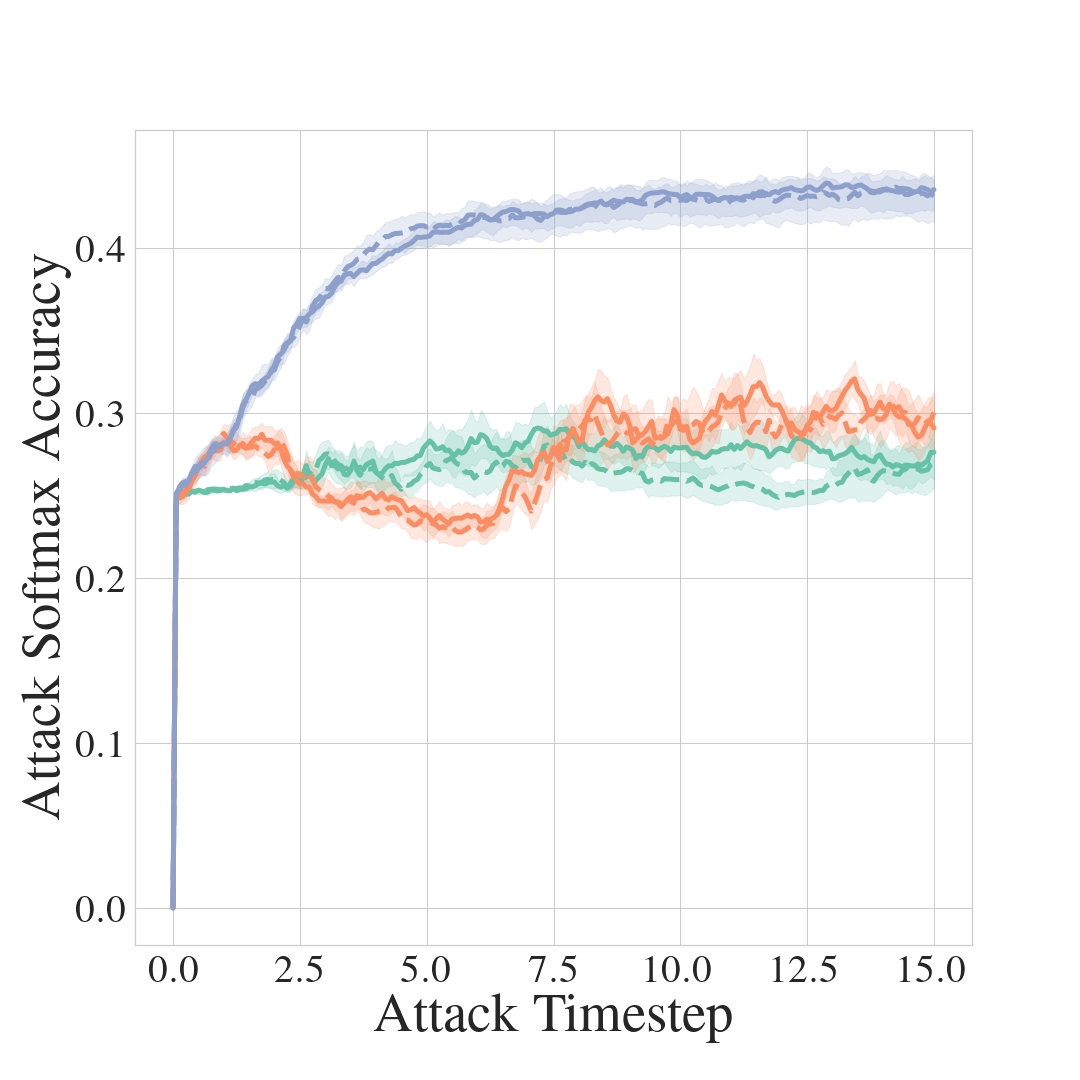}} &
    \subfloat[Test-Time @Effort]{\includegraphics[width=0.2\linewidth,trim={0cm 0cm 5cm 0cm},clip]{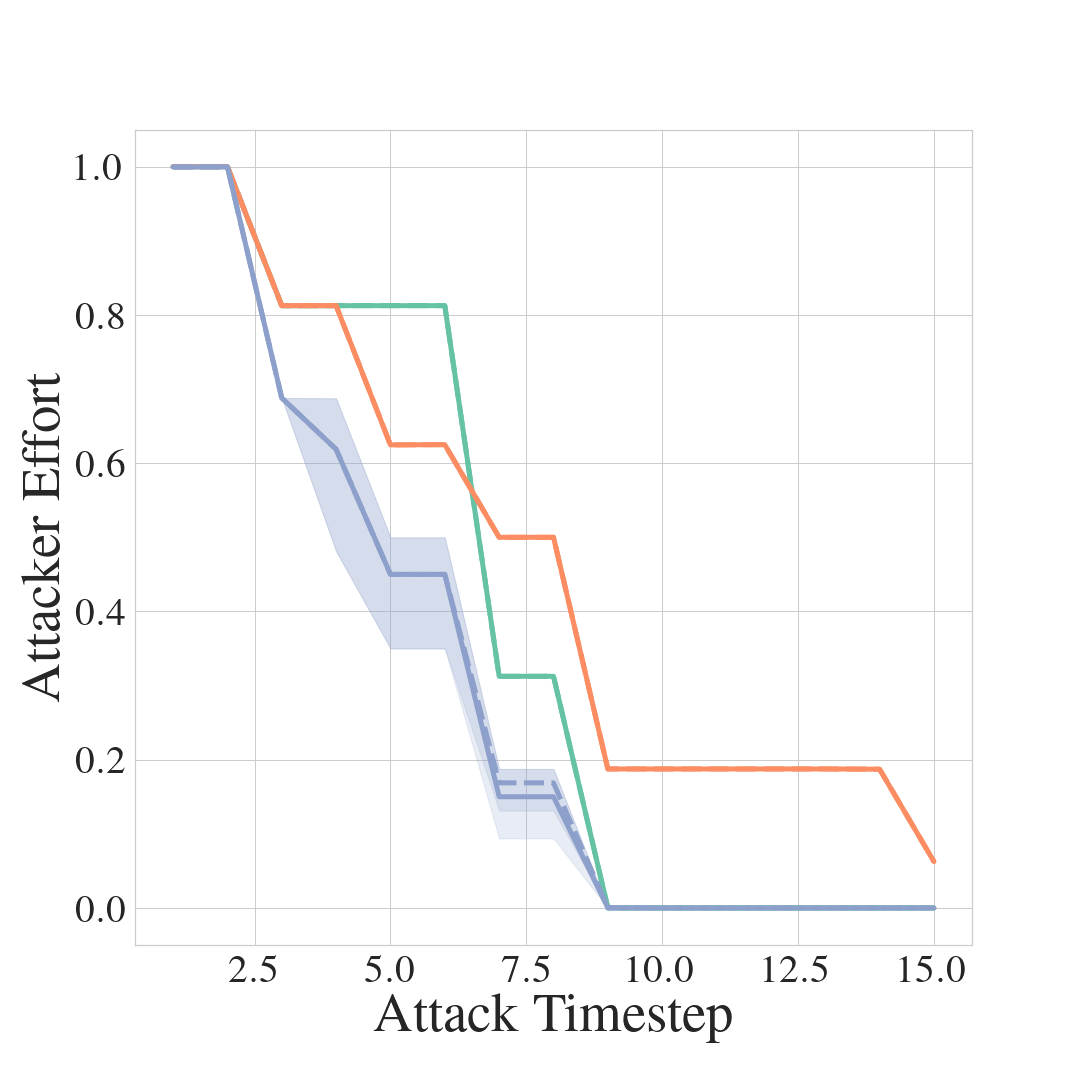}} &
    \subfloat[Test-Time @Time]
    {\includegraphics[width=0.2\linewidth,trim={0cm 0cm 5cm 0cm},clip]{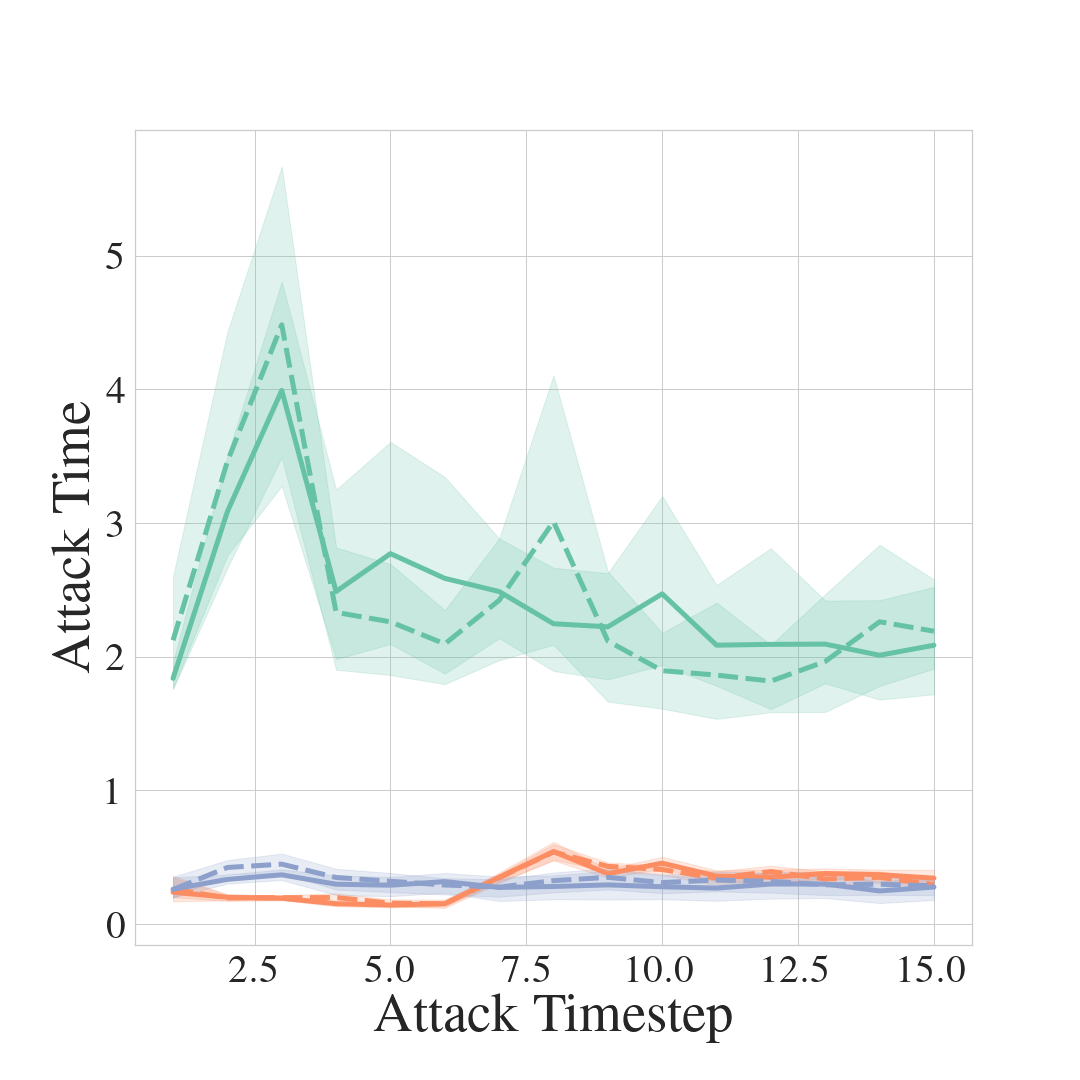}} \\
    \end{tabular}
    \caption{Training-Time statistics (a-c, e-h) and Test-Time performance (i-l) w.r.t. Accuracy (@Acc), Softmax Accuracy (@SoftAcc), Effort (@Effort), and Time (@Time) of baseline TEPA vs $\gamma$DDPG with dynamic discounts WD and TargetKLR. The dotted graphs in Test-Time plots (i-l) represent attacks on victims initialised with random numbers using different seeds.}
\label{Fig:study4}%
\end{figure*}

Furthermore, our results show that when TEPA, a SoTA baseline, is trained to enforce non-optimal target behaviour on a victim,
it gets stuck in a local optima, unable to exit it even after $\sim$20k training episodes. This can be seen, e.g., 
in Figure \ref{Fig:study4} that compares the best-performing effort+accuracy- and effort-based dynamic discounts, i.e. WD (Study 2) and TargetKLR (Study 3 in Appendix 8), with TEPA, 
highlighting the overall contribution of this work. Herein, plots e-h show that after approximately 11k training episodes, @Acc and @SoftAcc of TEPA begin to fluctuate within very small fixed ranges while @Effort and @Time plots become almost constant. In addition, the test-time plots (i-l) show that the best attack strategy found by TEPA performs worse than both WD and TargetKLR dynamic discounts w.r.t. @Acc, @SoftAcc and @Time.


\section{Conclusion and Future Work}
\label{sec:ConclusionLimitationsFutureWork}

This paper introduces a novel category of training-time environment-poisoning attacks wherein the attacker pushes the victim towards a non-optimal target behaviour. This target behaviour is un-adoptable in the original environment due to both, environment dynamics as well as non-optimality with respect to the victim’s objectives.
In order to make an attacker capable of carrying out such attacks, we introduce a novel reinforcement-learning algorithm titled $\gamma$DDPG that utilises an adaptive Bellman discount function to support dual-priority dual-objective optimisation in a partially observable setting. This dynamic discount bounds $\gamma$DDPG's search space conditioned on the accumulated modifications executed on the victim environment until the current attack step (and the difference between the current and target victim behaviour). The bounded search space, on one hand, bounds the lower priority objective (minimise Attacker Effort) and on the other hand, reduces uncertainty associated with the partially-observable environment and thereby aids in optimisation of the primary objective (maximise Attack Accuracy). The experiments conducted in this study show that Wasserstein distance based dynamic discount (WD) that respects the underlying geometry of the metric space and is insensitive to small differences in the probability distributions generalises better than the best fixed-discount found using a grid search (0.90) as well as the state-of-the-art baseline (TEPA). In fact, due to WD's exceptional generalisability to differently initialised victim agents, WD outperforms the best fixed-discount and TEPA attack models w.r.t. all four performance metrics.

In this work, however, the attacker approximates the victim's policy using the last action taken by the victim in each environment state. This mechanism can only be used for victim environments with discrete state spaces. Similarly, the current formulation of the dynamic discounts supports MDPs based on environments with discrete state and action spaces. Therefore, our next step entails extension of the proposed methodology to continuous environments. Furthermore, the proposed algorithm supports only dual-priority dual-objective optimisation. Future work constitutes expanding the developed methodology to multi-objective optimisation with more than 2 objectives and priority levels.    


\begin{acks}

This research was partly supported by the NTU SUG "Choice Manipulation and Security Games".

\end{acks}

\FloatBarrier


\bibliographystyle{plain}
\bibliography{references}

\FloatBarrier
\appendix
\section{(Expanded) Related Work}
\label{sec:ExpandedRelatedWork}

This research investigates the enforcement of non-optimal target behaviours on RL agents with a novel dual-priority dual-objective RL algorithm that uses the reward and a dynamic reward discounting function to encode and prioritise the two attack objectives. This section explores the literature related to this research including non-constant discounts, adaptive MDP and unsupervised environment design.

    \subsection{Adaptive Markov Decision Processes}
    \label{subsec:ExpandedRelatedWork/AdaptiveMDP}
    
    Predictive models when used to support/influence the system they model end up introducing themselves as a variable into the system \cite{perdomo2020performative,mendler2020stochastic,miller2021outside}. Such predictions are termed performative in nature as they can potentially modify the target distribution that they aim to predict. Predictive models that do not take performativity into account experience a shift in the underlying data distribution, over time, and deal with this problem of moving targets by periodically retraining the model using new data. This section presents the literature related to Adaptive MDPs and discusses how the issues that persist in predictive models are managed in the current research.
    
    \cite{perdomo2020performative} introduces the notion of performative stability that exists at equilibrium points wherein the model predicts the future that will manifest when the system will act on the model's predictions. In other words, equilibrium points are where the model becomes optimal for the distribution it induces. These equilibrium points are the stable points under retraining i.e. points at which the model becomes invariant to the retraining process. \cite{perdomo2020performative} proves conditions under which repeated risk minimisation converges to performative stability at a linear rate. Herein risk refers to the actual loss experienced by the decision-making process when it utilises the deployed predictive model. \cite{mendler2020stochastic} studies greedy vs lazy deploy during stochastic optimisation for performative prediction and finds that both deployment strategies are optimal under different settings of performative strength. \cite{brown2022performative} introduces a theoretical setting in which the shift in the target data distribution is a function of the predictive model and the current distribution; and studies conditions for convergence of repeated risk minimisation and one of its lazy variants to equilibrium. \cite{narang2022multiplayer} proposes a multi-player performative prediction setting wherein multiple predictive models compete against each other; and studies performatively stable as well as Nash equilibrium solutions in this setting. \cite{miller2021outside} states that performative stability does not ensure model optimality and strives to achieve the latter by directly optimising performative risk in a sample-efficient manner.
        
    \cite{mandal2022performative,bell2021reinforcement} study performative prediction in the sequential decision-making realm where the policy of a reinforcement-learning agent influences the underlying reward and transition dynamics of its environment. The agent's environment, modelled as an MDP, therefore adapts in response to the agent's behaviour. \cite{mandal2022performative} explores the deterministic adaptation of the MDP while \cite{bell2021reinforcement} investigates the setting wherein the transition and reward functions of the underlying MDP adapt non-deterministically to the agent's policy. The former work proves that when both the reward and transition functions change smoothly with respect to state-action occupancy, the agent converges to a stable equilibrium via repeated optimisation. On the other hand, the latter work shows that value-based RL agents cannot always converge to any optimal policy in non-deterministically adapting MDPs.
    
    This seemingly echoes the dual-MDP architecture proposed in the current research and bodes ill. However, in the setting proposed in the current research, the victim agents are unaware of being attacked and therefore the attacker's MDP does not actually undergo adaptation and the attacker cannot be considered as a "performative" agent. On the other hand, the victim (or victim population) MDP does undergo adaptation which is the reason behind the feasibility of the proposed training-time attack. Moreover, in the current research, this adaptation is made stable by ensuring that the attack is explicitly learned to {\em stably} influence the victim agent(s)).
    

    \subsection{Unsupervised Environment Design}
    \label{subsec:ExpandedRelatedWork/UED}
    
    Design of RL environments takes a lot of time and effort, is error-prone, and is infused with designer bias. Unsupervised Environment Design (UED) is a recent paradigm that aims to not only automate environment generation but also create environment distributions that are conducive to emergent complexity, robustness, and efficient transfer learning in RL agents. The UED paradigm shares certain features with the training-time environment-poisoning attack paradigm but the two domains are not equivalent to each other. In order to illustrate the differences between these two paradigms, this section presents literature related to the field of unsupervised environment design.
    
    Domain Randomisation (DR) \cite{tobin2017domain} creates a distribution of environments by assigning random values to the free parameters of the environment. Even though it exposes agents to a wide variety of environments, it does not create environments with complex structures and is not reactive to the capabilities of the learning agent. Minimax Adversarial Training (MAT) \cite{morimoto2005robust} makes environment distribution generation reactive to the learning agent’s capabilities by introducing an adversary that sequentially creates environments that minimize the learning agent's rewards. This worst-case tendency has however shown to create unsolvable environments that hinder the learning agent’s progress. PAIRED \cite{dennis2020emergent} does away with this worst-case tendency by introducing an additional "antagonist" agent which is smarter than the learning agent (protagonist). The adversary's objective is to maximise regret i.e. the difference between these two agents' rewards. This enables the generation of challenging but solvable environments. However, training the environment-creating adversary is challenging and is shown to produce weaker results than when main agents are trained on randomly-selected high-regret environments \cite{jiang2021replay}. This difficulty in training the adversary is due to sparse rewards and long-horizon credit assignment problems. Inspired by the teacher-student curriculum learning paradigm \cite{matiisen2019teacher}, Prioritised Level Replay (PLR) \cite{jiang2021prioritized}, and PLR+ \cite{jiang2021replay} do away with adversary-based mechanism and instead strategically sample the next environment by prioritising environments that have larger future learning potential. However, they cannot create new challenging environments with complex structures. This limits the frequency with which the learning agent can be exposed to complex structures and can thus hinder its progress, especially in high-dimensional complex environments. ACCEL \cite{parker2022evolving} does away with both, adversary-training-based and sampling-based methodologies’ drawbacks by utilising an evolutionary environment generator and a regret-based curator. The generator makes small modifications (mutations) to high-regret environments present inside the replay buffer. These modified environments are added to the replay buffer if they result in high regret. Regret thus functions as the fitness function for evolution and helps develop challenging environments for the learning agent.
        
    Training-time, environment-poisoning attacks can be compared to hypothetical negative-UED methodologies whose aim could be automatic design of environment distributions that result in the minimisation of the learning agent's (victim's) performance. However, a straightforward negative of existing UED methodologies is not equivalent to training-time, environment-poisoning attacks. First of all, adversary-based and evolution-based UED methodologies create challenging environments by aiming to minimise the learning agent's rewards, while sample-based UED methodologies sample environments with high learning potential. All these mechanisms serve to improve the learning agent's performance even in complex environments and tasks. A straightforward opposite of these approaches will not destroy the victim's performance or force the victim to learn different behaviours but will only train the victim agent to achieve good performance in simple environments and possibly average/poor performance in difficult environments. On the other hand, destructive environment-poisoning attacks destroy the victim's performance while constructive environment-poisoning attacks push the victim agent to adopt an attacker-desired target behaviour.
    
    Moreover, unlike UED, in training-time environment-poisoning attacks, the sequence of modifications is critical. While the victim begins training in its environment, the attacker begins to periodically modify the victim's environment. Since the attacker intends to push the victim towards a target behaviour that the victim has a low tendency of adopting by itself, in the original environment; the attacker must strategize modifications by carefully considering the current behaviours being learned by the victim. Moreover, the attack paradigm imposes constraints on the attacker in terms of the permissible magnitude of environment changes allowed at any single step. Any sizeable modification to the environment can therefore only be implemented through a sequence of small changes. The attack's sensitivity to current victim behaviours as well as magnitude constraints render the sequence of environment changes critical in the attack domain, which is not the case for the UED paradigm.

\section{Attacker's State and Action Spaces}
\label{sec:AttackerStateActionSpace}

The state space of the attacker constitutes the current behaviour of the victim RL agent and the current dynamics of the victim environment. This work, assumes the victim RL agent to be black-box and the victim environment to be white-box. The developed attack, therefore, does not access any internal mechanism of the victim RL agent for attack conditioning during training as well as testing of the attack. The victim's behaviour can, therefore, only be approximated through across-policy behaviour traces of the victim RL agent that the attacker can observe while the victim trains in its environment. Furthermore, this study works with the extreme case wherein the victim agent updates its internal policy after each interaction with its environment, and, therefore, each state-action pair of the victim's behaviour trace originates from a slightly different policy. On the other hand, as noted earlier, the victim environment is assumed to be white-box and therefore the attacker can directly obtain the transition dynamics of the victim environment at any time.
        
In order to appropriately approximate the victim's policy, behaviour traces across multiple epochs of the victim's training process would be required. However, conditioning an attack on such a large volume of data is extremely expensive and impractical. Therefore, this work preprocesses the victim's behaviour traces and stores the last observed victim action corresponding to each observed victim state and assigns a "no-action" symbol to unvisited states. A given victim agent's behaviour trace is denoted as $\tau_{u_{i-1}} = \{ s_1, a_1; s_2, a_2; ...; s_N, a_N\} \forall s_n \in S$, where $a_n$ is the latest action taken by the victim in state $n$ or the no-action symbol in case state $s_n$ was never visited by the victim, and $N$ is the total number of states in the victim environment. As $\tau_{u_{i-1}}$ contains the latest action / no-action symbol corresponding to all states, $\tau_{u_{i-1}}$'s size can still explode in high-dimensional environments. To combat this issue, this work learns a low-dimensional latent space, $\Phi$ of victim behaviours using an auto-encoder model. The auto-encoder model consists of an encoder $q_{e}$ that takes the victim's $\tau_{u_{i-1}}$ as input and outputs the corresponding latent behaviour $\phi_{u_{i-1}}$; and a decoder $q_{d}$ that takes two inputs, the latent behaviour $\phi_{u_{i-1}}$ and a victim environment state $s_n$, and outputs the probability with which the victim will take each available action in the given state $s_n$. The state space of the attacker can therefore be represented as $[T_{u_{i-1}}, \phi_{u_{i-1}}]$.

The attacker's action space is a vector of real numbers $[x^1, x^2, x^3, ..., x^M], x \in [-c,c]$ where $c$ is a constant and $M$ corresponds to the number of hyperparameters in the victim's environment that control the transition dynamics of the environment. When an attack action is executed, these real numbers are added to the existent values of the hyperparameters. Thereafter, the victim RL agent continues its training in this modified (poisoned) variation of the environment.

\section{(Expanded) $\boldsymbol{\gamma}$DDPG Algorithm}
\label{sec:Algorithm}

Appendix \ref{sec:Algorithm} presents $\gamma$DDPG algorithm in complete detail (Algorithm \ref{algo:DDPGgammaExpanded}) including the actor policy ($Q$) and target networks' ($Q', \sigma'$) update equations; and describes how victim behaviour $\tau$ is approximated in the given black-box setting with the aid of Algorithm \ref{algo:tau}.

$\gamma$DDPG is made capable of prioritising attack accuracy over attacker effort by maximising attack accuracy within an effort+accuracy-bounded search space. This search space is created at every attack step by adapting the attacker MDP's discount factor ($\gamma$) conditioned on the current attacker effort and attack accuracy. The adapting discount factor modifies $\gamma$DDPG's Bellman update to alter the level of importance that the algorithm accords to long-term rewards.
In further detail, when the attacker begins to train using $\gamma$DDPG and takes the first action to poison the victim's environment dynamics, its next-state is a low attacker-effort state due to existence of magnitude constraints on attack actions. However, the attack accuracy can be arbitrary. The discount function is proportional to attacker-effort and inversely proportional to attack-accuracy. Therefore, a decrease in effort and an increase in accuracy decreases the discount. Given that a small discount biases $\gamma$DDPG’s precedence to short-term rewards, a decrease in discount results in tightening of the search space of the attacker around the current state. On the other hand, an increase in effort and decrease in accuracy, increases the discount, which widens the search space of the attacker around the current state. Hence, after the first (constrained) attack action that results in a low effort state, in case the attack accuracy is high, the attacker's search space is tightened around the current state. As this state is a low-effort, high-accuracy (good) state, attacker should focus its exploration near this state in the given partially-observable setting where far-away states/rewards are less trustable. During this exploration, if the attacker visits better states (low-effort, high-accuracy), the search space gets tighter whereas whenever the attacker visits a bad state (high-effort, low-accuracy) its search space widens enabling the attacker to quickly move away from this bad state. In fact, the worse a given visited state, the wider the search space in the next time-step. However, the changes in the size of this search space are gradual due to the presence of magnitude constraints on the attack actions. Therefore, at each timestep, $\gamma$DDPG's adaptive discount factor creates a bounded effort+accuracy search space similar to the trust regions created by adaptive step sizes \cite{schulman2015trust}; and the attacker looks for the highest accuracy state within this space. The authors hypothesise that this bounded search space reduces the effects of uncertainty associated with future rewards in the partially observable state space and thereby aids in the optimisation process.

\begin{algorithm}[t]
\caption{$\boldsymbol{\gamma}$DDPG Algorithm (Expanded)}
\begin{algorithmic}[1]

    \STATE Randomly initialise critic $Q(x,u|\theta^Q)$ and actor $\sigma(x|\theta^\sigma)$ networks with weights $\theta^Q$ and $\theta^\sigma$
    \STATE Initialise target networks $Q'$ and $\sigma'$ with weights $\theta^{Q'} \leftarrow \theta^Q, \theta^{\sigma'} \leftarrow \theta^\sigma$
    \STATE Initialise replay buffer $R$
    
    \FOR{episode = 1, $M_a$}
        \STATE Initialise a random process $\chi$ for action exploration
        \STATE Receive initial observation state $x_0$
        
        \FOR{t = 1, $T_a$}
            \STATE Select action $u_{t} = \sigma(x_{t-1}|\theta^\sigma) + \chi_{t}$ according to the current policy and exploration noise
            \STATE Execute action $u_t$ to poison environment $T_{u_{t-1}}$
            \STATE Observe reward $r_{t}$  
            \STATE $x_t \leftarrow [T_{u_t}$, Auto\_Encoder( Algorithm \ref{algo:tau} ($T_{u_t}$) )]
            \STATE \textbf{Compute $\boldsymbol{\gamma_t}$ using Equation \ref{eq:discountFunction}}
            \STATE \textbf{Store transition $\boldsymbol{(x_{t-1}, u_t, r_t, x_t, \gamma_t)}$ in $\boldsymbol{R}$}
            \STATE \textbf{Sample a random minibatch of $\boldsymbol{N}$ transitions $\boldsymbol{(x_{i-1}, u_i, r_i, x_i, \gamma_i)}$ from $\boldsymbol{R}$}
            \STATE \textbf{Set $\boldsymbol{y_i = r_i + \gamma_i Q'(x_i, \sigma'(x_i|\theta^{\sigma'})|\theta^{Q'})}$}
            \STATE Update critic $\sigma$ by minimising the loss: $L = \frac{1}{N} \sum_{i}(y_i - Q(x_{i-1}, u_i|\theta^Q))^2$
            \STATE Update actor policy $Q$ using sampled policy gradient:
                \begin{align*}
                    \nabla_{\theta^\sigma} J \approx &\frac{1}{N} \sum_{i} \nabla_{u} Q(x,u|\theta^Q)\\
                    &\mid_{x = x_{i-1}, u = \sigma(x_{i-1})} \nabla_{\theta^\sigma} \sigma(x | \theta^\sigma) \mid_{x_{i-1}}
                \end{align*}
            \STATE Update the target networks $Q'$ and $\sigma'$:
                \begin{align*}
                    &\theta^{Q'} \leftarrow \rho \theta^Q + (1 - \rho)\theta^{Q'}\\
                    &\theta^{\sigma'} \leftarrow \rho \theta^\sigma + (1 - \rho)\theta^{\sigma'}
                \end{align*}
        \ENDFOR

    \ENDFOR
\end{algorithmic}
\label{algo:DDPGgammaExpanded}
\end{algorithm}

\section{Victim Environment Grid}
\label{sec:VictimEnvironmentGrid}

\begin{figure}[tbp]
  \centering
  \includegraphics[width=0.8\linewidth]{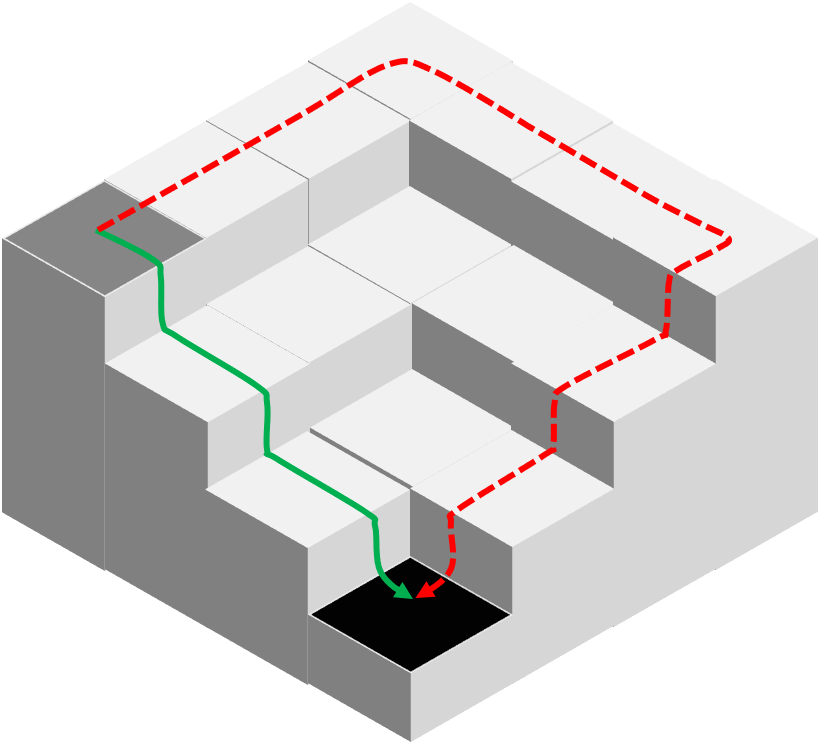}
  \caption{Default (Un-Attacked) Victim Environment}
  \label{fig:gridMp}
\end{figure}

The experiments conducted in this study employ the 4x4 3D grid depicted in Figure \ref{fig:gridMp} as the default (un-attacked) victim environment. In this grid visualisation, the start cell's top face is painted dark grey, and the goal cell's top face is painted black while the top faces of all other grid cells are light grey. Herein the green solid line is the shortest path between the start and the goal cell that is adopted by the victim RL agent in the absence of the attack. This path is adopted as not only is the path optimal with respect to the victim's objective (shortest path) but the path is also favoured by the transition dynamics of the environment. The red dotted line is the target path that the attacker wants the victim agent to adopt. The attacker forces adoption of this red dotted path by sequentially modifying the altitudes of the grid cells (thereby altering the transition dynamics of the environment) with constrained attack actions.

\section{Attack Strategy Selection}
\label{sec:AttackStrategySelection}

After each attacker training episode, the attack strategy employed in that episode is saved if it is better or equal to the best attack strategy found so far, with respect to last-timestep, mean or cumulative value of at least one strategy quality criterion. A given fixed-discount strategy's quality is approximated using 6 internal (KLR and WD based) and 5 external quality criteria, while that of a dynamic-discount strategy is approximated using 3 internal (KLR or WD based depending on the metric used to compute the dynamic/adaptive discount) and 5 external quality criteria. Herein criteria that are approximated by the attacker are referred to as internal while criteria computed by the external system for the purpose of training the attacker are termed external. The quality criteria are:

\begin{enumerate}
    \item KLR (Internal): KLR between vanilla-current MDP(current env * current victim behaviour) and perfect MDP(default env * target victim behaviour)
    \item TargetKLR (Internal): KLR between target-current MDP(current env * target victim behaviour) and perfect MDP(default env * target victim behaviour)
    \item DefaultKLR (Internal): KLR between default-current MDP(default env * current victim behaviour) and perfect MDP(default env * target victim behaviour)
    \item WD (Internal): Wasserstein Distance (WD) between vanilla-current MDP(current env * current victim behaviour) and perfect MDP(default env * target victim behaviour)
    \item TargetWD (Internal): WD between target-current MDP(current env * target victim behaviour) and perfect MDP(default env * target victim behaviour)
    \item DefaultWD (Internal): WD between default-current MDP(default env * current victim behaviour) and perfect MDP(default env * target victim behaviour)
    \item Attack Accuracy (External): same as Attack Accuracy (@Acc) performance metric
    \item Attack Softmax Accuracy (External): same as Attack SoftMax Accuracy (@SoftAcc) performance metric
    \item Attack Partial-Softmax Accuracy (External): unlike Attack SoftMax Accuracy where probability of attacker-desired actions in all attacker-desired states are added, here probability of attacker-desired actions in only those attacker-desired states are added where @Acc is 1.0 i.e. only probability of those attacker-desired actions are taken into account which are already assigned maximum probability by the victim. This quality criterion enables the attacker to identify (and thereby save) strategies that are capable of inducing strong adoption of target behaviour but were not able to achieve this in all target states in the given trial.
    \item Attacker Effort (External): same as Attacker Effort (@Effort) performance metric
    \item Attack Time (External): same as Attack Time (@Time) performance metric
\end{enumerate}

\section{Computing Infrastructure \& Hyperparameters}
\label{sec:Hyperparameters}

This section presents the hardware and software infrastructure used for running the experiments in Tables \ref{tbl:hardware} - \ref{tbl:software3} and lists the important hyperparameters relevant to the training of the victim and the attacker in Sections \ref{subsec:Hyperparameters/victim} and \ref{subsec:Hyperparameters/attacker} respectively. In the experiments conducted as part of this paper, each attack model is trained with four different seeds (0, 7, 16, 25) except TEPA which is only trained with seed 0 as it requires an exorbitant amount of time to train.

\begin{table}[tbp]
    \centering
    \caption{Hardware Infrastructure}
    \begin{tabulary}{\textwidth}{LCC}
        \hline
        \textbf{Name} & \textbf{Description} \\
        \hline
        CPU & AMD Ryzen Threadripper 3960X 24-Core \\
        Memory & 258GiB System Memory \\
        GPU & 4 NVIDIA RTX A6000 \\
        \hline
    \end{tabulary}
    \label{tbl:hardware}
\end{table} 

\begin{table}[tbp]
    \centering
    \caption{Software Infrastructure - I}
    \begin{tabulary}{\textwidth}{LCC}
        \hline
        \textbf{Name} & \textbf{Version} \\
        \hline
        Ubuntu & 20.04 \\
        \_libgcc\_mutex      &      0.1         \\
        \_openmp\_mutex      &      4.5         \\
        \_pytorch\_select    &      0.2         \\
        absl-py            &      0.13.0      \\
        aiohttp            &      3.7.0       \\
        alsa-lib           &      1.2.3       \\
        async-timeout      &      3.0.1       \\
        attrs              &      21.2.0      \\
        blas               &      1.0         \\
        blinker            &      1.4         \\
        brotlipy           &      0.7.0       \\
        bzip2              &      1.0.8       \\
        c-ares             &      1.17.2      \\
        ca-certificates    &      2021.10.8   \\
        cachetools         &      4.2.2       \\
        certifi            &      2021.10.8   \\
        cffi               &      1.14.6      \\
        chardet            &      3.0.4       \\
        click              &      8.0.1       \\
        cloudpickle        &      1.6.0       \\
        colorama           &      0.4.4       \\
        cryptography       &      3.4.7       \\
        cudatoolkit        &      11.3.1      \\
        cycler             &      0.10.0      \\
        cython             &      0.29.24     \\
        dataclasses        &      0.8         \\
        dbus               &      1.13.6      \\
        expat              &      2.4.1       \\
        ffmpeg             &      4.3.1       \\
        fontconfig         &      2.13.1      \\
        freetype           &      2.10.4      \\
        fsspec             &      2021.8.1    \\
        future             &      0.18.2      \\
        gettext            &      0.19.8.1    \\
        glib               &      2.68.4      \\
        glib-tools         &      2.68.4      \\
        gmp                &      6.2.1       \\
        gnutls             &      3.6.13      \\
        google-auth        &      1.35.0      \\
        google-auth-oauthlib  &   0.4.6       \\
        grpcio                &   1.38.1      \\
        gst-plugins-base      &   1.18.5      \\
        gstreamer             &   1.18.5      \\
        gym                   &   0.19.0      \\
        icu                   &   68.1        \\
        idna                  &   2.10        \\
        importlib-metadata    &   4.8.1       \\
        intel-openmp          &   2020.2      \\
        jbig                  &   2.1         \\
        joblib                &   0.17.0      \\
        jpeg                  &   9d          \\
        kiwisolver            &   1.3.2       \\
        krb5                  &   1.19.2      \\
        lame                  &   3.100       \\
        lcms2                 &   2.12        \\
        \hline
    \end{tabulary}
    \label{tbl:software1}
\end{table}

\begin{table}[tbp]
    \centering
    \caption{Software Infrastructure - II}
    \begin{tabulary}{\textwidth}{LCC}
        \hline
        \textbf{Name} & \textbf{Version} \\
        \hline
        ld\_impl\_linux-64      &   2.35.1      \\
        lerc                  &   2.2.1       \\
        libclang              &   11.1.0      \\
        libdeflate            &   1.7         \\
        libedit               &   3.1.20191231    \\
        libevent              &   2.1.10          \\
        libffi                &   3.3             \\
        libgcc-ng             &   11.2.0          \\
        libgfortran-ng        &   7.5.0           \\
        libgfortran4          &   7.5.0           \\
        libglib               &   2.68.4          \\
        libgomp               &   11.2.0          \\
        libiconv              &   1.16            \\
        libllvm11             &   11.1.0          \\
        libogg                &   1.3.4           \\
        libopus               &   1.3.1           \\
        libpng                &   1.6.37          \\
        libpq                 &   13.3            \\
        libprotobuf           &   3.18.0          \\
        libstdcxx-ng          &   11.2.0          \\
        libtiff               &   4.3.0           \\
        libuuid               &   2.32.1      \\
        libuv                 &   1.40.0      \\
        libvorbis             &   1.3.7       \\
        libwebp-base          &   1.2.1       \\
        libxcb                &   1.13        \\
        libxkbcommon          &   1.0.3       \\
        libxml2               &   2.9.12      \\
        lz4-c                 &   1.9.3       \\
        markdown              &   3.3.4       \\
        matplotlib            &   3.4.3       \\
        matplotlib-base       &   3.4.3       \\
        mkl                   &   2019.4      \\
        mkl-service           &   2.3.0       \\
        mkl\_fft               &   1.2.0       \\
        mkl\_random            &   1.1.0       \\
        mpi4py                &   3.1.2       \\
        multidict             &   5.1.0       \\
        mysql-common          &   8.0.25      \\
        mysql-libs            &   8.0.25      \\
        ncurses               &   6.2         \\
        nettle                &   3.6         \\
        ninja                 &   1.10.2      \\
        nspr                  &   4.30        \\
        nss                   &   3.69        \\
        numpy                 &   1.19.1      \\
        numpy-base            &   1.19.1      \\
        oauthlib              &   3.1.1       \\
        olefile               &   0.46        \\
        openh264              &   2.1.1       \\
        openjpeg              &   2.4.0       \\
        openssl               &   1.1.1o      \\
        packaging             &   21.0        \\
        pandas                &   1.1.3       \\
        pcre                  &   8.45        \\
        pillow                &   8.3.1       \\
        pip                   &   21.2.2      \\
        pot                   &   0.7.0       \\
        protobuf              &   3.18.0      \\
        \hline
    \end{tabulary}
    \label{tbl:software2}
\end{table}

\begin{table}[tbp]
    \centering
    \caption{Software Infrastructure - III}
    \begin{tabulary}{\textwidth}{LCC}
        \hline
        \textbf{Name} & \textbf{Version} \\
        \hline
        pthread-stubs         &   0.4         \\
        pyasn1                &   0.4.8       \\
        pyasn1-modules        &   0.2.7       \\
        pycparser             &   2.20        \\
        pydeprecate           &   0.3.1       \\
        pyglet                &   1.5.16      \\
        pyjwt                 &   2.1.0       \\
        pyopenssl             &   20.0.1      \\
        pyparsing             &   2.4.7       \\
        pyqt                  &   5.12.3      \\
        pyqt-impl             &   5.12.3      \\
        pyqt5-sip             &   4.19.18     \\
        pyqtchart             &   5.12        \\
        pyqtwebengine         &   5.12.1      \\
        pysocks               &   1.7.1       \\
        python                &   3.8.5       \\
        python-dateutil       &   2.8.2       \\
        python\_abi            &   3.8         \\
        pytorch               &   1.10.0      \\
        pytorch-lightning     &   1.5.2       \\
        pytorch-mutex         &   1.0         \\
        pytz                  &   2020.1      \\
        pyu2f                 &   0.1.5       \\
        pyyaml                &   5.4.1       \\
        qt                    &   5.12.9      \\
        readline              &   8.1         \\
        requests              &   2.25.1      \\
        requests-oauthlib     &   1.3.0       \\
        rsa                   &   4.7.2       \\
        scikit-learn          &   0.23.2      \\
        scipy                 &   1.6.2       \\
        seaborn               &   0.11.0      \\
        setuptools            &   52.0.0      \\
        six                   &   1.15.0      \\
        sqlite                &   3.36.0      \\
        tensorboard           &   2.6.0       \\
        tensorboard-data-server &  0.6.0      \\
        tensorboard-plugin-wit  & 1.8.0       \\
        tensorboardx            & 2.5         \\
        threadpoolctl           & 2.1.0       \\
        tk                      & 8.6.10      \\
        torchmetrics            & 0.6.0       \\
        torchvision             & 0.11.1      \\
        tornado                 & 6.1         \\
        tqdm                    & 4.62.2      \\
        typing\_extensions       & 3.10.0.0    \\
        urllib3                 & 1.26.6      \\
        werkzeug                & 2.0.1       \\
        wheel                   & 0.37.0      \\
        x264                    & 1!152.20180806   \\
        xorg-libxau             & 1.0.9       \\
        xorg-libxdmcp           & 1.1.3       \\
        xz                      & 5.2.5       \\
        yacs                    & 0.1.8       \\
        yaml                    & 0.2.5       \\
        yarl                    & 1.6.3       \\
        zipp                    & 3.5.0       \\
        zlib                    & 1.2.11      \\
        zstd                    & 1.5.0       \\
        \hline
    \end{tabulary}
    \label{tbl:software3}
\end{table}

    \subsection{Victim}
    \label{subsec:Hyperparameters/victim}

    Like prior works in constructive, training-time, environment-poisoning works; in this work, the victim utilises Q Learning with discount factor $\gamma_v = 0.90$, and learning rate $\alpha = 0.100$ for training. The attacker observes the victim's training for 80 episodes before modifying the victim environment with an attack action. These values were chosen as they enable the victim to converge to optimal behaviour (w.r.t its objectives) in the default (un-attacked) environment. Also, it is important to note that the state-of-the-art environment poisoning work, TEPA \cite{Hang2}, tested their attack on an $\epsilon$-Greedy victim that is heavily biased towards exploitation from the beginning of the training period. This work, on the other hand, tests the developed attack on more-realistic SoftMax victims that spend substantial time exploring the environment at the beginning of their training.

    \subsection{Attacker}
    \label{subsec:Hyperparameters/attacker}

    The attacker in the baseline (TEPA) trains using Deep Deterministic Policy Gradient (DDPG) algorithm while the attacker in the current work trains using the developed Gamma Deep Deterministic Policy Gradient ($\gamma$DDPG) algorithm. The baseline attacker algorithm (DDPG) uses a fixed discount factor ($\gamma = 0.95$), and a target network update rate of 0.005. The proposed attacker algorithm (DDPG), on the other hand, uses a dynamic discount function, and the same target network update rate of 0.005 as the baseline. The TEPA attacker collects data for 100 attack episodes before beginning to train while the attacker developed in this work utilises a shorter initial data collection period of 30 attack episodes. The policy function of both the attackers is a fully connected, feedforward neural network with specification: INPUT(21)-FC(400)-ReLU-FC(300)-ReLU-FC(16)-Tanh.
    
    TEPA attacker uses an auto-encoder that trains in parallel with the attacker in order to learn the latest victim behaviours with higher accuracy. On the other hand, the proposed $\gamma$DDPG attacker uses a pre-trained auto-encoder. Both encoders are fully connected, feedforward neural networks with specification: INPUT(12)-FC(36)-ReLU-FC(36)-ReLU-FC(5) that use a learning rate of 0.001. In order to test the TEPA attacker, its auto-encoder parameters are saved during training, after every 20 attack episodes. At the time of evaluation of a particular strategy, the auto-encoder parameters saved closest to the given strategy are used by the attacker.

\begin{figure*}[htbp]%
        \centering
        \includegraphics[width=0.25\linewidth,trim={0cm 0cm 0cm 0cm},clip]{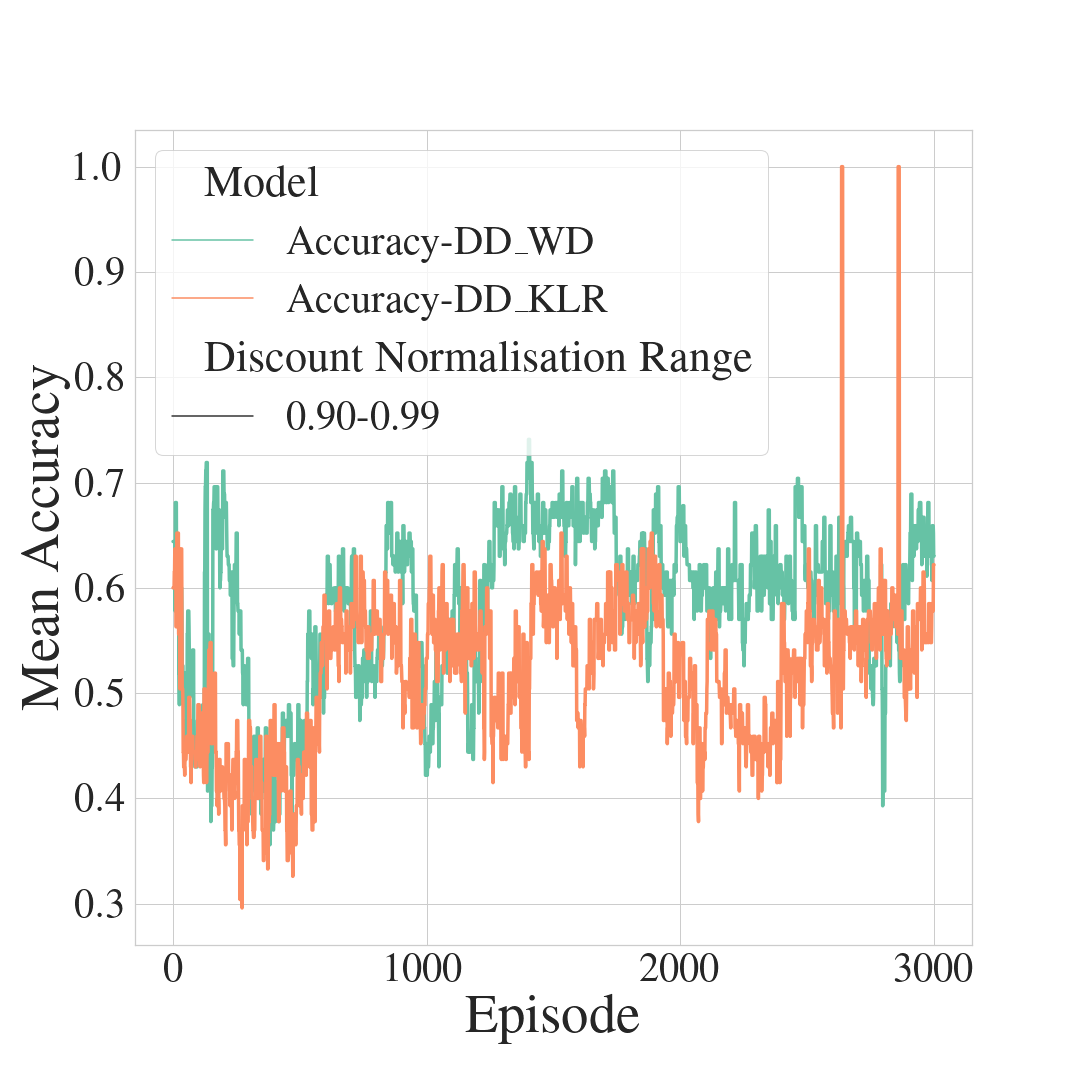}
        \includegraphics[width=0.25\linewidth,trim={0cm 0cm 0cm 0cm},clip]{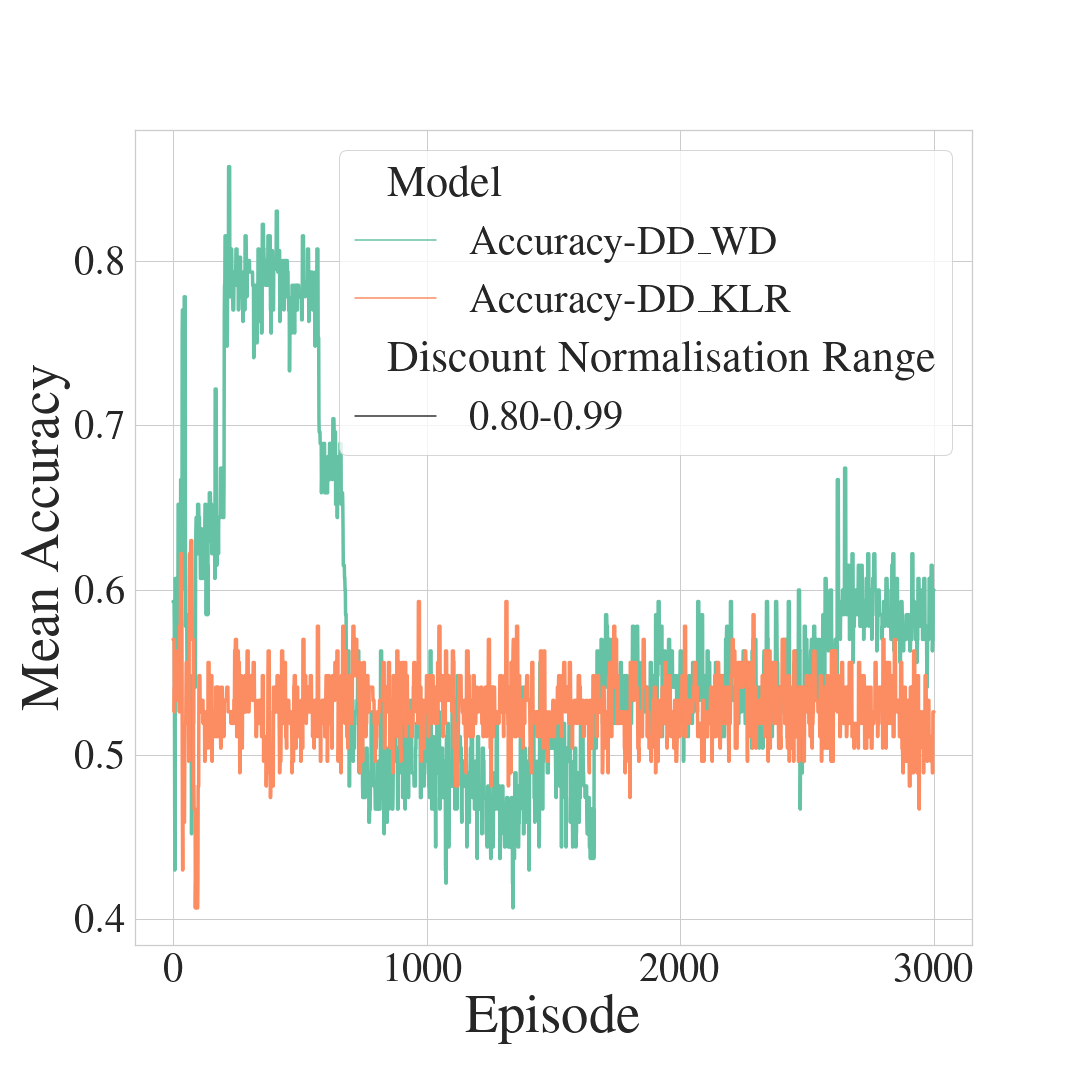}
        \includegraphics[width=0.25\linewidth,trim={0cm 0cm 0cm 0cm},clip]{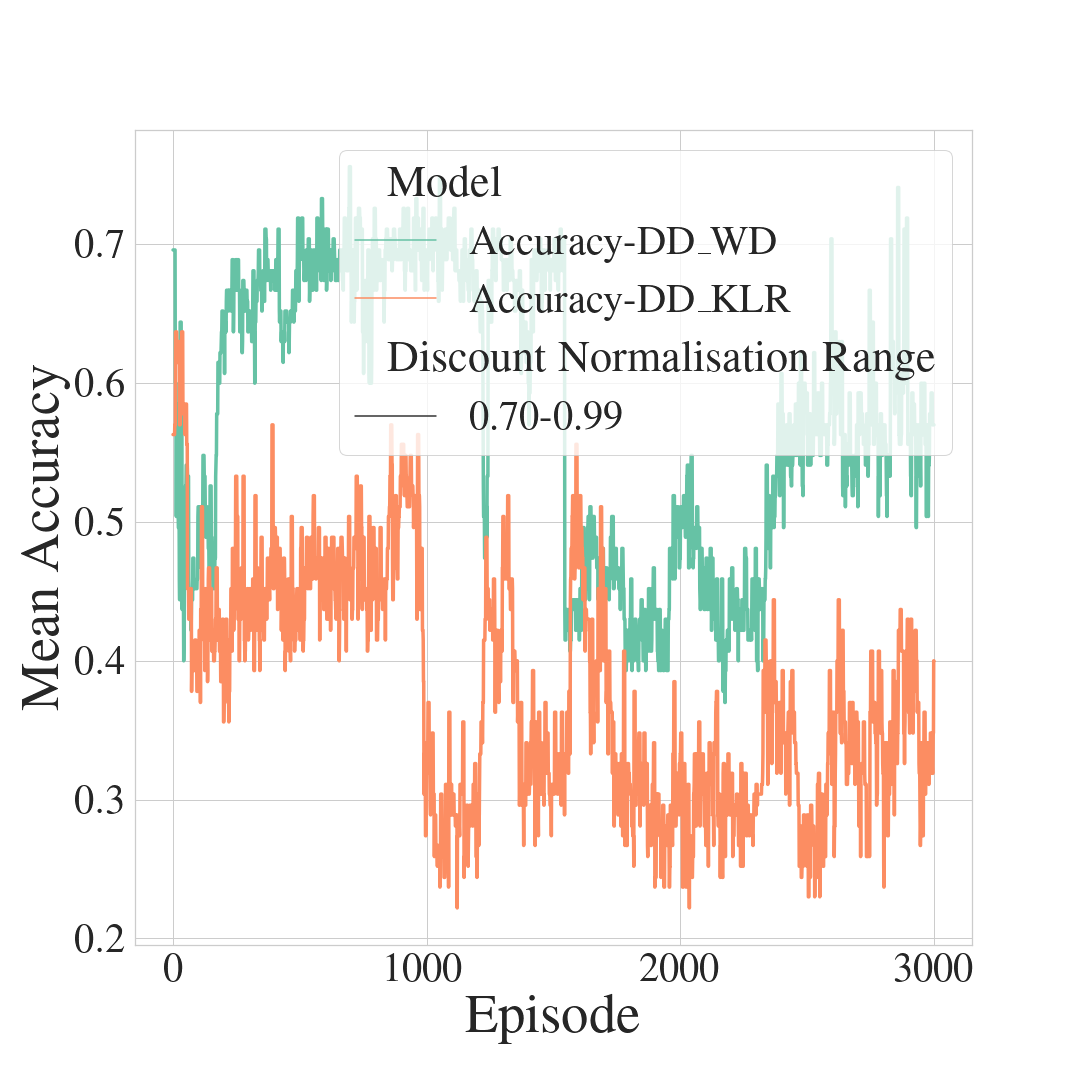}\\
        \includegraphics[width=0.25\linewidth,trim={0cm 0cm 0cm 0cm},clip]{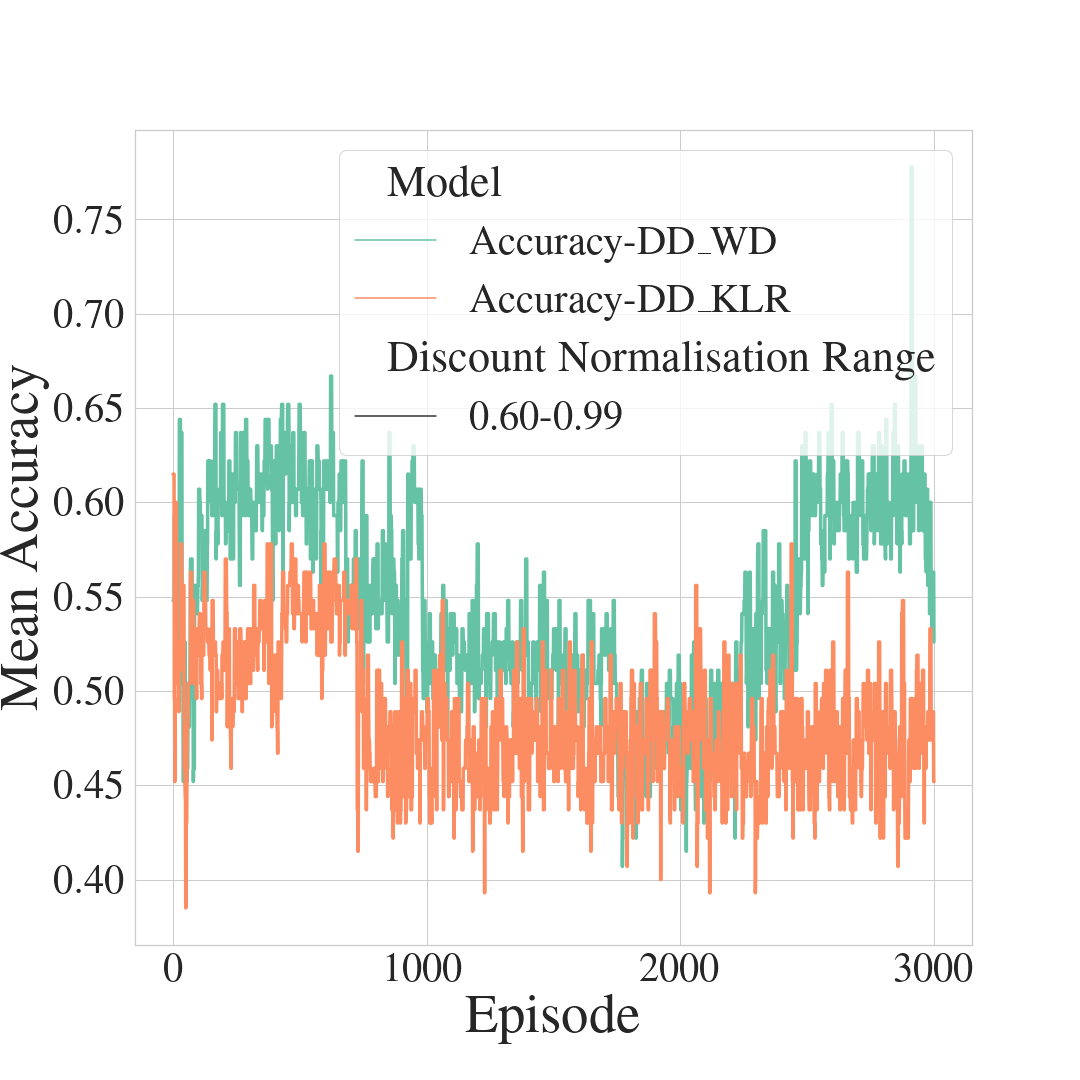}
        \includegraphics[width=0.25\linewidth,trim={0cm 0cm 0cm 0cm},clip]{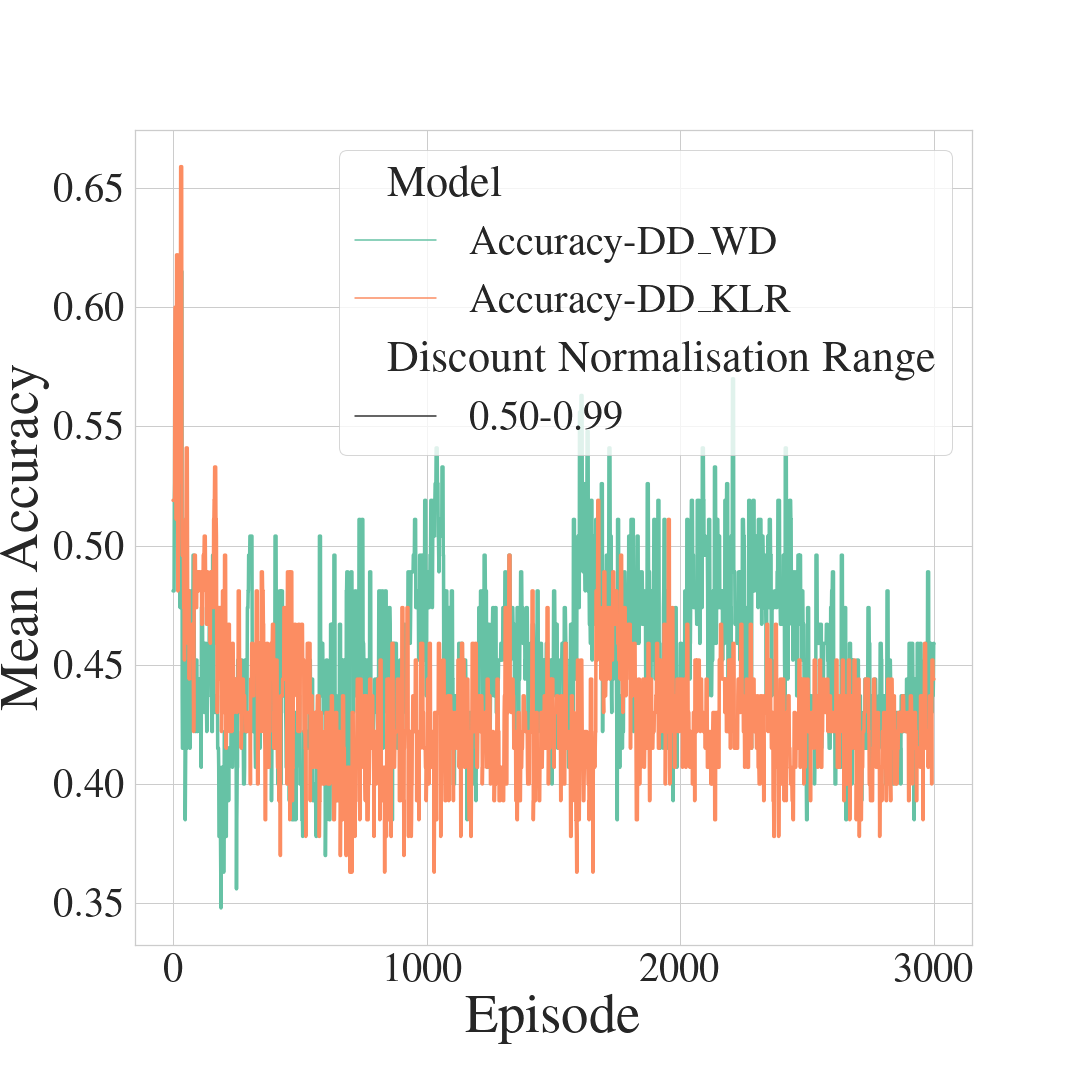}\\
        \caption{Training-Time Mean Attack Accuracy of KLR and WD dynamic discounts w.r.t. normalisation ranges [0.90-0.99], [0.80-0.99], [0.70-0.99], [0.60-0.99] and [0.50-0.99]}
        \label{fig:DDnormalisation}%
    \end{figure*}

    The range of KL divergence \citep{kullback1951information} as well as Wasserstein distance \citep{vaserstein1969markov} is $[0, \infty]$. Therefore the four adaptive discount functions (TargetKLR, TargetWD, KLR, WD) must undergo normalisation. In order to ensure that maximising accuracy is prioritised over minimising effort, each divergence/distance formulation is normalised into a range whose lower bound is greater than 0.5. A short single-seed experiment of 3k training episodes is conducted to test the effect of different normalisation ranges on performance of KLR and WD adaptive discounts. Figure \ref{fig:DDnormalisation} shows that WD achieves higher mean @Acc with high frequency in all ranges. In order to give KLR adaptive discounts a better chance, this work optimises both ranges to compare the best performers from each method. Therefore KLR-based formulations are normalised to [0.90, 0.99] while WD-based formulations are normalised to [0.80, 0.99].

\section{Extended Experiments}
\label{sec:ExtendedExperiments}

\begin{figure*}[htbp]%
    \begin{tabular}{cccc}
    \subfloat[Accuracy KDE]{\includegraphics[width=0.2\linewidth,trim={0cm 0cm 7cm 0cm},clip]{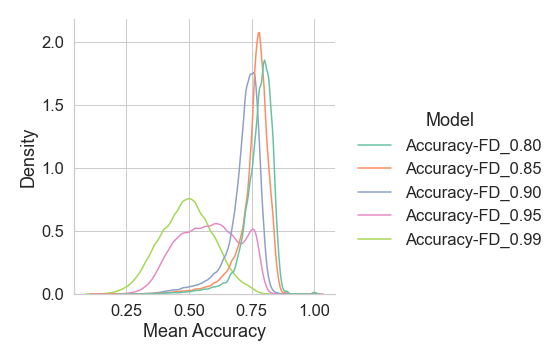}} &
    \subfloat[SoftMax Accuracy KDE]{\includegraphics[width=0.2\linewidth,trim={0cm 0cm 7cm 0cm},clip]{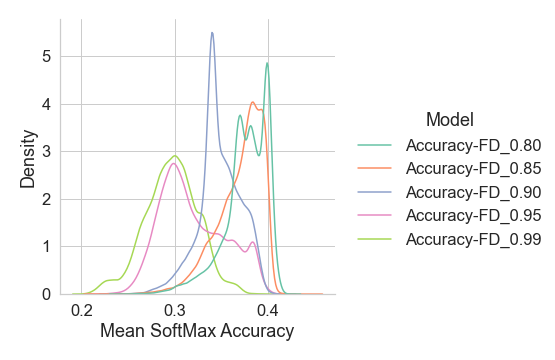}} &
    \subfloat[Effort KDE]{\includegraphics[width=0.2\linewidth,trim={0cm 0cm 7cm 0cm},clip]{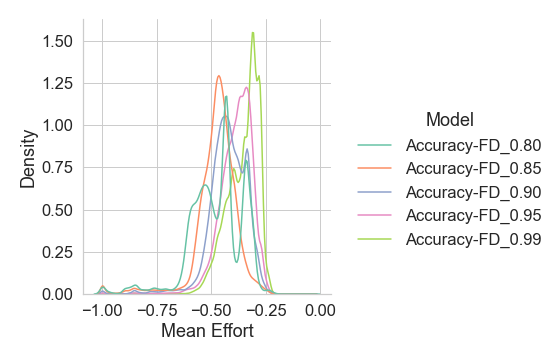}} &
    \subfloat[Legend]{\includegraphics[width=0.16\linewidth,trim={13cm 3cm 0cm 2cm},clip]{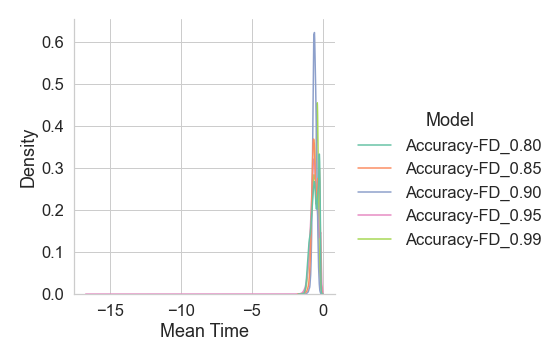}}\\

    \subfloat[Accuracy Line Graph]{\includegraphics[width=0.24\linewidth,trim={2.5cm 0cm 2.5cm 2.5cm},clip]{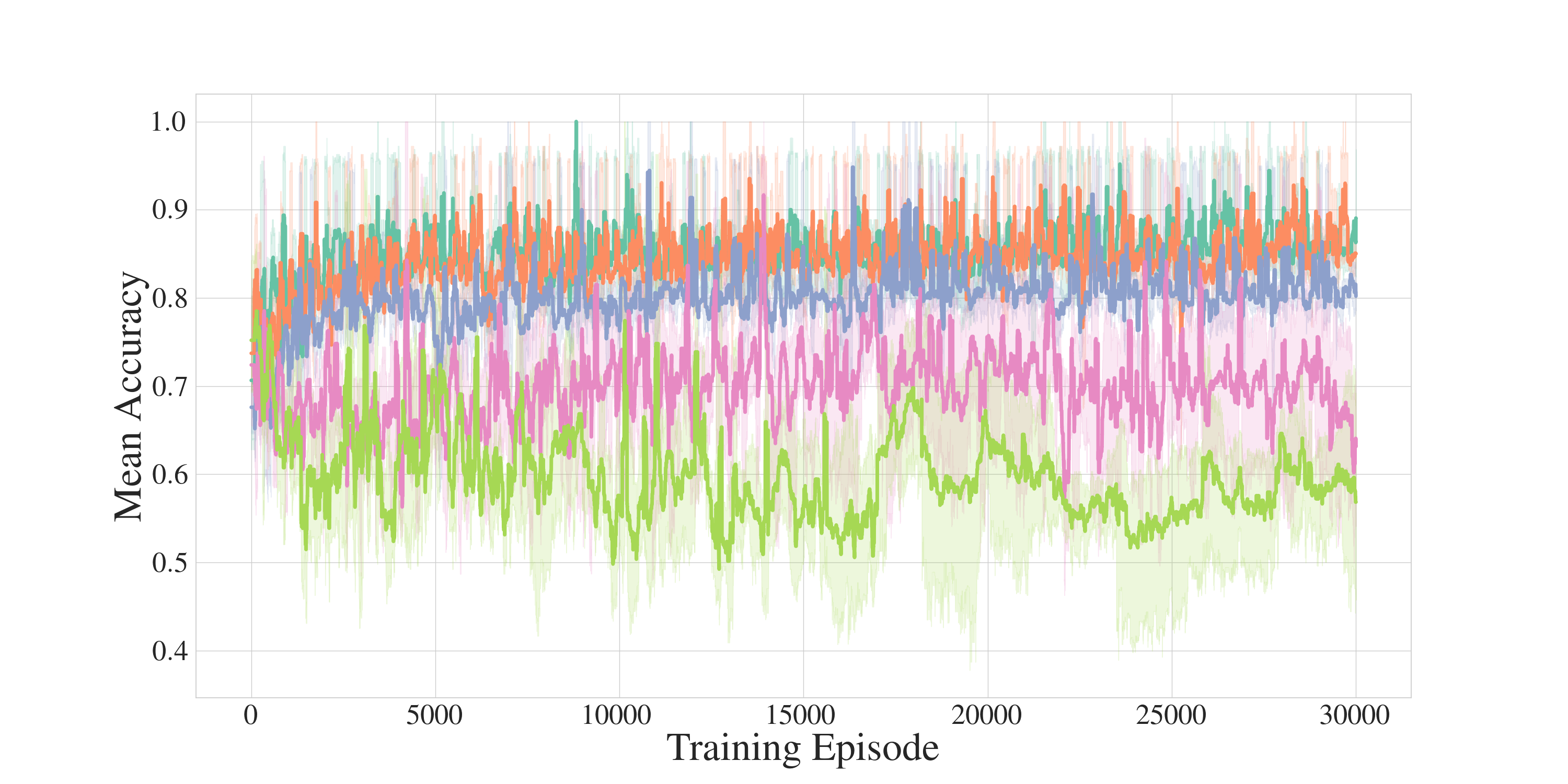}} &
    \subfloat[Softmax Accuracy Line Graph]{\includegraphics[width=0.24\linewidth,trim={2.5cm 0cm 2.5cm 2.5cm},clip]{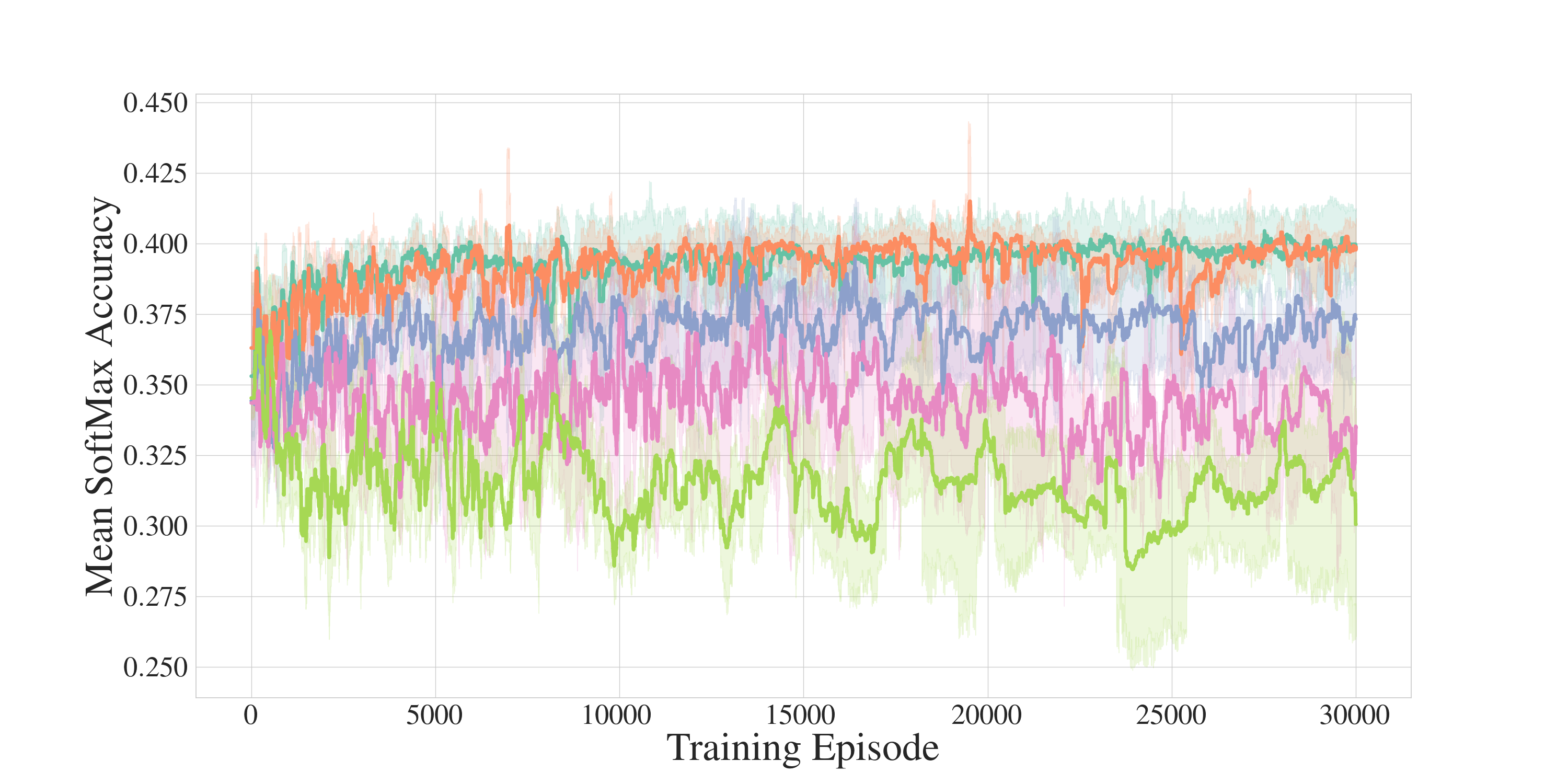}} &
    \subfloat[Effort Line Graph]{\includegraphics[width=0.24\linewidth,trim={2.5cm 0cm 2.5cm 2.5cm},clip]{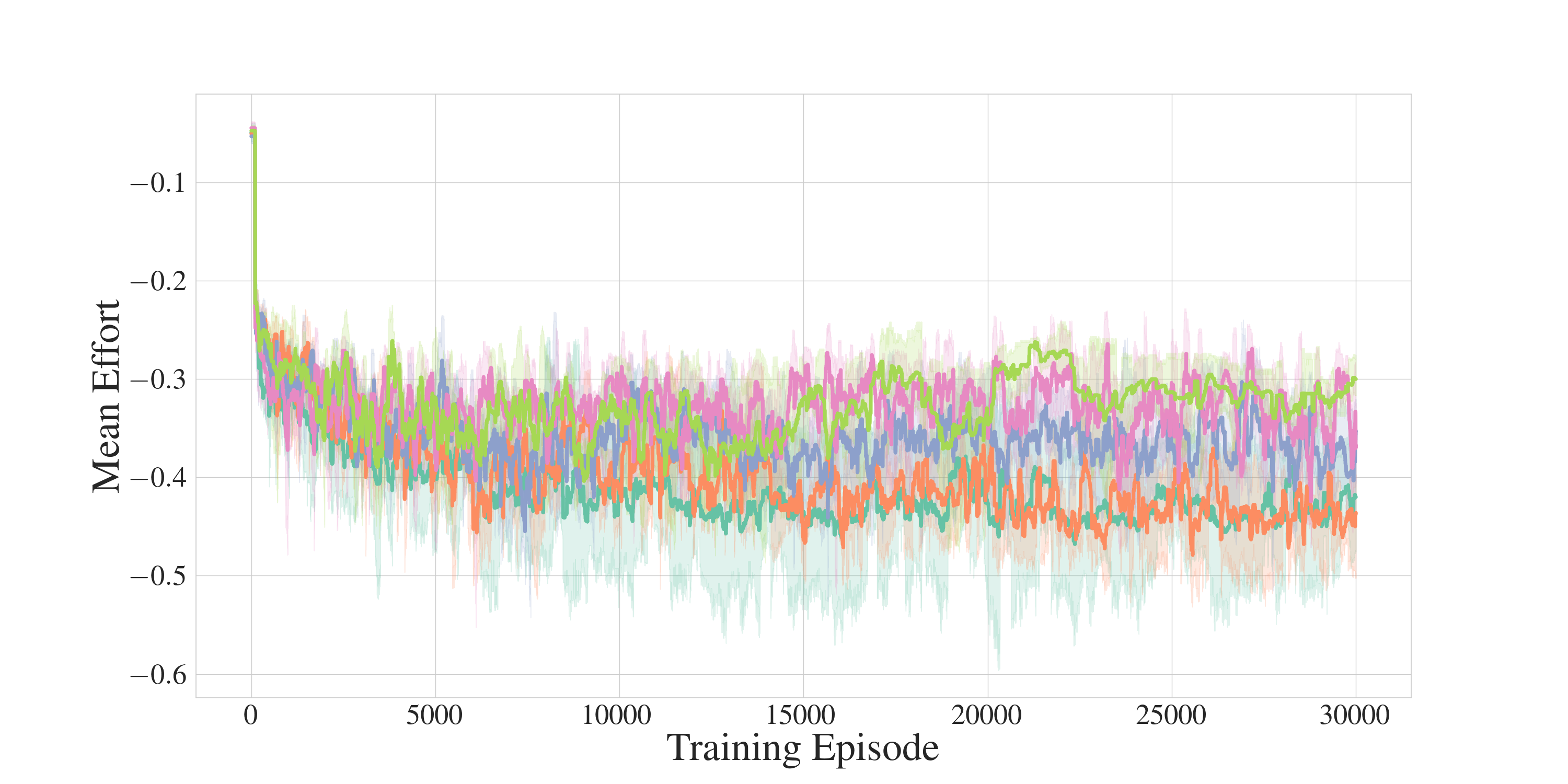}} &
    \subfloat[Time Line Graph]{\includegraphics[width=0.24\linewidth,trim={2.5cm 0cm 2.5cm 2.5cm},clip]{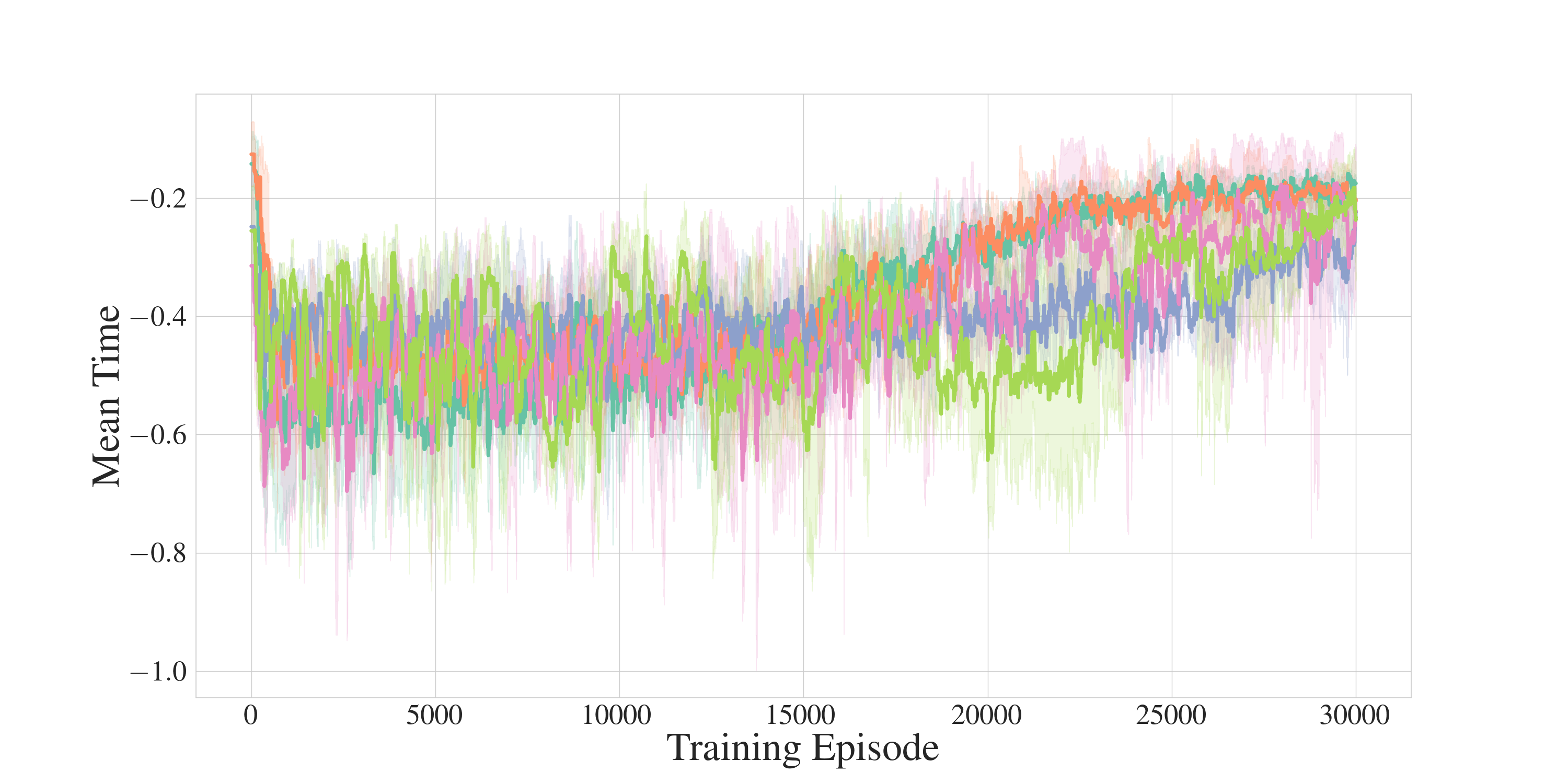}} \\

    \subfloat[Test-Time @Acc]{\includegraphics[width=0.2\linewidth,trim={0cm 0cm 5cm 0cm},clip]{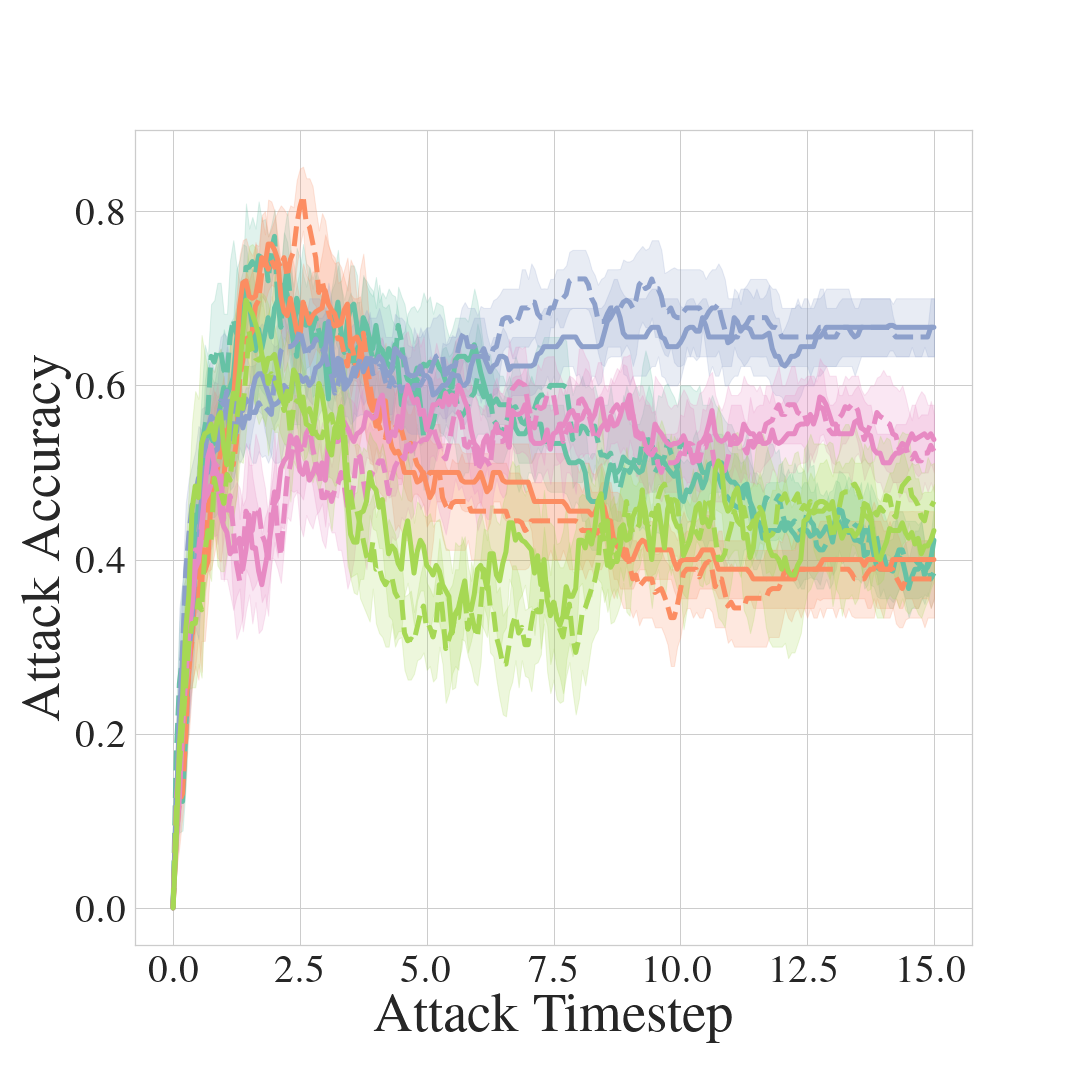}} &
    \subfloat[Test-Time @SoftAcc]{\includegraphics[width=0.2\linewidth,trim={0cm 0cm 5cm 0cm},clip]{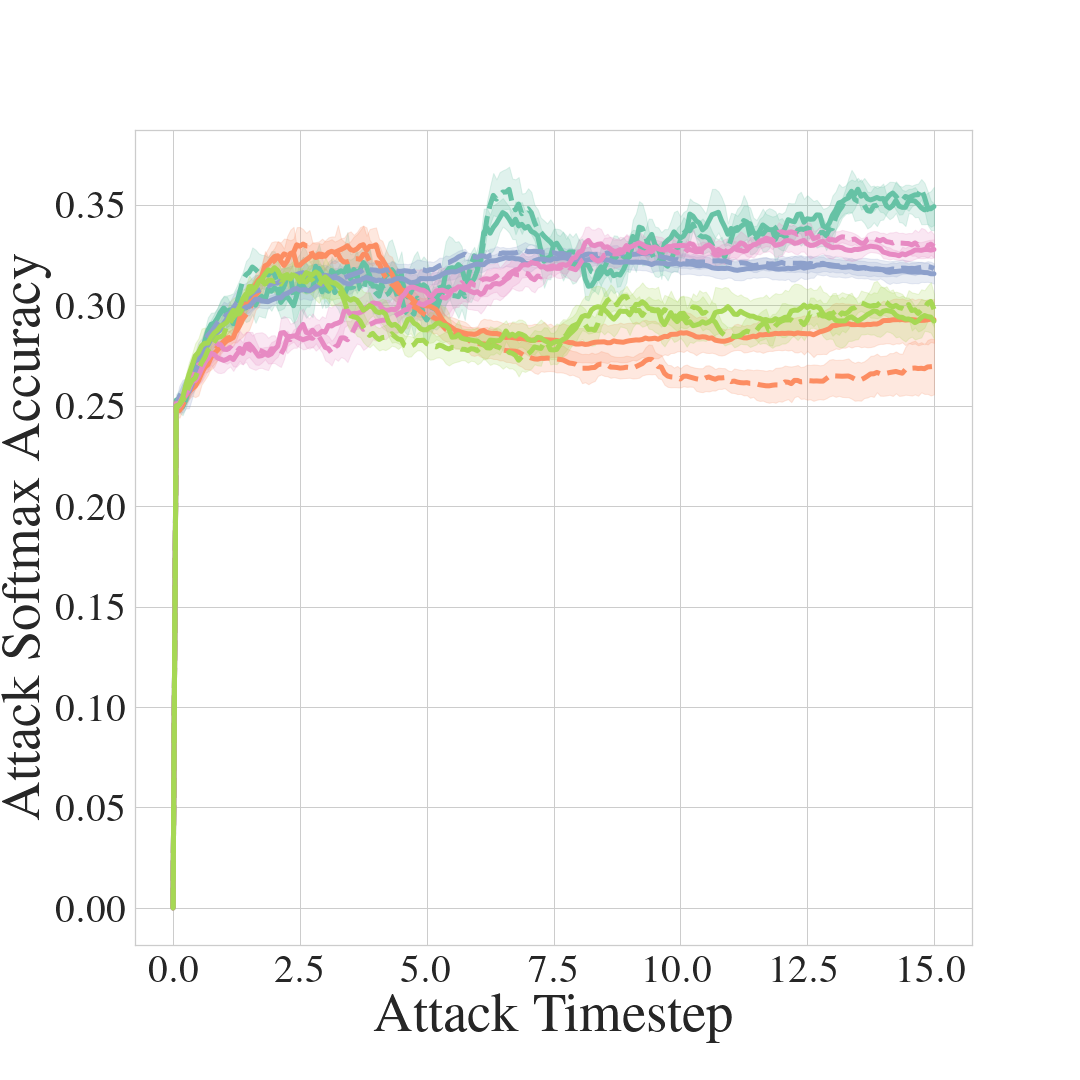}} &
    \subfloat[Test-Time @Effort]{\includegraphics[width=0.2\linewidth,trim={0cm 0cm 5cm 0cm},clip]{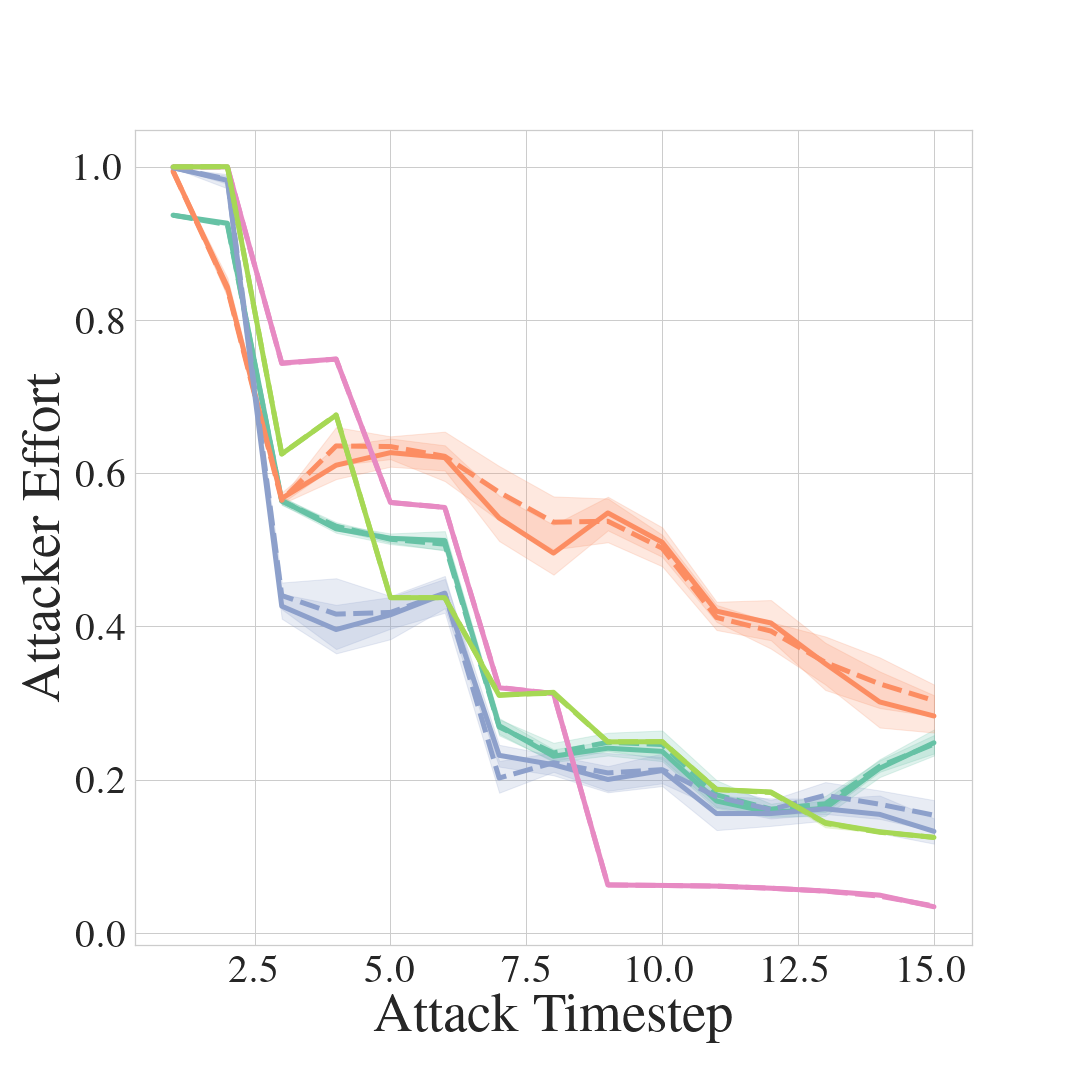}} &
    \subfloat[Test-Time @Time]
    {\includegraphics[width=0.2\linewidth,trim={0cm 0cm 5cm 0cm},clip]{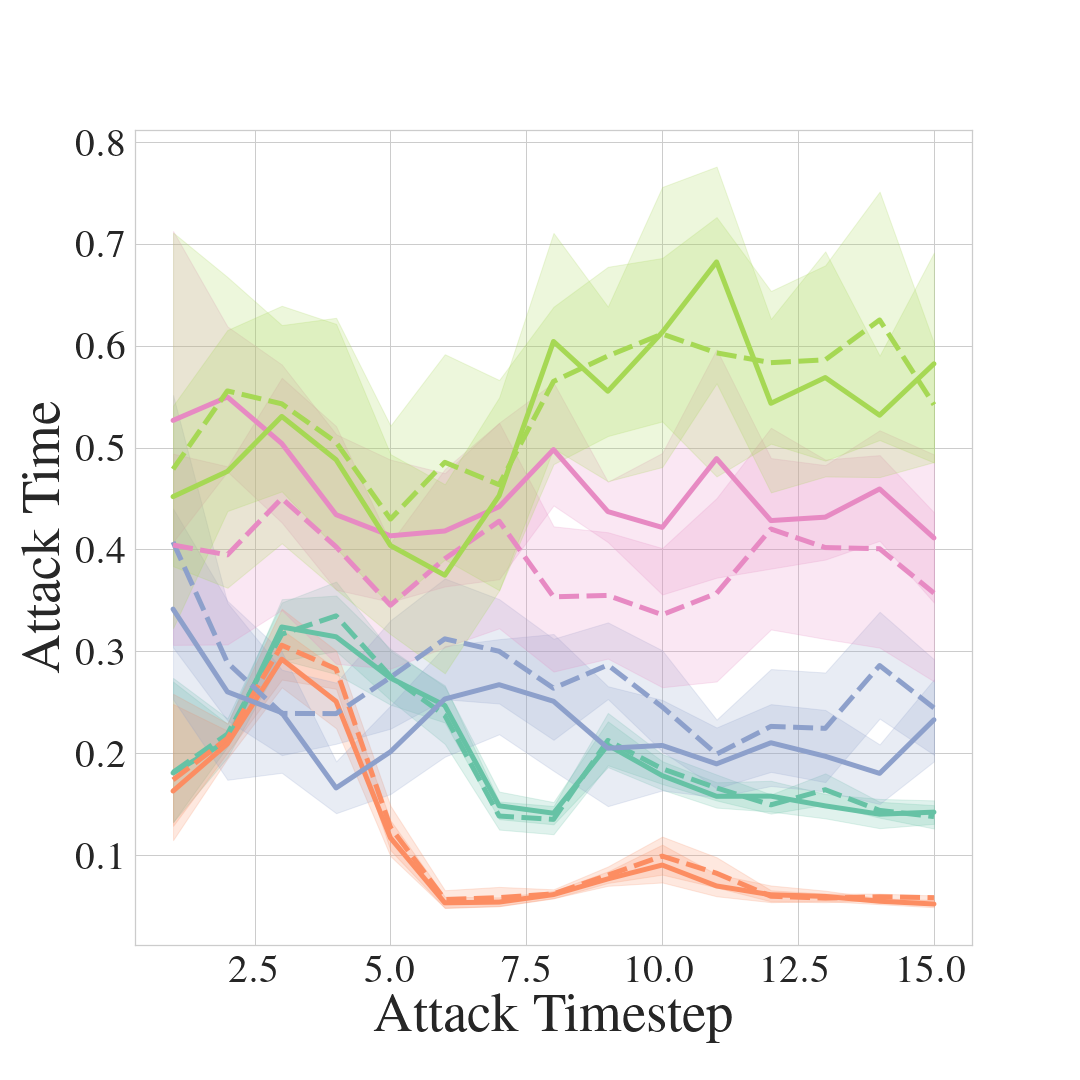}} \\
    \end{tabular}
    \caption{Training-Time statistics (a-c, e-h) and Test-Time performance (i-l) w.r.t. Accuracy (@Acc), Softmax Accuracy (@SoftAcc), Effort (@Effort), and Time (@Time) of $\gamma$DDPG with fixed Bellman discounts 0.80, 0.85, 0.90, 0.95, and 0.99. The dotted graphs in Test-Time plots (i-l) represent attacks on victims initialised with random numbers using different seeds.}
\label{Fig:study1}%
\end{figure*}

Study 1 is designed to demonstrate the capability of the discount factor to function as a means of bounding the lower-priority objective (minimise @Effort) while reducing the effect of uncertainty so as to aid in the optimisation of the primary objective (maximise @Acc) in a high-dimensional space. Figure \ref{Fig:study1} shows that strategies found by lower discount factors during training exert a slightly higher @Effort to achieve high @Acc and @SoftAcc. This implies that reducing the search space around the current state reduces the effect of uncertainty in the high-dimensional black-box (partially-observable) setting; enabling $\gamma$DDPG to find strategies that achieve high @Acc while exerting a bounded @Effort.
    

Additionally, the variance of @Acc and @SoftAcc during training increases with increasing $\gamma$. An increased value of $\gamma$ implies that the optimisation algorithm is taking into consideration and allotting importance to rewards further in the future. This reflects the difficulty faced by RL algorithms while optimising in large high-dimensional non-convex spaces and illustrates the potential of the discount factor in facilitating optimisation by reducing/bounding these spaces.

Figure \ref{Fig:study1} compares the test-time performance of the best strategies (highest mean value) found using fixed discounts 0.80, 0.85, 0.90, 0.95, and 0.99 respectively. As suggested by training-time statistics, the strategies that are learned using smaller fixed-discounts exert higher effort during test-time. However, unlike the trend observable in training-time histograms, smaller fixed-discounts do not achieve highest @Acc during test-time. This discrepancy is due to the low generalisability of the best attack strategies found using smaller fixed-discount models. But, interestingly, instead of taking approximately equal time to carry out the attack, lower fixed-discount strategies execute faster attacks. Overall, fixed-discount 0.90 achieves the best performance during test-time evaluation as it achieves the highest @Acc with bounded @Effort, high @SoftAcc, and low @Time. 

As fixed discount 0.90 offers the best balance between @Acc and @Effort, it is chosen as the best fixed discount in this research. This study demonstrates the capability of the discount factor to encode and prioritise attack objective(s). However, the fixed-discount approach requires a grid-search to be performed in order to identify the optimal fixed discount factor ($\gamma$) for attacking a given victim in the given victim environment. Furthermore, the fixed-discount approach cannot be applied in settings where the victim task and/or the victim environment change during the victim's training; in a manner that causes modification of the value of the optimal discount factor for the attacker. These problems associated with fixed discounts are solved in the current work with the aid of effort-based and effort+accuracy-based adaptive/dynamic discount functions.

\begin{figure*}[htbp]%
    \begin{tabular}{cccc}
    \subfloat[Accuracy KDE]{\includegraphics[width=0.2\linewidth,trim={0cm 0cm 9cm 0cm},clip]{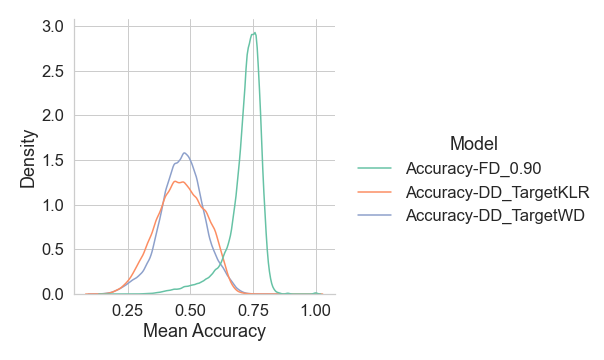}} &
    \subfloat[SoftMax Accuracy KDE]{\includegraphics[width=0.2\linewidth,trim={0cm 0cm 9cm 0cm},clip]{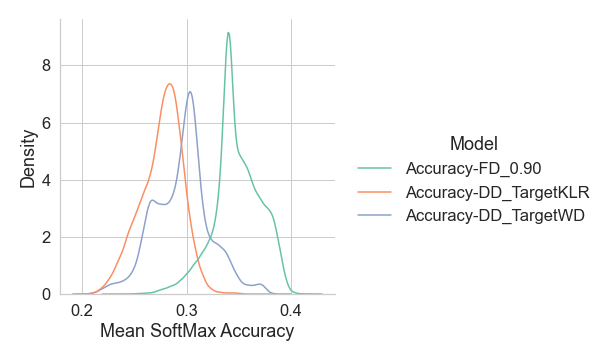}} &
    \subfloat[Effort KDE]{\includegraphics[width=0.2\linewidth,trim={0cm 0cm 9cm 0cm},clip]{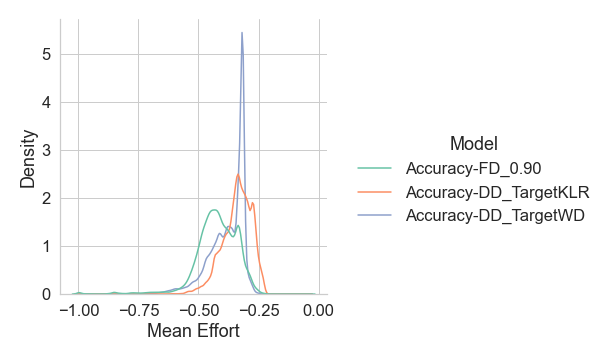}} &
    \subfloat[Legend]{\includegraphics[width=0.2\linewidth,trim={13cm 3cm 0cm 2cm},clip]{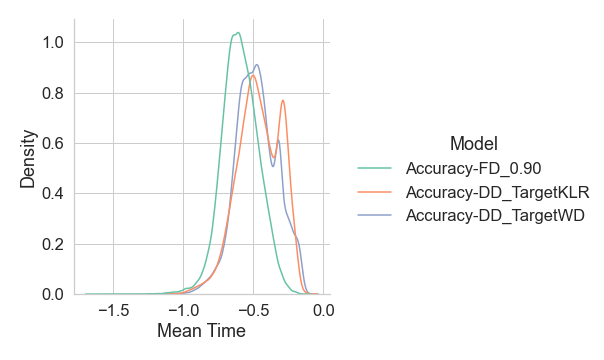}}\\

    \subfloat[Accuracy Line Graph]{\includegraphics[width=0.24\linewidth,trim={2.5cm 0cm 2.5cm 2.5cm},clip]{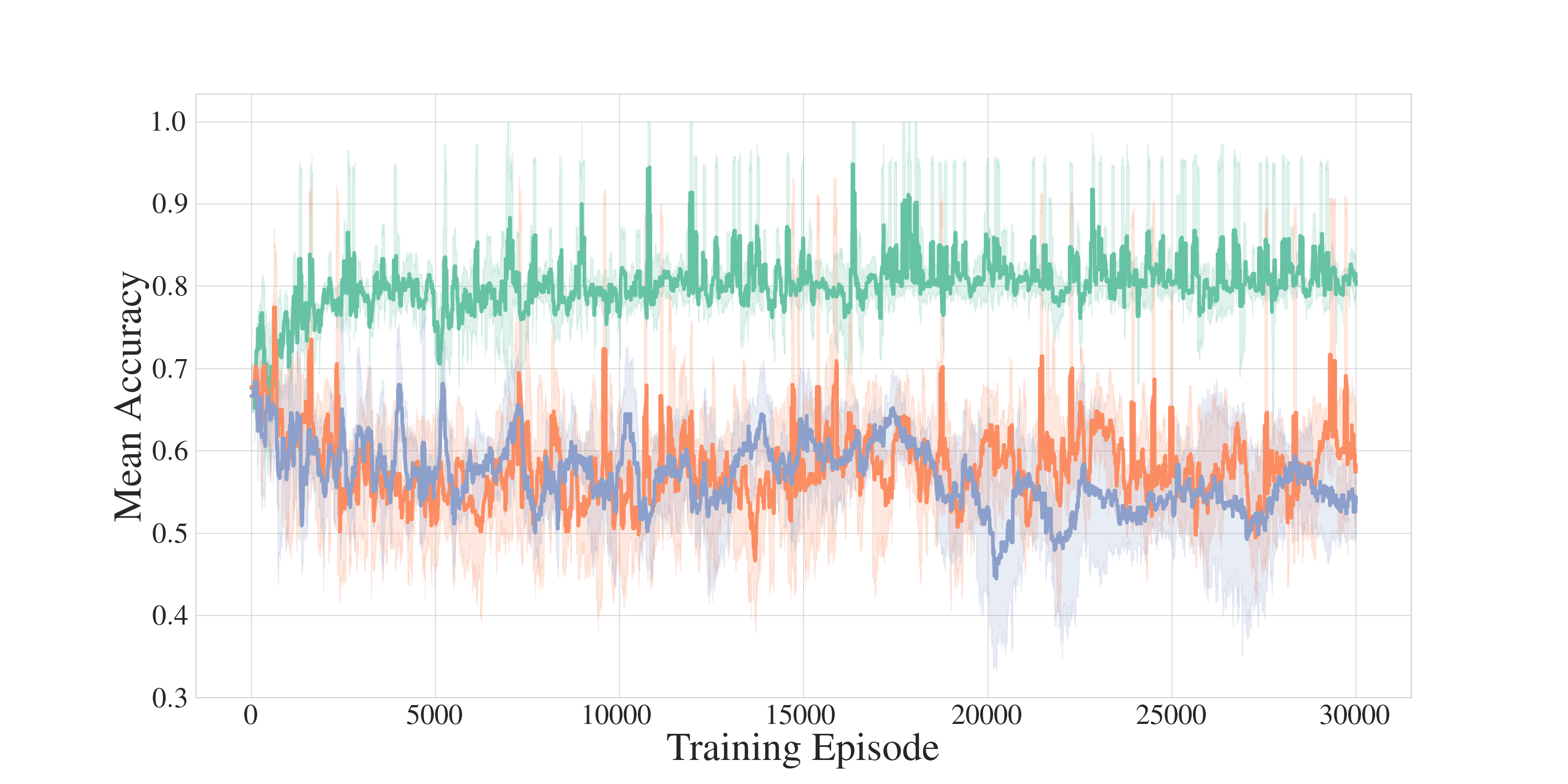}} &
    \subfloat[Softmax Accuracy Line Graph]{\includegraphics[width=0.24\linewidth,trim={2.5cm 0cm 2.5cm 2.5cm},clip]{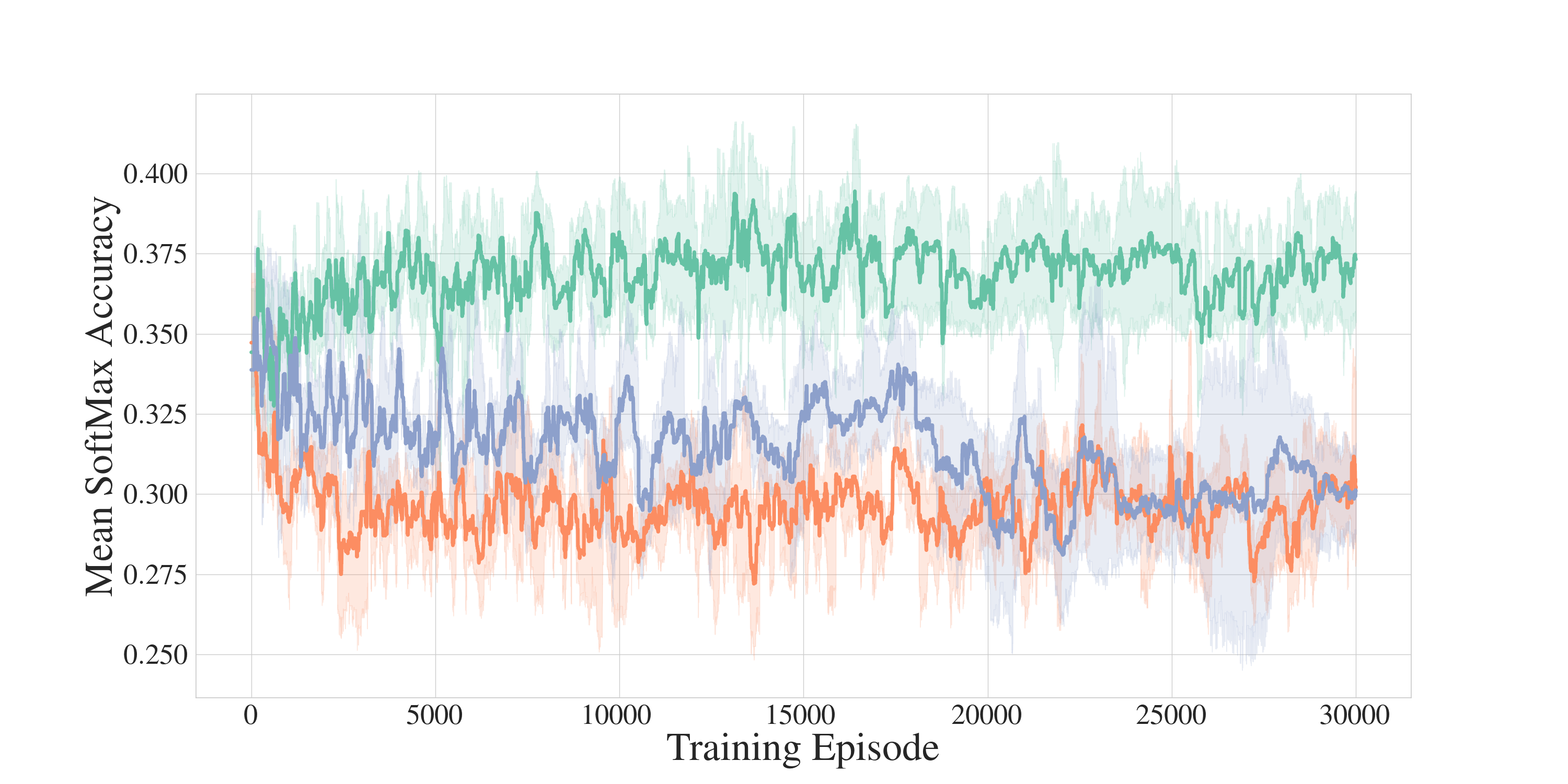}} &
    \subfloat[Effort Line Graph]{\includegraphics[width=0.24\linewidth,trim={2.5cm 0cm 2.5cm 2.5cm},clip]{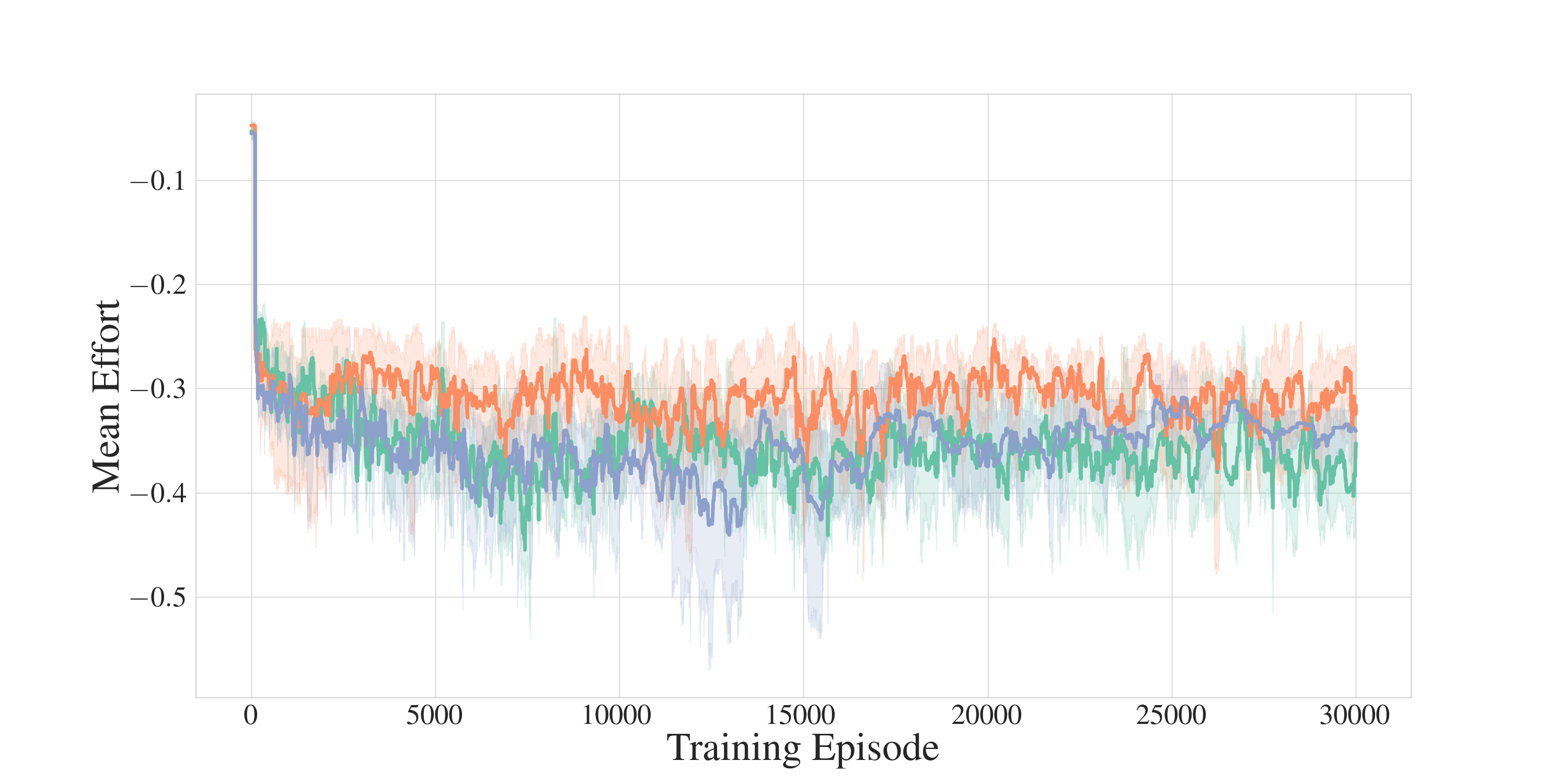}} &
    \subfloat[Time Line Graph]{\includegraphics[width=0.24\linewidth,trim={2.5cm 0cm 2.5cm 2.5cm},clip]{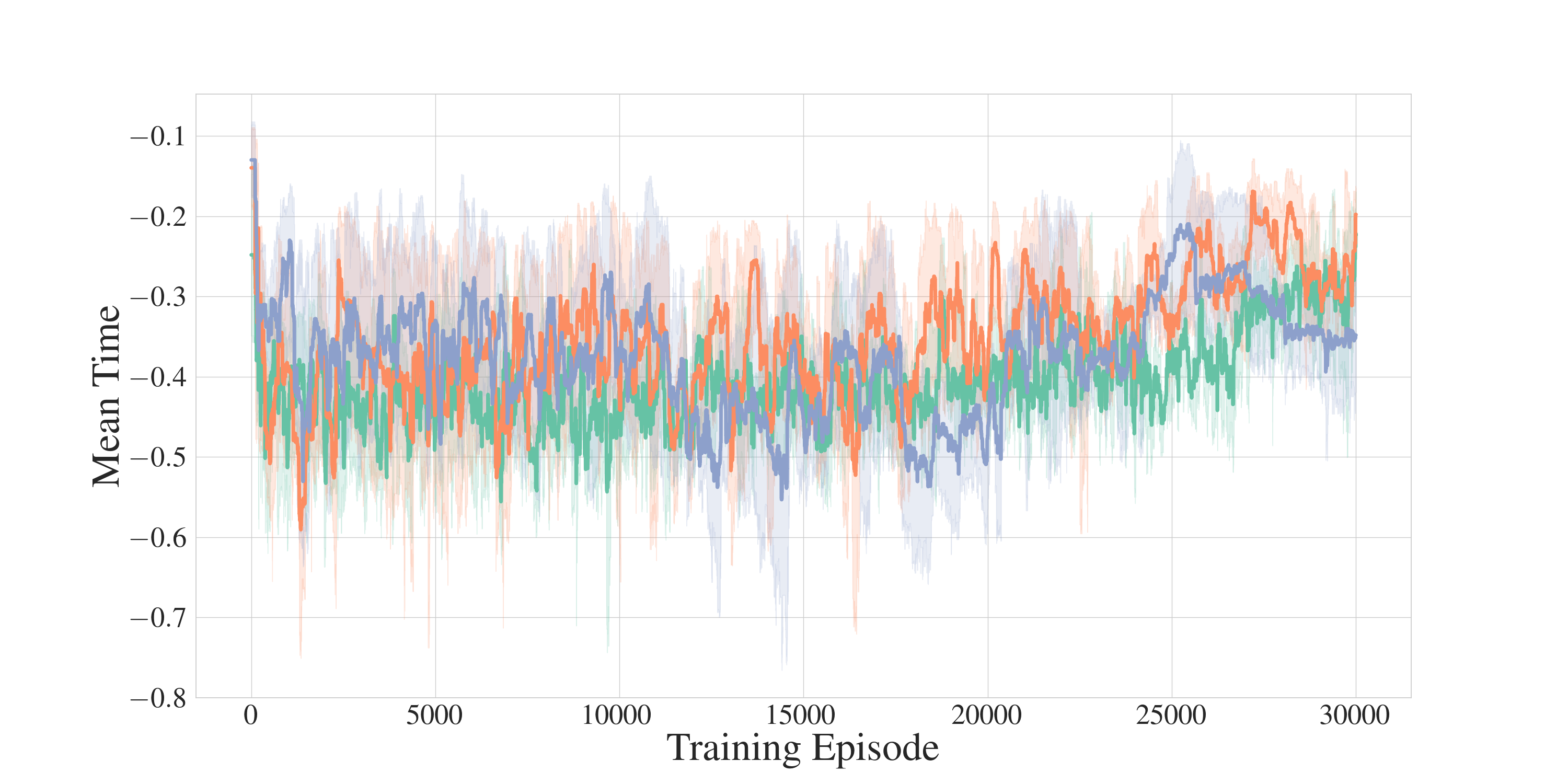}}\\

    \subfloat[Test-Time @Acc]{\includegraphics[width=0.2\linewidth,trim={0cm 0cm 5cm 0cm},clip]{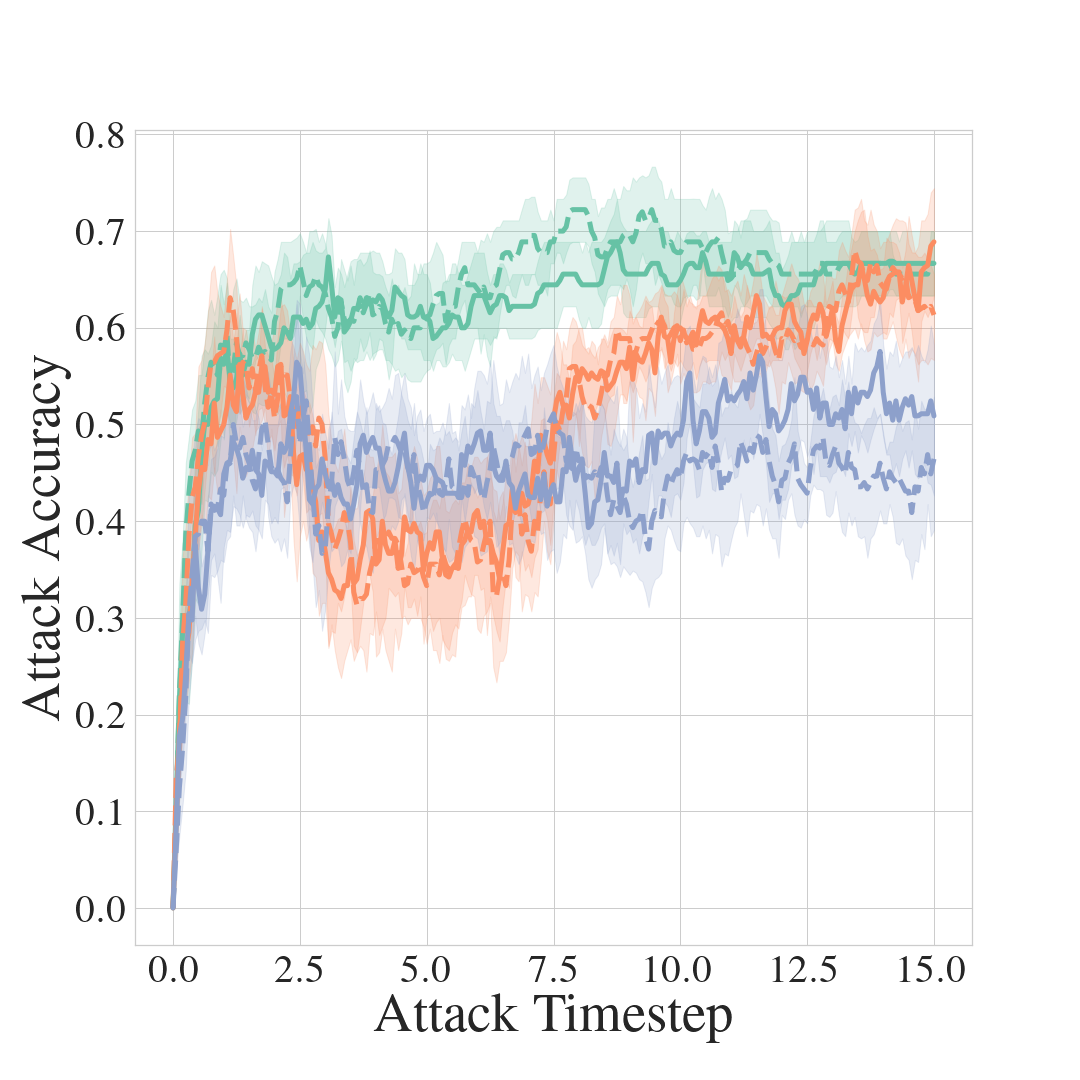}} &
    \subfloat[Test-Time @SoftAcc]{\includegraphics[width=0.2\linewidth,trim={0cm 0cm 5cm 0cm},clip]{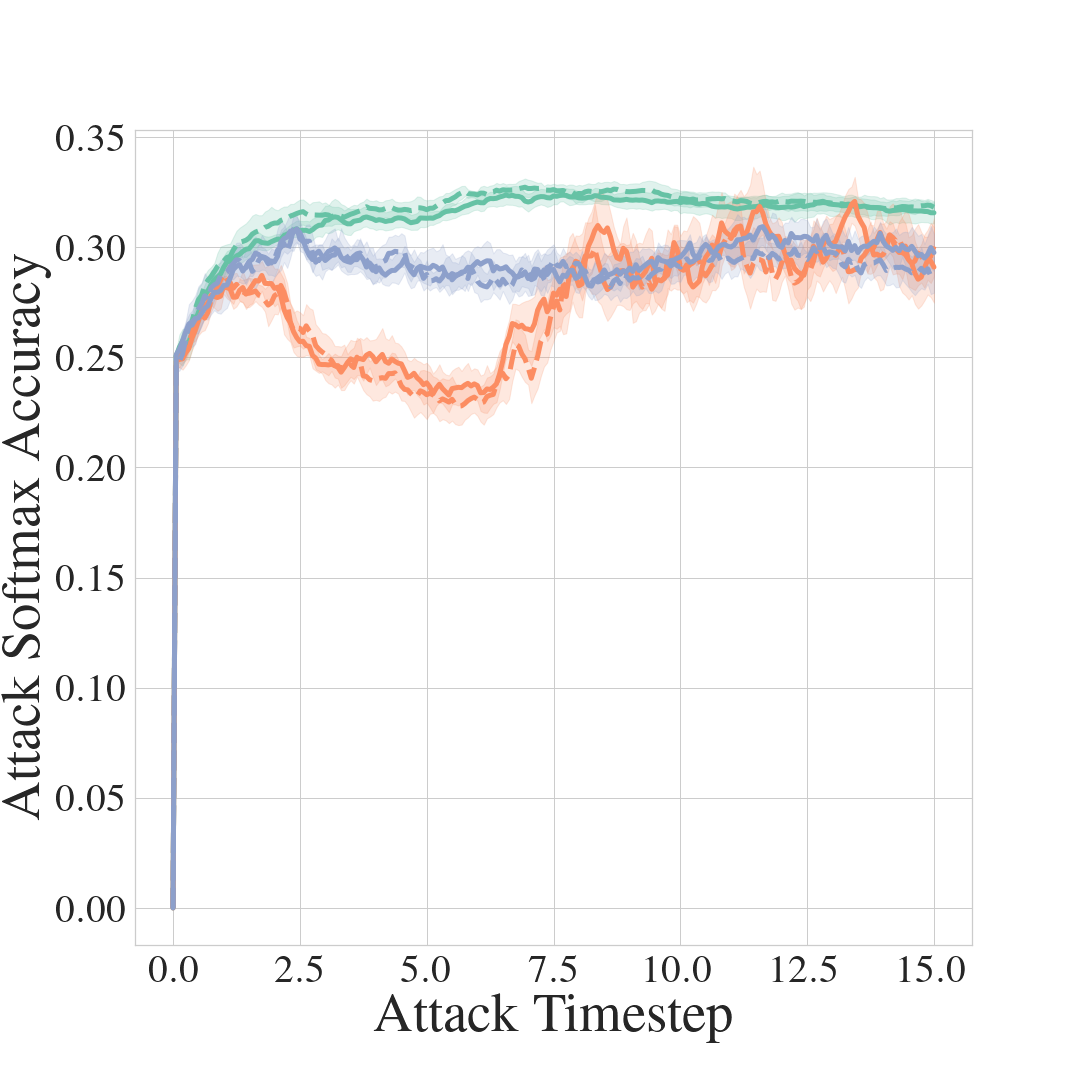}} &
    \subfloat[Test-Time @Effort]{\includegraphics[width=0.2\linewidth,trim={0cm 0cm 5cm 0cm},clip]{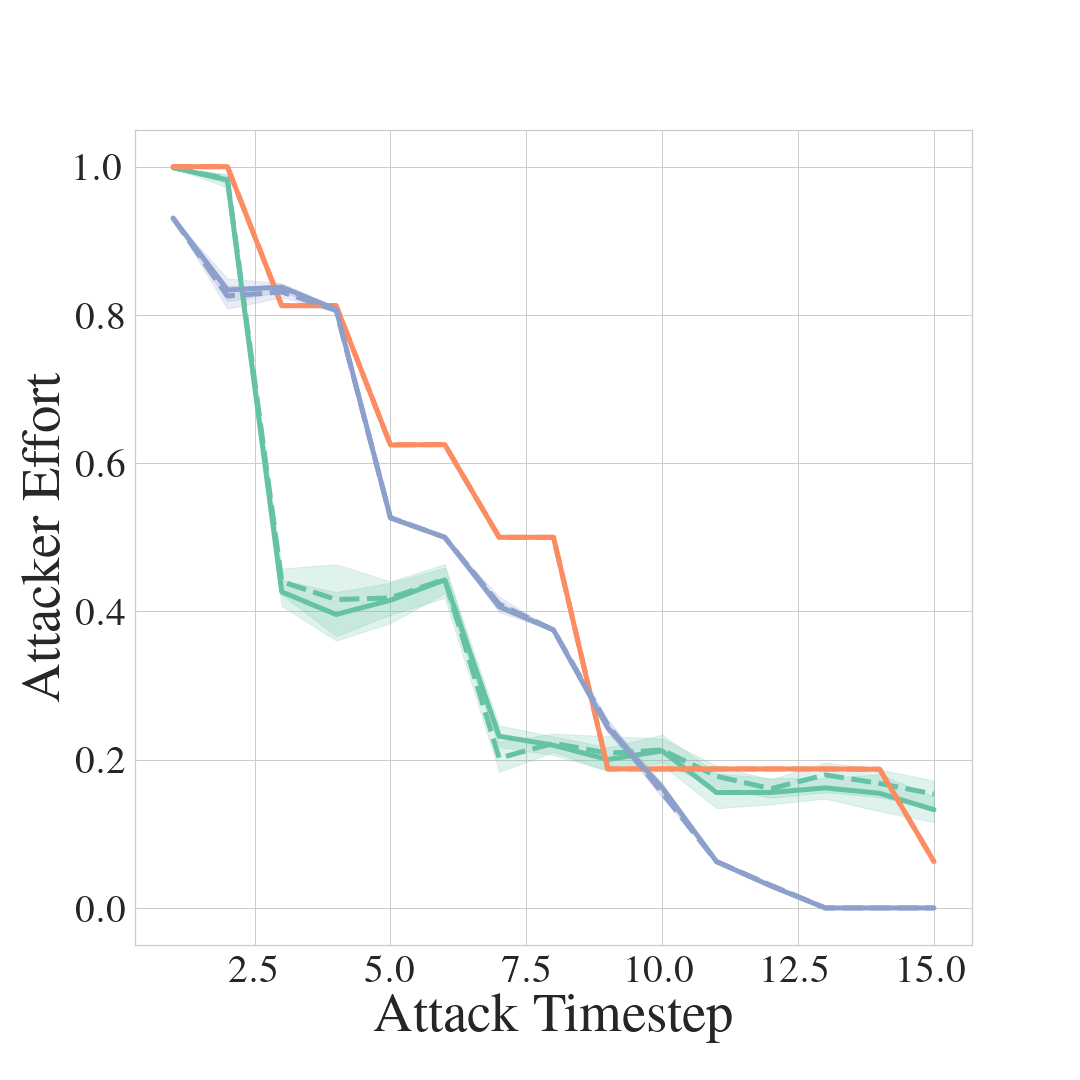}} &
    \subfloat[Test-Time @Time]
    {\includegraphics[width=0.2\linewidth,trim={0cm 0cm 5cm 0cm},clip]{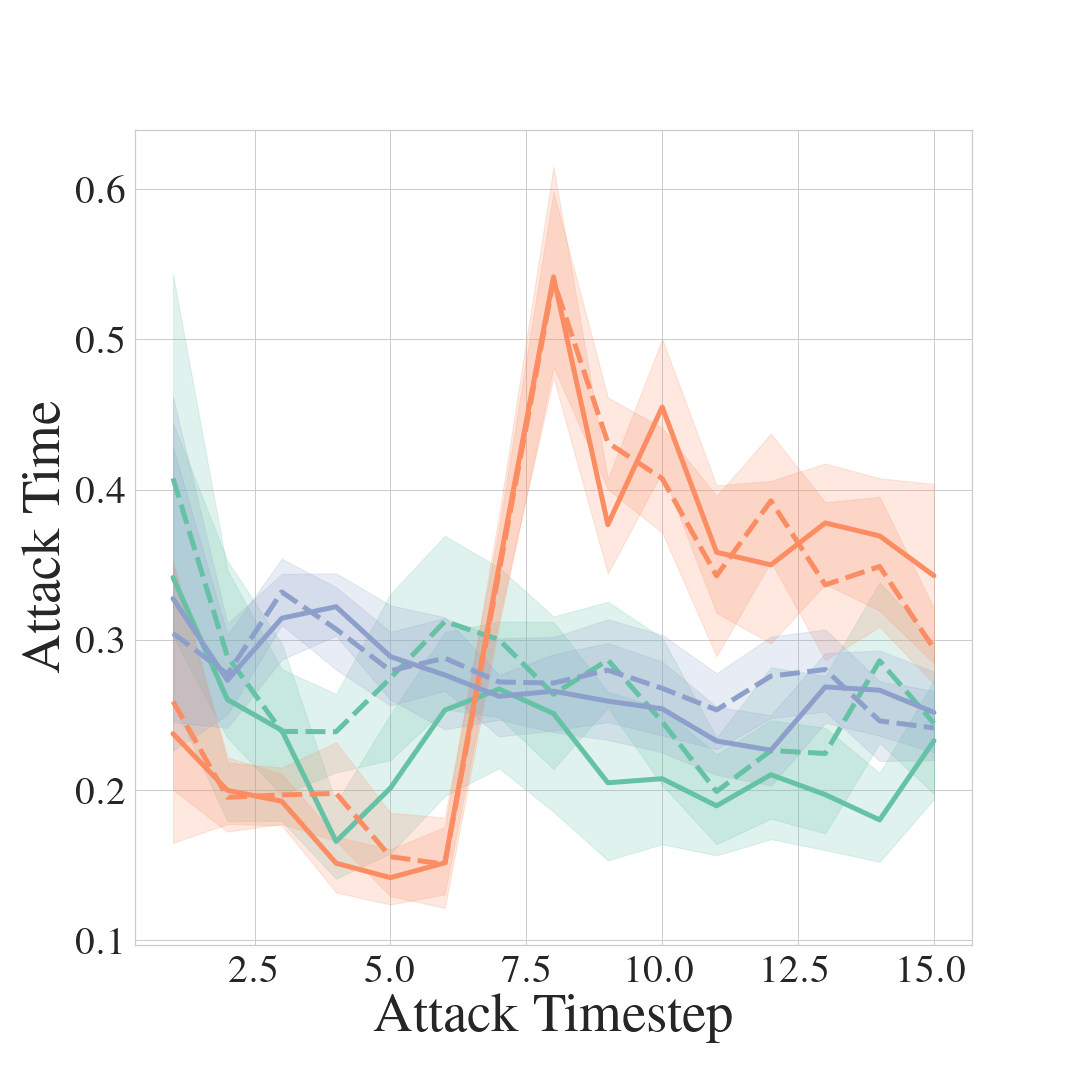}} \\
    \end{tabular}
    \caption{Training-Time statistics (a-c, e-h) and Test-Time performance (i-l) w.r.t. Accuracy (@Acc), Softmax Accuracy (@SoftAcc), Effort (@Effort), and Time (@Time) of $\gamma$DDPG with best fixed-discount (0.90) and dynamic discounts TargetKLR and TargetWD. The dotted graphs in Test-Time plots (i-l) represent attacks on victims initialised with random numbers using different seeds.}
\label{Fig:study3}%
\end{figure*}
            
Study 3 presented in Figure \ref{Fig:study3} compares TargetKLR and TargetWD dynamic discounts to the best fixed-discount found in Study 1. $\gamma$DDPG with TargetKLR and TargetWD dynamic discounts frequently finds strategies with @Acc above 0.5. Herein, strategies found by TargetWD achieve better @SoftAcc than strategies found by TargetKLR while fixed-discount 0.90 frequently finds strategies with better @Acc as well as better @SoftAcc than both the dynamic discounts. However, most of the strategies found by TargetWD and TargetKLR achieve these accuracies while executing lower level of @Effort on the victim environment than fixed-discount 0.90. Furthermore, the @Time graph shows that both the dynamic discounts find strategies that take lesser time to carry out the attack compared to strategies found by the best fixed-discount. These results imply that in the given setting, fixed-discount 0.90 achieves the best training-time statistics while TargetWD is better than TargetKLR with respect to @SoftAcc and TargetKLR is better than TargetWD with respect to @Effort.

The test-time performance of the best strategies found by fixed-discount (0.90) and dynamic discounts TargetKLR and TargetWD are presented in plots i-l of Figure \ref{Fig:study3}. These plots show that TargetKLR generalises better than TargetWD with respect to @Acc and thereby achieves a similar @Acc of $\sim$0.65 as the fixed-discount 0.90 by the end of the attack while TargetWD achieves a lower @Acc of $\sim$0.5. However, TargetWD is able to reduce @Effort to 0.0 by the end of the attack while TargetKLR and fixed-discount 0.90 continue to exert effort between 0.1 and 0.2 until the end of the attack. Lastly, TargetWD and fixed-discount 0.90 execute a faster attack than TargetKLR. Given that @Acc is the higher-priority objective for the attacker, TargetKLR is the better effort-based dynamic discount in this research.

    \subsection{Significance of Improvement in Attack Performance}
    \label{subsec:ExtendedExperiments/significance}

    \begin{table}[tbp]
        \centering
        \caption{Wilcoxon Signed-Rank Test to Establish Significance of Difference in Test-Time Mean Attack Accuracy (@Acc) of Different Attack Models}
        \begin{tabulary}{\textwidth}{LCCCC}
            \hline
            \textbf{Attack Model 1} & \textbf{Attack Model 2} & \textbf{Statistic} & \textbf{P Value} \\
            \hline
            DD WD & FD 0.90 & 210.0 & 9.53674e-07 \\ 
            DD WD & DD KLR & 210.0 & 9.53674e-07 \\
            DD WD & DD TargetKLR & 210.0 & 9.53674e-07 \\
            DD WD & DD TargetWD & 210.0 & 9.53674e-07 \\
            DD WD & TEPA & 210.0 & 9.53674e-07 \\
            \hline
        \end{tabulary}
        \label{tbl:sigAccuracy}
    \end{table}

    \begin{table}[tbp]
        \centering
        \caption{Wilcoxon Signed-Rank Test to Establish Significance of Difference in Test-Time Mean Attacker Effort (@Effort) of Different Attack Models}
        \begin{tabulary}{\textwidth}{LCCCC}
            \hline
            \textbf{Attack Model 1} & \textbf{Attack Model 2} & \textbf{Statistic} & \textbf{P Value} \\
            \hline
            DD WD & FD 0.90 & 0.0 & 9.53674e-07 \\
            DD WD & DD KLR & 0.0 & 9.53674e-07 \\
            DD WD & DD TargetKLR & 0.0 & 9.53674e-07 \\
            DD WD & DD TargetWD & 0.0 & 9.53674e-07 \\
            DD WD & TEPA & 0.0 & 9.53674e-07 \\
            \hline
        \end{tabulary}
        \label{tbl:sigEffort}
    \end{table}

    $\gamma$DDPG with WD dynamic discount achieves the best test-time performance with respect to all four performance metrics as shown in plots i-l in Figures 2 and 3 of the main paper. The significance of this performance is tested with the Wilcoxon Signed-Rank Test and the results are presented in Tables \ref{tbl:sigAccuracy} and \ref{tbl:sigEffort} where DD refers to Dynamic Discount while FD refers to Fixed Discount.

    As mentioned in Section 4 of the main paper, the test-time evaluation of the attack models is conducted by carrying out 20 separate attacks with each attack model. Within these 20 attacks, 10 attacks are carried out on victim agents initialised with the same random number generation seed (as the seed used during training of the attack model) and 10 attacks are carried out on victim agents initialised with different random number generation seeds. As all attack models are tested on randomly and independently sampled victim agents (with same/different seed) from the same population; and the mean performance metrics (@Acc, @Effort) are continuous variables that cannot be assumed to follow a normal distribution, the Wilcoxon Signed-Rank Test is used to test the following two hypothesis for @Acc and @Effort respectively:

    \noindent Hypothesis for mean @Acc:
    \begin{itemize}
        \item \textbf{Null hypothesis (H0):} The median difference between mean @Acc of attacks executed by two given attack models is zero
        \item \textbf{Alternative hypothesis (HA):} The median difference is positive $\alpha$ = 0.005
    \end{itemize}

    \noindent Hypothesis for mean @Effort:
    \begin{itemize}
        \item \textbf{Null hypothesis (H0'):} The median difference between mean @Effort of attacks executed by two given attack models is zero
        \item \textbf{Alternative hypothesis (HA'):} The median difference is negative $\alpha$ = 0.005
    \end{itemize}

    Tables \ref{tbl:sigAccuracy} and \ref{tbl:sigEffort} present the significance of the difference in mean @Acc and mean @Effort achieved by attack model $\gamma$DDPG with WD dynamic discount as compared to every other attack model. Herein, the statistic is the sum of the ranks of the differences above zero, and p-value is the probability of getting a test statistic as large or larger assuming both distributions are the same. As the statistic $>$ 173 and p-value $<$ 0.005 for each comparison in Table \ref{tbl:sigAccuracy}, we can reject the null hypothesis in favour of the alternative i.e. the mean @Acc achieved by $\gamma$DDPG with WD is significantly larger than the mean @Acc of every other attack model. Similarly, as the statistic $<$ 37 and p-value $<$ 0.005 for each comparison in Table \ref{tbl:sigEffort}, we can reject the null hypothesis in favour of the alternative i.e. the mean @Effort achieved by $\gamma$DDPG with WD is significantly smaller than the mean @Effort of every other attack model. In fact, for each of the 20 attacks executed by all the attack models, $\gamma$DDPG with WD dynamic discount achieves a larger mean @Acc and a smaller mean @Effort than every other attack model.

\FloatBarrier

\end{document}